%% file: neurips_2024.tex
\documentclass{article}

    \usepackage[preprint]{neurips_2024}

\usepackage[utf8]{inputenc} 
\usepackage[T1]{fontenc}    
\usepackage{hyperref}       
\usepackage{url}            
\usepackage{booktabs}       
\usepackage{amsfonts}       
\usepackage{nicefrac}       
\usepackage{microtype}      
\usepackage{xcolor}         
\usepackage{algorithm}
\usepackage{algpseudocode}
\usepackage{microtype}
\usepackage{graphicx}
\usepackage{subfigure}
\usepackage{booktabs} 
\usepackage{xcolor}
\usepackage{soul}
\usepackage{epsfig}
\usepackage{subfigure}
\usepackage{booktabs}
\usepackage{caption}
\usepackage{array}
\usepackage{etoolbox}
\usepackage{tabulary}
\usepackage{colortbl}
\usepackage{bbding}
\usepackage{wrapfig}
\usepackage{multirow}
\usepackage{soul}
\usepackage{amsmath}
\usepackage{hyperref}
\usepackage{tabularx}
\usepackage{amssymb}
\usepackage{multirow}
\usepackage{pifont} 
\usepackage{wrapfig}
\usepackage{amssymb}
\usepackage{mathtools}
\usepackage{amsthm}

\title{Two Trades are not Baffled: Condensing Graph via Crafting Rational Gradient Matching}

\author{
Tianle Zhang$^{1}$\footnotemark[1],\;   
Yuchen Zhang$^{1}$\footnotemark[1],\; 
Kun Wang$^{1}$,\; 
Kai Wang$^{1}$\footnotemark[3],\; 
Beining Yang$^{1}$,\; \;\\ 
\textbf{Kaipeng Zhang}$^{3}$\textbf{,}\; 
\textbf{Wenqi Shao}$^{3}$\textbf{,}\;  
\textbf{Ping Liu}$^{4}$\textbf{,} \;  
\textbf{Joey Tianyi Zhou}$^{2}$\footnotemark[2]\textbf{,} \;    
\textbf{Yang You}$^{1}$\footnotemark[2].
	\\
	$^{1}$National University of Singapore \quad $^{2}$A*STAR Centre for Frontier AI Research (CFAR) \quad \\ $^{3}$Shanghai AI Laboratory 
	$^{4}$University of Nevada, Reno
}

\begin{document}

\maketitle
\renewcommand{\thefootnote}{\fnsymbol{footnote}} 
\footnotetext[1]{Equal contribution} 
\footnotetext[2]{Corresponding authors} 
\footnotetext[3]{Project lead} 

\begin{abstract}
Training on large-scale graphs has achieved remarkable results in graph representation learning, but its cost and storage have raised growing concerns. 
As one of the most promising directions, graph condensation methods address these issues by employing gradient matching, aiming to condense the full graph into a more concise yet information-rich synthetic set. Though encouraging, these strategies primarily emphasize matching directions of the gradients, which leads to deviations in the training trajectories. 
Such deviations are further magnified by the differences between the condensation and evaluation phases, culminating in accumulated errors, which detrimentally affect the performance of the condensed graphs.
In light of this, we propose a novel graph condensation method named \textbf{C}raf\textbf{T}ing \textbf{R}ationa\textbf{L} trajectory (\textbf{CTRL}), which offers an optimized starting point closer to the original dataset's feature distribution and a more refined strategy for gradient matching. 
Theoretically, CTRL can effectively neutralize the impact of accumulated errors on the performance of condensed graphs.  
We provide 
extensive experiments on various graph datasets and downstream tasks to support the effectiveness of CTRL. 
\end{abstract}

\input{1.Introduction}
\input{3.Method}
\input{4.Experiments}

\input{5.Conclusion}


\bibliographystyle{unsrt}



\appendix
\onecolumn
\input{2.Related_work}
\section{Datasets and implementation details}\label{section: details}
\subsection{Datasets}\label{section: details_data}
We evaluate CTRL on three transductive datasets: Cora, Citeseer~\cite{kipf2016semi}, Ogbn-arxiv~\cite{hu2020open}, Ogbn-xrt~\cite{chien2021node}, and two inductive datasets: Flickr~\cite{zeng2020graphsaint}, and Reddit~\cite{hamilton2017inductive}.
We obtain all datasets from PyTorch Geometric\cite{fey2019fast} with publicly available splits and consistently utilize these splits in all experiments. For graph classification, we use multiple molecular datasets from Open Graph Benchmark (OGB)~\cite{hu2020open} and TU Datasets (MUTAG and NCI1)~\cite{morris2020tudataset} for graph-level property classification, and one superpixel dataset CIFAR10~\cite{dwivedi2020benchmarking}. In addition, we use Pubmed\cite{kipf2016semi} in our neural architecture search (NAS) experiments. Dataset statistics are shown in Table \ref{tab:ncdatasets} and \ref{tab:gcdatasets}.

\input{tables/NC_dataset}
\input{tables/GC_dataset}

\subsection{Implementation Details}\label{section: details_im}
During the implementation of CTRL, considering datasets with a large number of nodes, where the feature distribution plays a crucial role in the overall data, such as Ogbn-arxiv, Ogbn-xrt, and Reddit, we employ the specific initialization method depicted in Sec. \ref{sec:CTRL} on them. 
Additionally, due to the substantial volume of data and numerous gradient matching iterations, we introduce a threshold $tau$ during the matching process. We match gradients smaller than this threshold on both direction and magnitude, while gradients exceeding the threshold were matched solely based on their directions. 
Through extensive experimentation, we observe that this strategy reduced computational costs and led to further performance improvements. For other datasets, due to the smaller dataset size, we utilize both gradient magnitude and direction for matching throughout the entire matching process and we directly employ random sampling for initialization. For graph classification tasks, we employ gradient magnitude and direction matching throughout the graph condensation and adopt a strategy of random sampling for initialization. And the beta in Fig.~\ref{fig:reddit}and~\ref{fig:ogb} are 0.1 and 0.15, respectively.

\subsection{Implementation Details of Ablation Experiments}\label{sec_abl}
The values of $\beta$ in the training of models in Fig.\ref{fig:reddit} and Fig.\ref{fig:ogb} are 0.1 and 0.15, respectively. 
In Fig.\ref{fig:beta}, we employ the specific initialization method(K-MEANS-based) in CTRL for Ogbn-arxiv, and random sampling for Cora and Citeseer, all the results are obtained by repeating experiments 5 times.

\subsection{Hyper-parameter Setting}
For node classification, without specific mention, we adopt a 2-layer SGC~\cite{wu2019simplifying} with 256 hidden units as the GNN used for gradient matching. We employ a multi-layer perceptron (MLP) as the function $g_{\Phi}$ models the relationship between $\mathbf{A}^{\prime}$
and $\mathbf{X}^{\prime}$. Specifically, we adopt a 3-layer MLP with 128 hidden units for small graphs (Cora and Citeseer)
and 256 hidden units for large graphs (Flickr, Reddit, and Ogbn-arxiv). We tune the training epoch for CTRL in a range of {400, 600, 1000}. 
For the choices of condensation ratio $r$, we divide the discussion into two parts. The first part is about transductive datasets. For Cora and Citeseer, since their labeling rates are very small (5.2\% and 3.6\%, respectively), we choose r to be {25\%, 50\%, 100\%} of the labeling rate. Thus, we finally
choose {1.3\%, 2.6\%, 5.2\%} for Cora and {0.9\%, 1.8\%, 3.6\%} for Citeseer. For Ogbn-arxiv and Ogbn-xrt, we choose r to be {0.1\%, 0.5\%, 1\%} of its labeling rate (53\%), thus being {0.05\%, 0.2\%, 0.5\%}. 
The second part is about inductive datasets. As the nodes in the training graphs are all labeled in inductive datasets, we choose {0.1\%, 0.5\%, 0.1\%} for Flickr and 0.05\%, 0.1\%, 0.2\% for Reddit. The $beta$ that measures the weight of the direction and magnitude is in {0.1, 0.2, 0.5, 0.7, 0.9}. For graph classification, we vary the number of learned synthetic graphs per class in the range of {1, 10, 50} ({1, 10} for MUTAG, in the case where the condensation ratio has already reached 13.3\% with a factor of 10, and we will refrain from further increasing the condensation ratio to ensure the effectiveness of data condensation.) and train a GCN on these graphs. The $beta$ that measures the weight of the direction and magnitude is between 0.05 and 0.9 for node classification tasks, while between 0.1 and 0.9 for graph classification tasks.


\section{More details about quantitative analysis}



\subsection{Multiple Metrics}\label{sec-metric}
We select six complementary metrics to provide a comprehensive evaluation of the distribution of high and low-frequency signals between the synthetic and original graphs. 
The proportion of low-frequency nodes directly quantifies the relative abundance of high versus low-frequency signals.
Meanwhile, the mean of the high-frequency signal region for each feature dimension can describe the right-shift phenomenon in the whole spectrum~\cite{pmlr-v162-tang22b} and get a measure of the overall strength of the high-frequency content.
Additionally, spectral peakedness and skewness directly characterize the shape of the frequency distribution~\cite{gallier2016spectral,ao2022spectral}. 
Spectral radius indicates the overall prevalence of high-frequency signals, with a larger radius corresponding to more high-frequency content~\cite{gao2017spectral,fan2022toughness}. 
Finally, the variance of eigenvalues measures the spread of the frequency distribution~\cite{Min_2016,sharma2019note}. 

Together, these metrics enable multi-faceted quantitative analysis of the frequency distribution from diverse perspectives, including intuitive reflection of frequency distribution, spectral characterization to assess distribution shapes, and statistical distribution properties like spread and central tendency.
To provide a more intuitive summarized measure of the similarity between the synthetic and original graphs across the six metrics, we first normalize all these metrics by simply multiplying or dividing by powers of 10 in order to mitigate the impact of scale differences, then we compute the Pearson correlation coefficient which reflects the overall trend, complementing the individual metric comparisons. 

\subsection{Details about Gradient Magnitudes and Graph Spectral Distribution.}\label{sec_cv}

\textbf{Measurement of graphs signals.} Graph spectral analysis provides insights into the frequency content of signals on graphs. This allows the graph signal to be decomposed into components of different frequencies based on the graph discrete Fourier transform (GDFT)~\cite{stankovic2019graph,cheng2022graph}, as shown in Eq. (\ref{eqn-3}) and Eq. (\ref{eqn-4}).

\begin{equation}\label{eqn-3}
\boldsymbol{L}=\boldsymbol{I}-\boldsymbol{D}^{-\frac{1}{2}}\boldsymbol{A}\boldsymbol{D}^{-\frac{1}{2}}\quad=\boldsymbol{U}\mathbf{\Lambda} \boldsymbol{U}^{T},
\end{equation}
\begin{equation}\label{eqn-4}
{\hat{\boldsymbol{x}}}=\boldsymbol{U}^{T}\boldsymbol{x},  
\quad {\hat{\boldsymbol{x}}}_{H}=\boldsymbol{U}^{T}_{H}\boldsymbol{x}, 
\quad {\hat{\boldsymbol{x}}}_{L}=\boldsymbol{U}^{T}_{L}\boldsymbol{x},
\end{equation}

where \textbf{\textit{L}} denotes a regularized graph Laplacian matrix, \textbf{\textit{U}} is a matrix composed of the eigenvectors of \textbf{\textit{L}}. $\boldsymbol{\Lambda}$ is a diagonal matrix with the eigenvalues of \textbf{\textit{L}} on its diagonal, and ${\boldsymbol{x}}=({x}_{1},{x}_{2},\cdots,{x}_{N})^{T}\in\mathbb{R}^{N}$ and $\hat{\boldsymbol{x}}=(\hat{x}_{1},\hat{x}_{2},\cdots,\hat{x}_{N})^{T}=\boldsymbol{U}^{T}\boldsymbol{x}\in\mathbb{R}^{N}$ represent a signal on the graph and the GDFT of \textbf{\textit{x}}, respectively, while $\boldsymbol{{U}}_{H}$ and $\boldsymbol{{U}}_{L}$ correspond to matrices containing eigenvectors associated with eigenvalues that are above and below the specified cutoff frequency $\tau$. Furthermore, $\hat{\boldsymbol{x}}_{H}$ and $\hat{\boldsymbol{x}}_{L}$ represent signals attributed to high-frequency and low-frequency components respectively.

However, this process is often time-consuming due to the Laplacian matrix decomposition required. Following the previous work~\cite{pmlr-v162-tang22b}, we introduce the definition of the high-frequency area, represented by Eq. (\ref{eqn-5}).

\begin{equation}\label{eqn-5}
S_\mathrm{high}=\frac{\sum_{k=1}^N\lambda_k\hat{x}_k^2}{\sum_{k=1}^N\hat{x}_k^2}=\frac{\boldsymbol{x}^T\boldsymbol{L}\boldsymbol{x}}
{\boldsymbol{x}^T\boldsymbol{x}},
\end{equation}

where $\hat{x}_{k}^{2}/\Sigma_{i=1}^{N}\hat{x}_{i}^{2}$ denotes the spectral energy distribution at $\lambda_k(1\leq k \leq N)$, as the spectral energy on low frequencies contributes less on $S_\mathrm{high}$, this indicator increases when spectral energy shifts to larger eigenvalues. Note that $S_\mathrm{high}$ is defined under the premise of \textbf{\textit{x}} being a one-dimensional vector. In subsequent computations, $S_\mathrm{high}$ is computed for each feature dimension and averaged.

\begin{figure}[t]
        \centering
        \subfigure[Train with SGC model]
        {\includegraphics[width=0.45\linewidth, angle=0]{figures/fre.pdf}
        \label{fig:fre}}
         \hspace{6.5mm}
        \subfigure[Train with ChebNet model]
        {\includegraphics[width=0.45\linewidth, angle=0]{figures/fre2.pdf}
        \label{fig:fre2}}
    \caption{(a) and (b) demonstrate the relationship between average gradient magnitude and graph spectral distribution for synthetic graphs conforming to the Erdos-Renyi model, training with SGC and ChebNet respectively. The curves shown are the result of fitting a fourth-order polynomial.
    Altering the frequency distribution causes a significant impact on the model gradient during training.}\label{fig:fres}
    \vspace{-5mm} 
\end{figure}

Employing the methods outlined in \cite{pmlr-v162-tang22b} to create artificial graphs, we first initialize the composite graphs with three graph models, Erdos-Renyi~\cite{Erd_s_2013}, Barabasi-Albert~\cite{stamatelatos2022exact}, and Watts-Strogatz~\cite{Song_2014}.
Subsequently, we conducted 3*4 sets of experiments on four commonly used GNN models: SGC~\cite{wu2019simplifying}, ChebNet~\cite{defferrard2016convolutional}, GCN, and APPNP~\cite{gasteiger2018predict}, and repeat each experiment 1000 times by varying the standard deviation of the Gaussian distribution and the random seed.
The results reveal a fairly strong correlation between the high-frequency area and the gradient magnitudes, as shown in Fig.~\ref{fig:fres}, with the high-frequency area increasing gradually, the gradient size generated has an obvious upward trend. Furthermore, 8 of the 12 experiments showed a correlation greater than 0.95. We conducted several experiments in this study, utilizing the following implementation details: 

We initialize a list of 1,000 random seeds ranging from 0 to 100,000,000. With a fixed random seed of 0, we created random synthetic graphs using the Erdos-Renyi~\cite{Erd_s_2013} (edge probability 0.2), Barabasi-Albert (2 edges per new node)~\cite{stamatelatos2022exact}, and Watts-Strogatz~\cite{Song_2014} (4 initial neighbors per node, rewiring probability 0.2) models. 
For each random seed, we followed the same procedure: We selected a bias term from [1, 2, 3, 4, 5] in sequence and then used these different bias terms to generate node features following a Gaussian distribution. With a fixed 200 nodes and 128 feature dimensions, we computed the mean of high-frequency area for each feature dimension based on the node feature matrix and adjacency matrix of the synthetic graph. We then trained a simple 2-layer SGC~\cite{wu2019simplifying}, ChebNet~\cite{defferrard2016convolutional}, GCN, and APPNP~\cite{gasteiger2018predict} for 50 epochs, recording the average gradient magnitude. We calculated the Spearman correlation coefficient using the 1,000 rounds of results obtained. The results presented in Fig.~\ref{fig:fres} were from two rounds under the Erdos-Renyi model, while Table \ref{tab:corre} shows the corresponding results under different graph models and GNNs.
\input{tables/Corre}

\subsection{More Details about Matching Trajectories.} \label{sec:matching}
The experiments of gradient difference analysis in the optimization process are conducted on the Reddit dataset with a condensation ratio of 0.50\%, utilizing four distinct gradient matching methods. It is noteworthy that the hyperparameter $\beta$ for both the CTRL and Cos + Norm methods were uniformly set to 0.3.

\begin{figure}[ht]
    \centering
    \begin{minipage}{0.3\linewidth}
        \centering
        \subfigure[Cosine distance - class 0]
        {\includegraphics[width=\linewidth, angle=0]{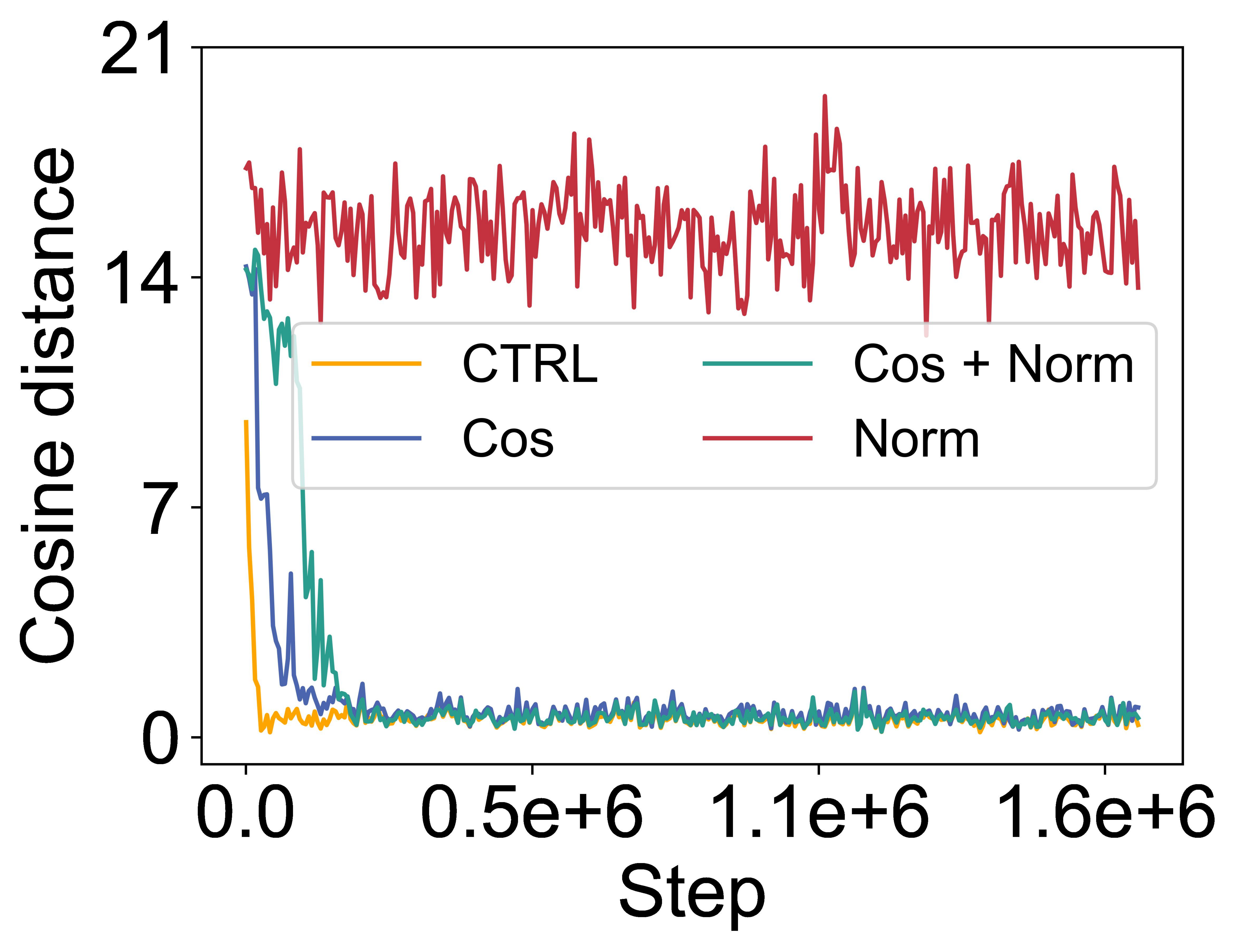}
        \label{fig:cos_0}}
    \end{minipage}%
    \hspace{4.5mm}%
    \begin{minipage}{0.29\linewidth}
        \centering
        \subfigure[Cosine distance - class 28]
        {\includegraphics[width=\linewidth, angle=0]{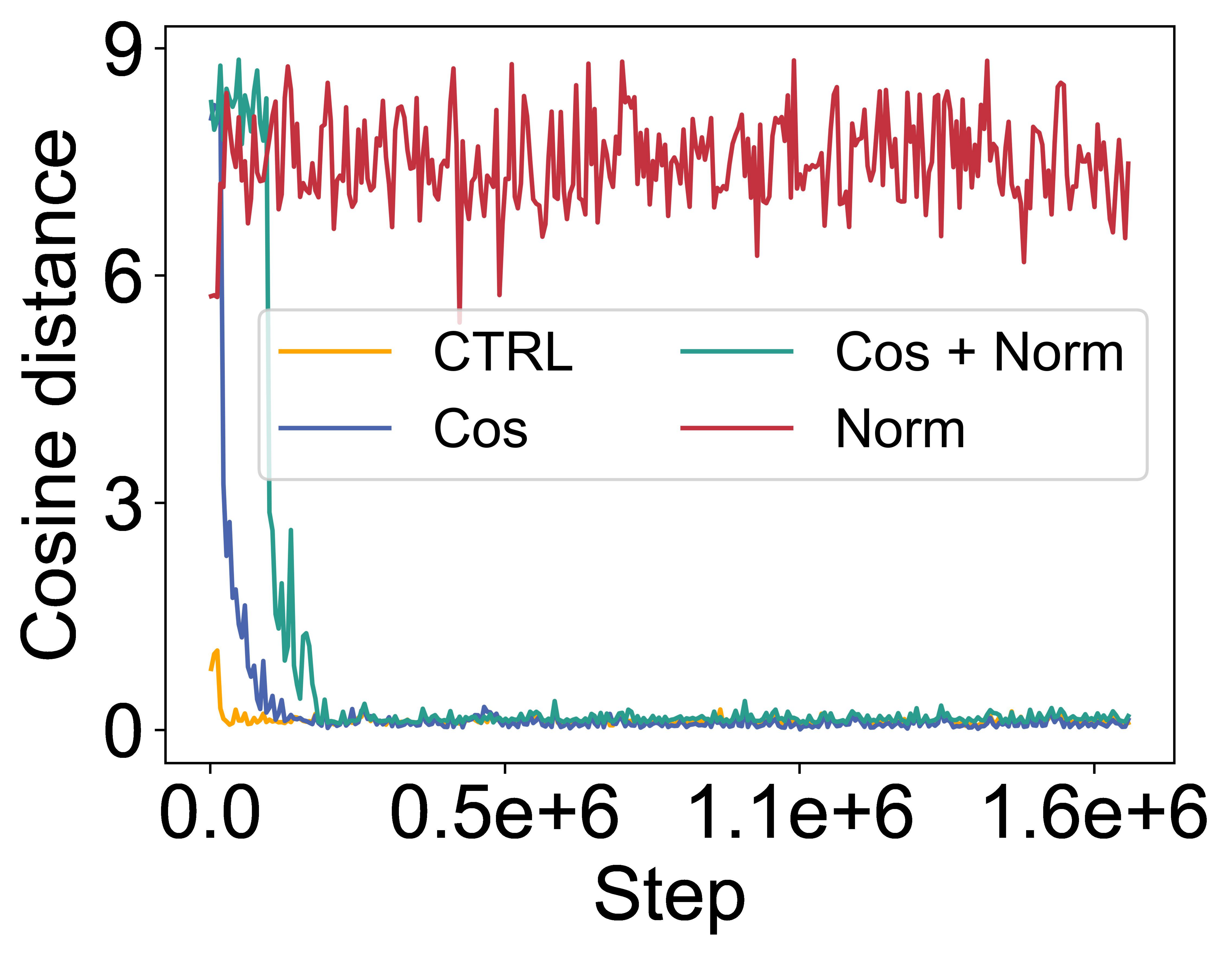}
        \label{fig:cos_28}}
    \end{minipage}%
    \hspace{4.5mm}%
    \begin{minipage}{0.3\linewidth}
        \centering
        \subfigure[Cosine distance - class 32]
        {\includegraphics[width=\linewidth, angle=0]{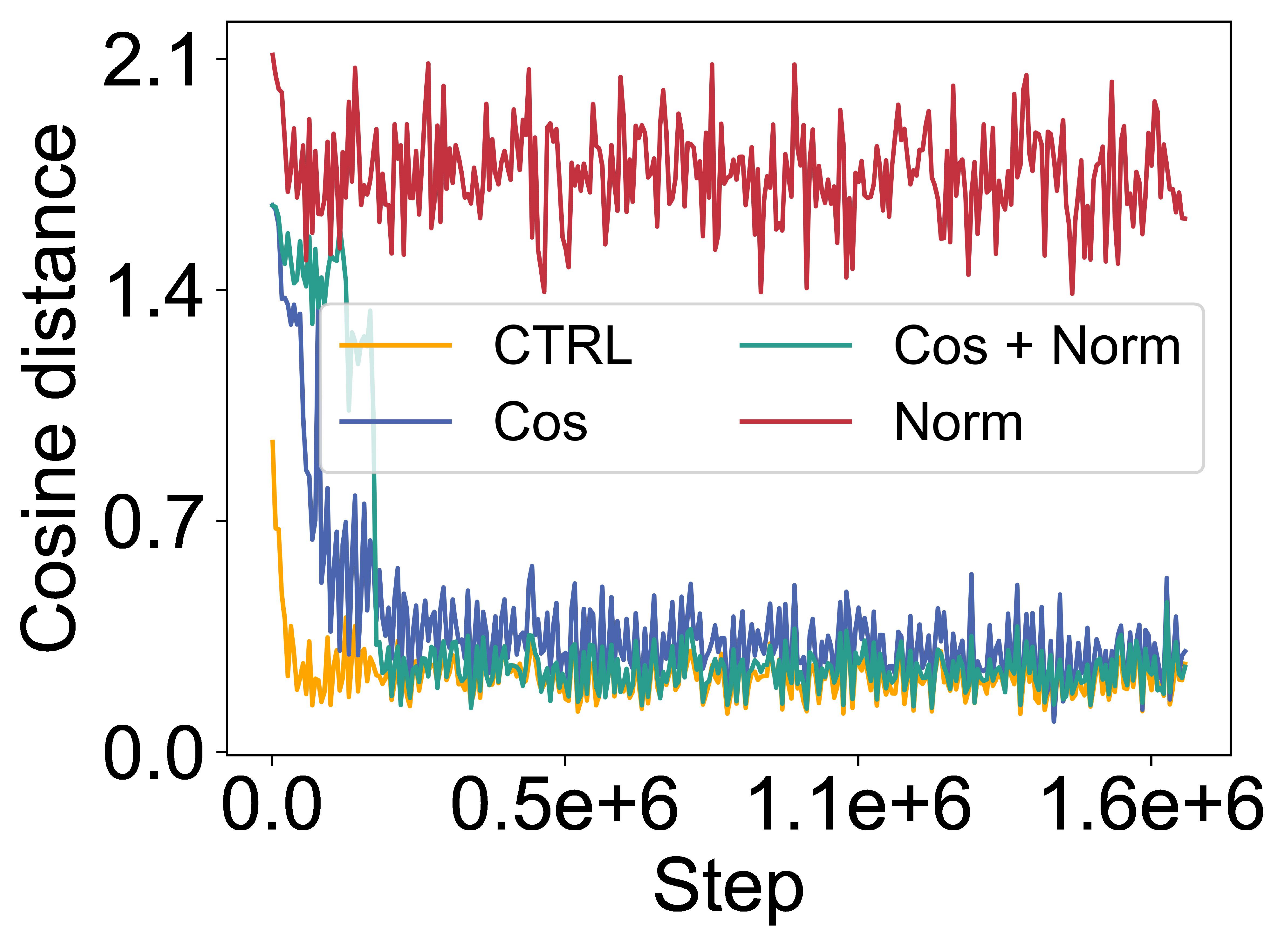}
        \label{fig:cos_32}}
    \end{minipage}

    \vspace{4.5mm} 

    \begin{minipage}{0.3\linewidth}
        \centering
        \subfigure[Magnitude difference - class 0]
        {\includegraphics[width=\linewidth, angle=0]{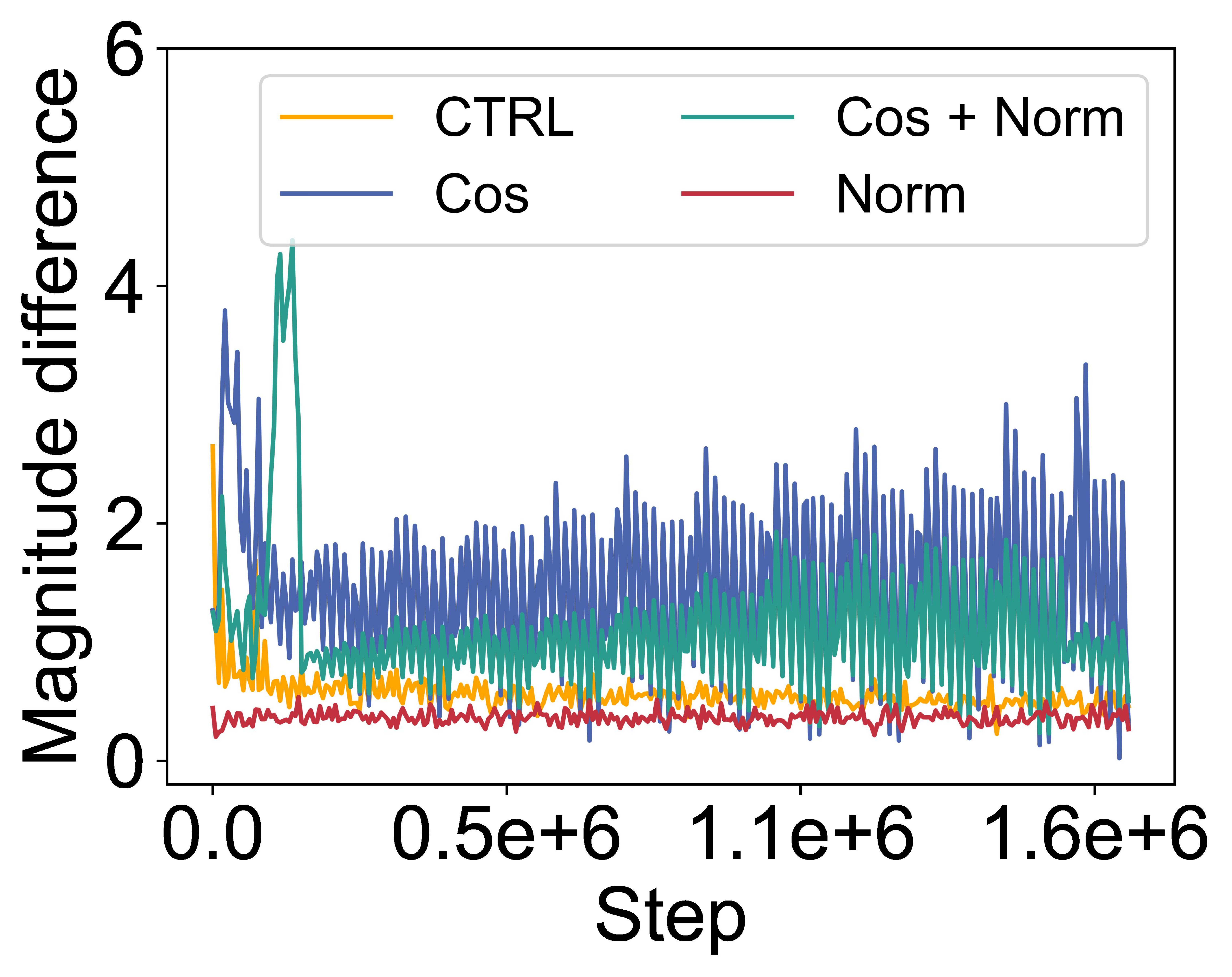}
        \label{fig:amp_0}}
    \end{minipage}%
    \hspace{4.5mm}%
    \begin{minipage}{0.29\linewidth}
        \centering
        \subfigure[Magnitude difference - class 28]
        {\includegraphics[width=\linewidth, angle=0]{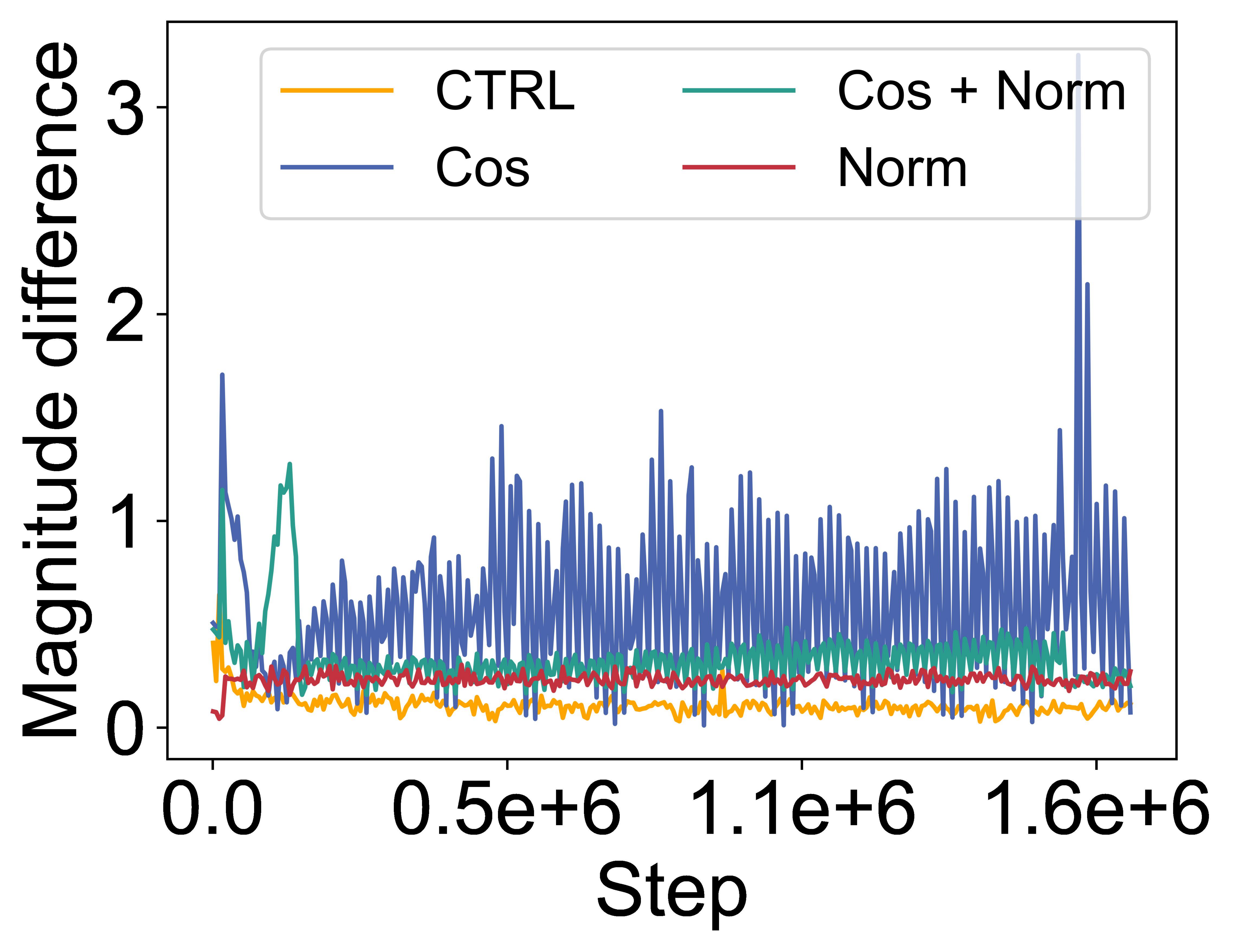}
        \label{fig:amp_28}}
    \end{minipage}%
    \hspace{4.5mm}%
    \begin{minipage}{0.3\linewidth}
        \centering
        \subfigure[Magnitude difference - class 32]
        {\includegraphics[width=\linewidth, angle=0]{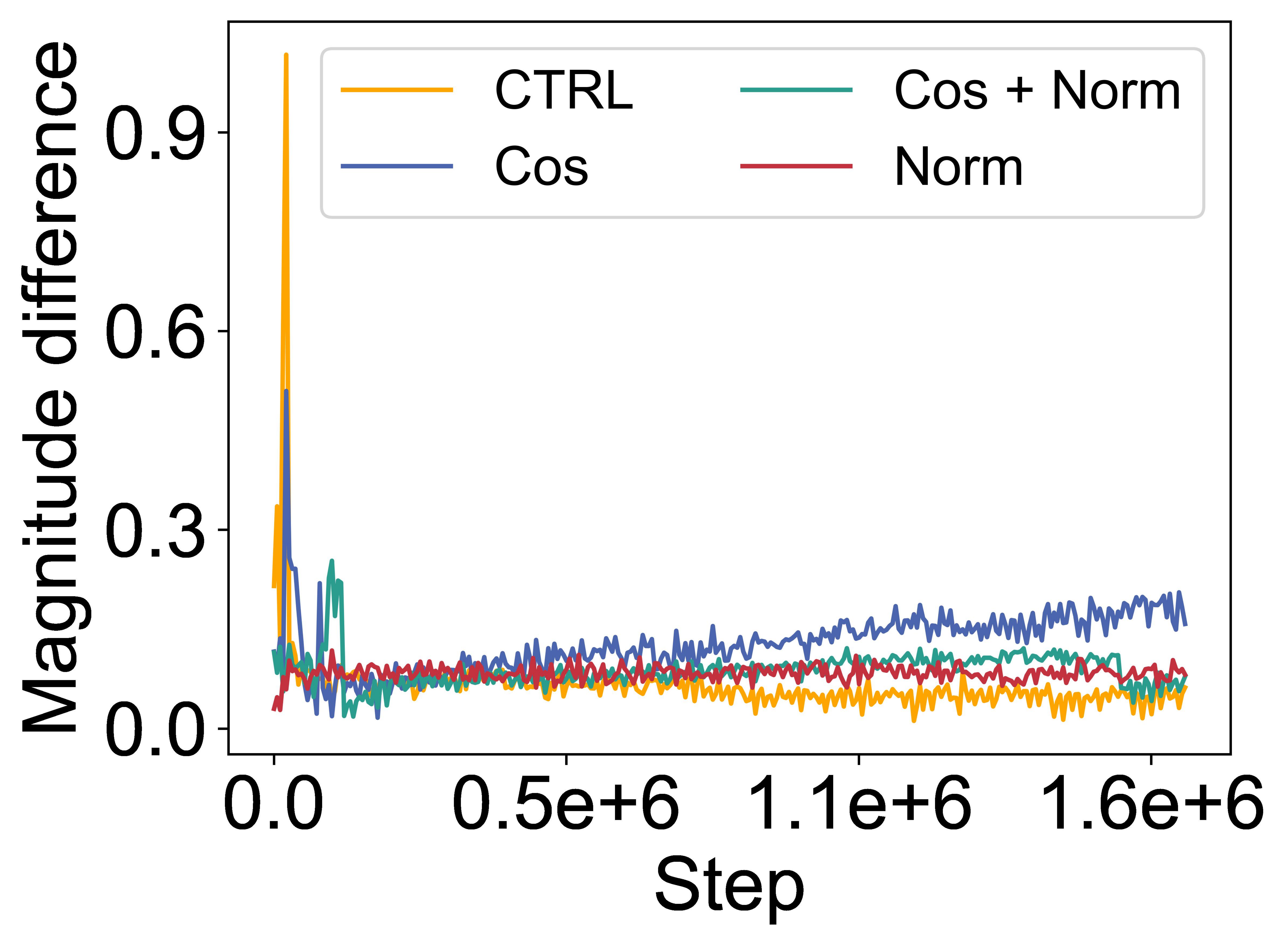}
        \label{fig:amp_32}}
    \end{minipage}

    \hspace{4.5mm}%
    
    \caption{(a)(b)(c)(d)(e) and (f)illustrate the gradient difference during the optimization process of the condensed graphs using CTRL and the basic
    gradient matching methods with three gradient discrepancy metrics, step refers to the times of calculating gradient matching losses. 
    Notably, these results align closely with the conclusions drawn in Sec.~\ref{introduction}.
    }\label{fig:B_Better}
\end{figure}

In the training process, the optimization loss for synthetic data is obtained by training the GNN model on subgraphs composed of nodes from each class. 
To enhance the evaluation of optimization effectiveness, we monitored the variations in gradients during the training process across nodes of distinct classes within both synthetic and original datasets. 
In the previous text, due to space limitations (Reddit has a total of 40 classes), we presented results for class 3 with a relatively high proportion of nodes. 
In this context, to make the experimental results more representative, as depicted in Fig.~\ref{fig:B_Better}, we showcase representative results for classes with a lower proportion of nodes (class 32), classes with a moderate proportion of nodes (class 28), and another class with a higher proportion of nodes (class 28). 
\section{More experiments}
\subsection{Application of CTRL to other Condensation Methods}\label{section: APP}
To further demonstrate the difference between distillation methods in Graph and CV yields, we tried to add CTRL to the CV distillation methods~\cite{jiang2022delving}, which can also achieve improvements, in some cases. However, it is not as obvious as in graph distillation.
\input{tables/cv}
We also conducted experiments on the unstructured graph compression method SFGC, which similarly demonstrated notable improvements.
\input{tables/cross_sfgc}

\subsection{Visualizations}
To better understand the effectiveness of CTRL, we can visualize the condensed graphs in node classification tasks.
It is worth noting that while utilizing the CTRL technique to generate synthetic graphs, it is possible to encounter outliers in contrast with GCOND, especially when working with extensive datasets. 
This phenomenon arises due to the difficulty in learning and filtering out high-frequency signals present in larger datasets, given the low-pass filtering nature of GNNs.
The presence of outliers in the condensed graphs generated by the CTRL method implies that some of the high-frequency signals in the original data are well-preserved in the synthesized graph~\cite{Sandryhaila_2013}. Previous research has demonstrated the importance of both high and low-frequency signals in graph data for effective GNN training~\cite{bo2021beyond,wu2023extracting}. This further substantiates the validity of this observation.
\begin{figure}[ht]
	\centering
	\subfigure[Cora, $r=2.5\%$]
    {\includegraphics[width=0.25\textwidth, angle=0]{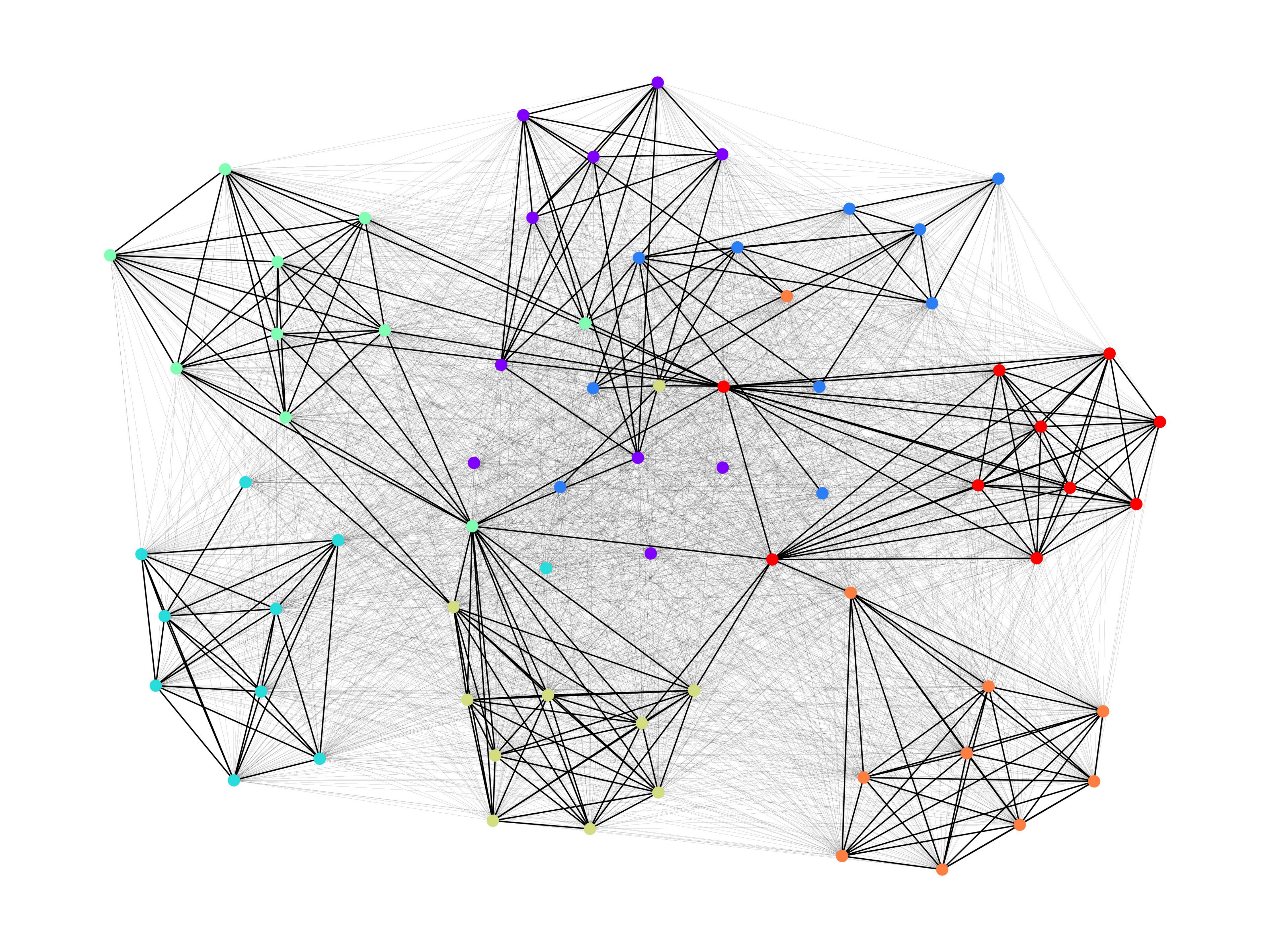}
    \label{fig:vis_cora}}
    \hspace{1mm}
    \subfigure[Citeseer, $r=1.8\%$]
    {\includegraphics[width=0.25\textwidth, angle=0]{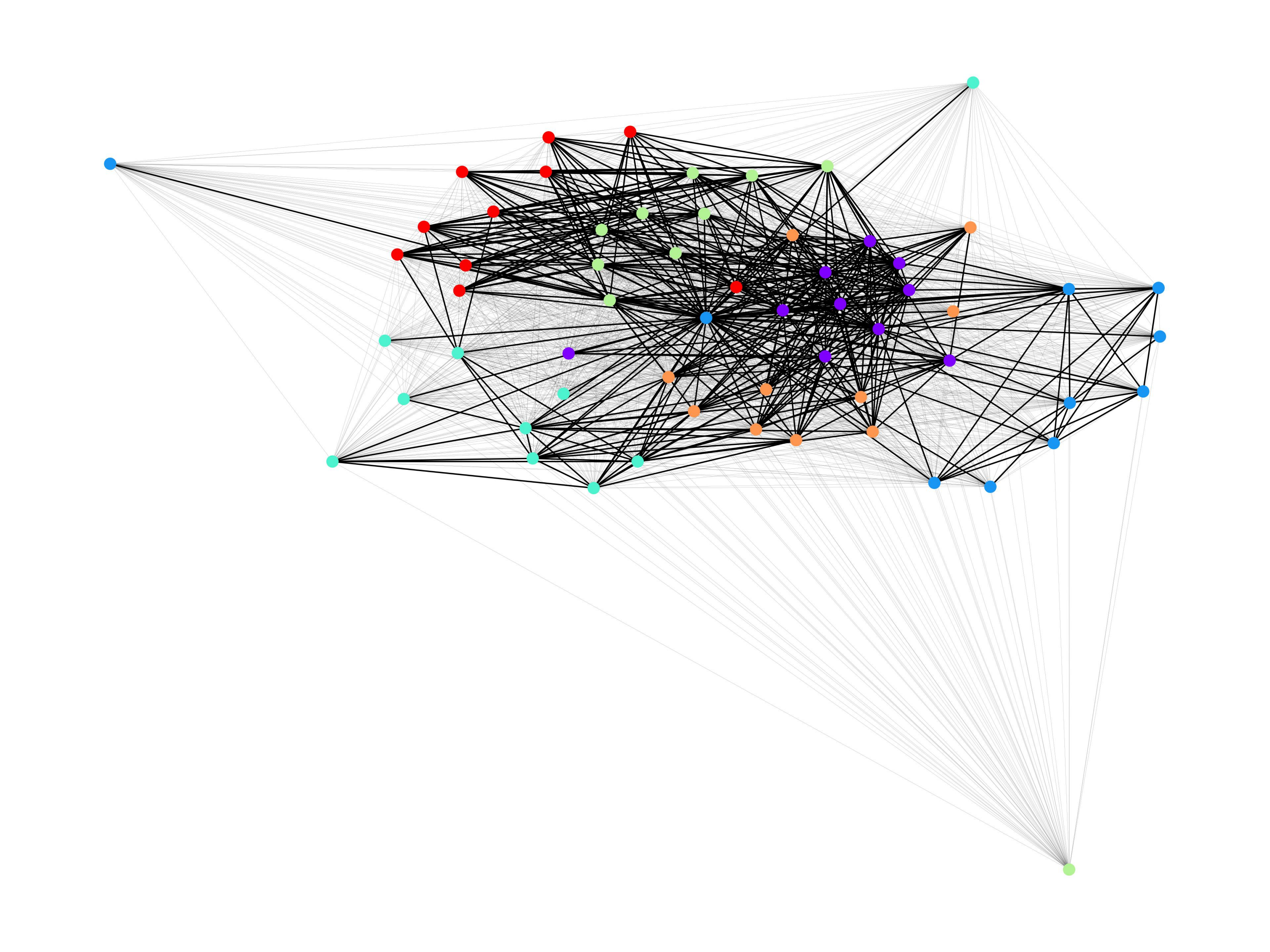}
    \label{fig:vis_citeseer}}
    \hspace{1mm}
    \subfigure[Flickr, $r=0.1\%$]
    {\includegraphics[width=0.25\textwidth, angle=0]{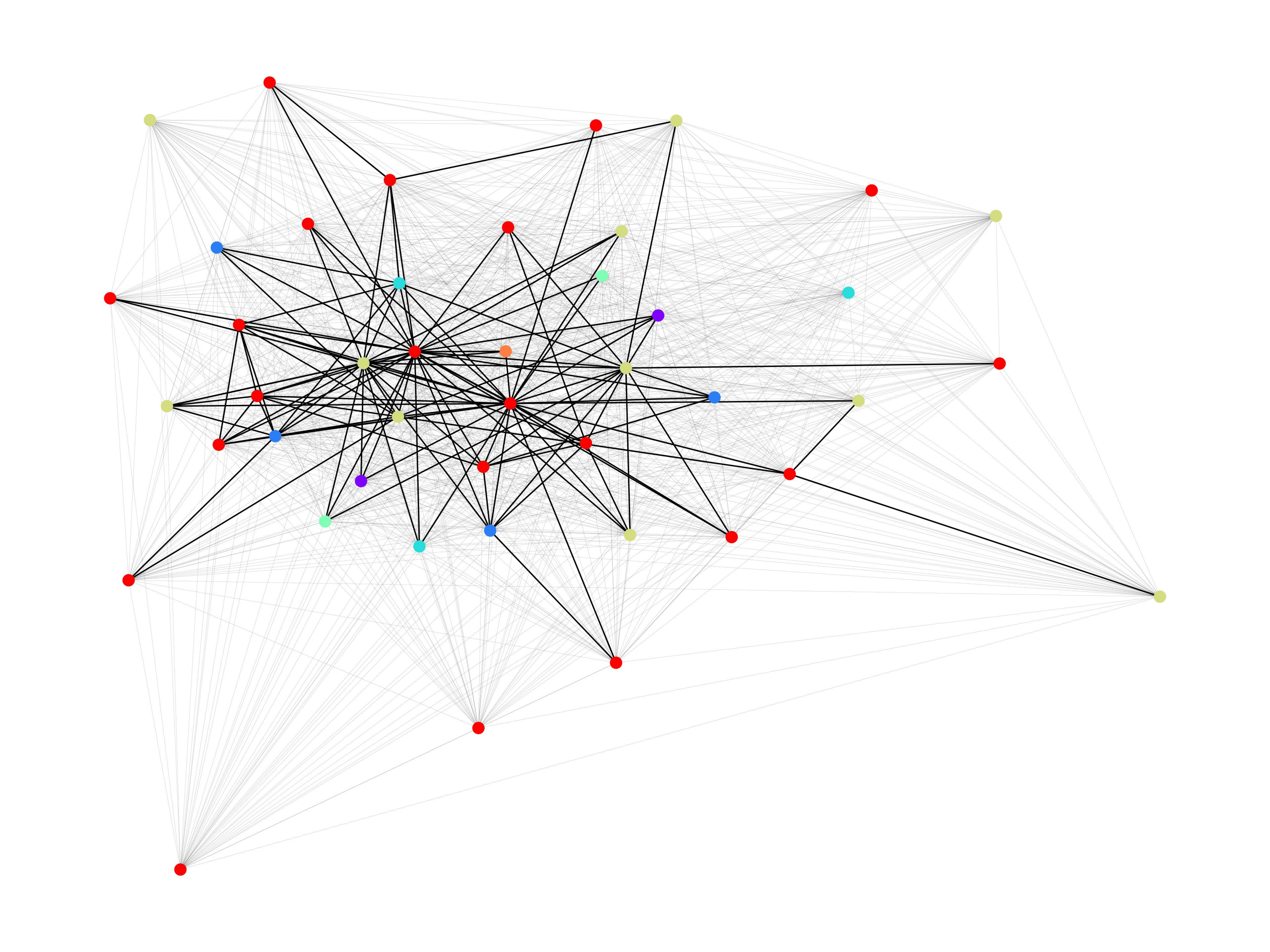} 
    \label{fig:vis_flickr}}
    \hspace{1mm}
    \subfigure[Ogbn-arxiv, $r=0.05\%$]
    {\includegraphics[width=0.45\textwidth, angle=0]{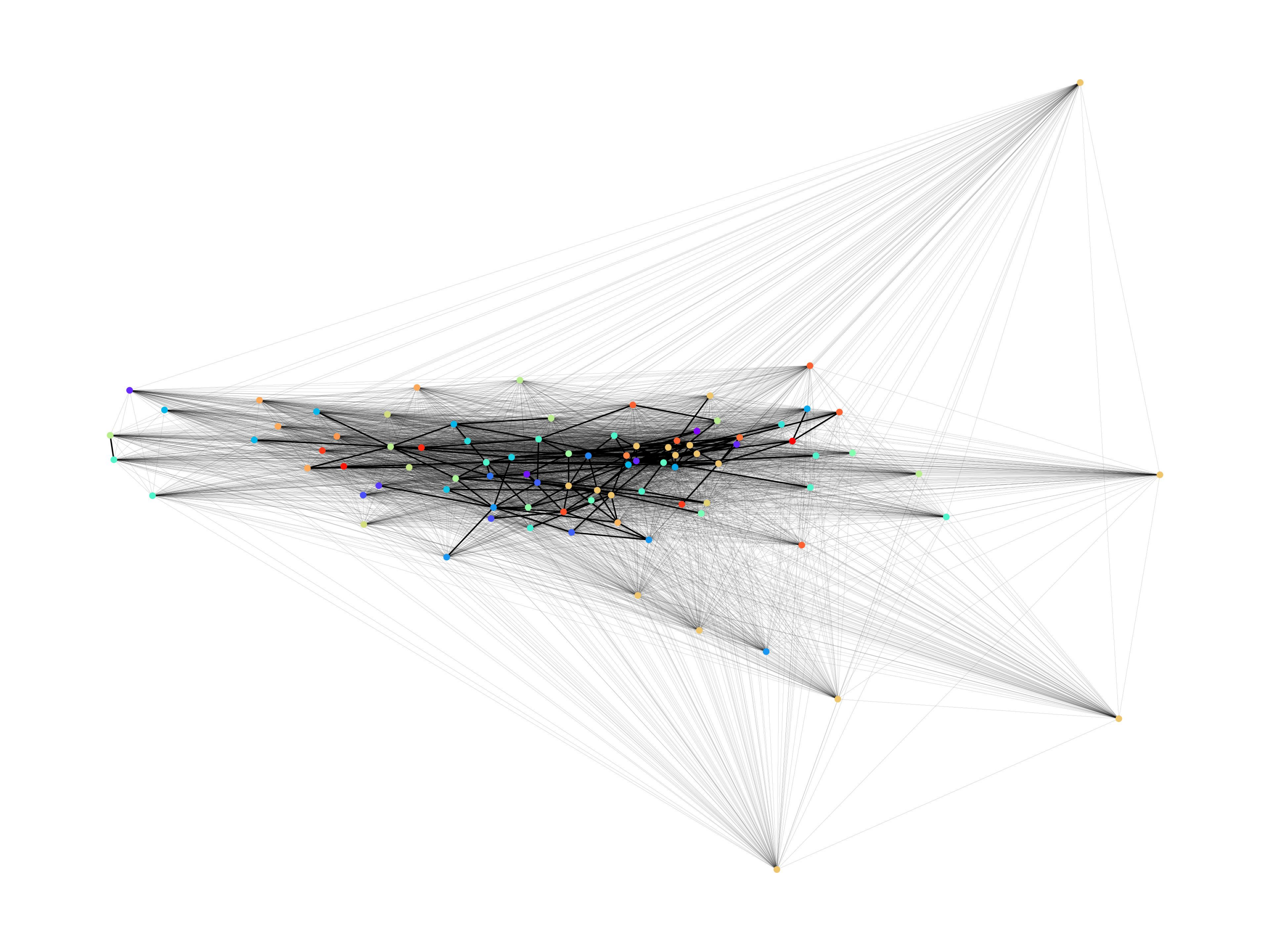} 
    \label{fig:vis_reddit}}
    \hspace{1mm}
    \subfigure[Reddit, $r=0.1\%$]
    {\includegraphics[width=0.45\textwidth, angle=0]{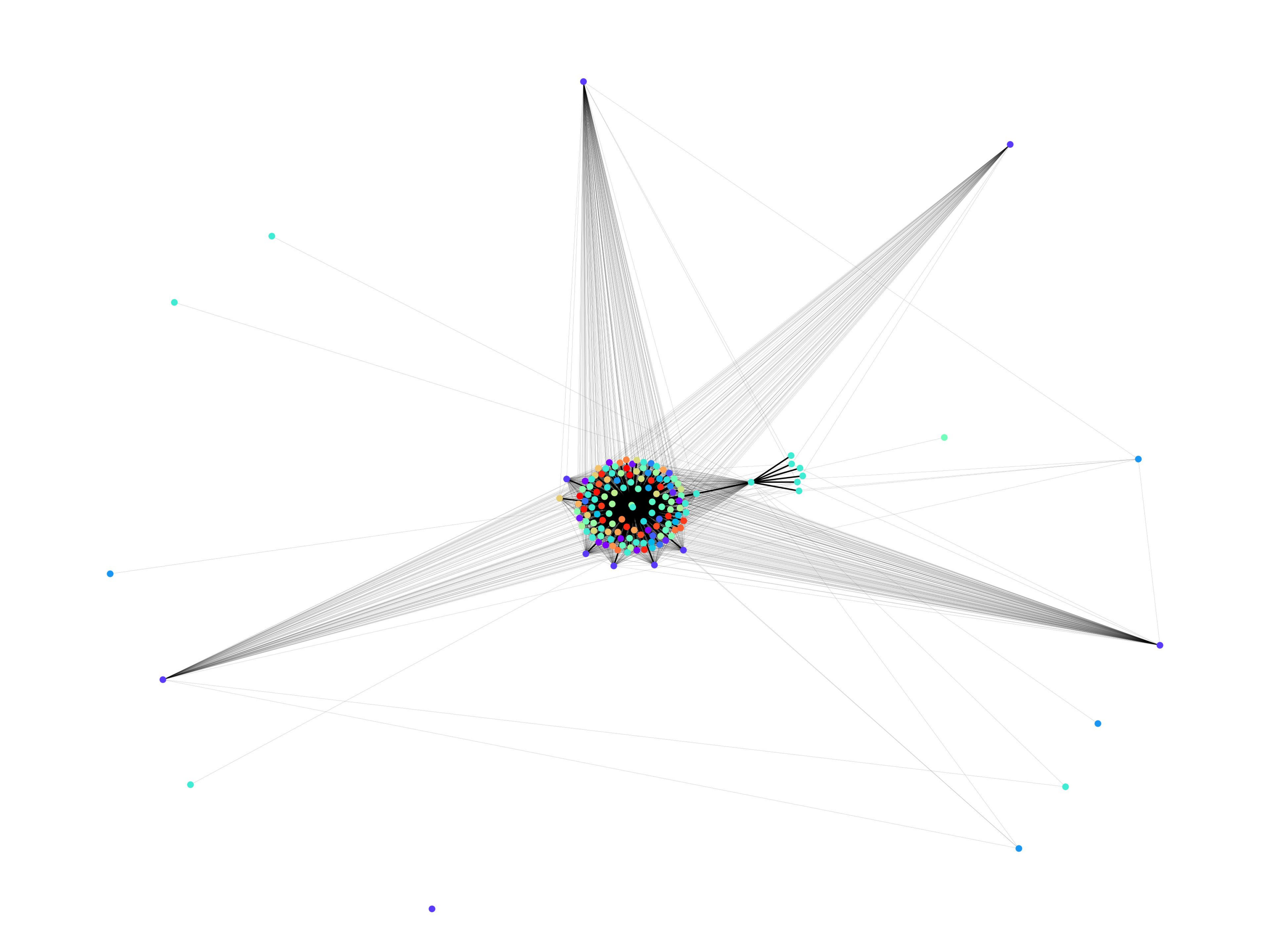} 
    \label{fig:vis_arxiv}}
	\caption{When the size of the original datasets is relatively small(a)(b), the resulting composite dataset tends to exhibit fewer outliers. In contrast, when working with compressed results of large datasets(c)(d)(e), it is more likely to encounter a higher number of outliers}\label{fig:vis_syngraph}
    \vspace{-8.0pt}
\end{figure}

\subsection{Time complexity and Runtime}\label{section: time}
\textbf{Time complexity}. For simplicity, let the number of MLP layers in the adjacency matrix generator be $L$, and all the hidden units are $d$. 
In the forward process, we first calculate the $A'$, which has the complexity of $O(N'^2d^2)$. 
Second, the forward process of GNN on the original graph has a complexity of $O(mLNd^2)$, where m denotes the sampled size per node in training.
Third, the complexity of training on the condensed graph is $O(LN'd)$. 
Then, taking into account additional matching metrics calculations, the complexity of the gradient matching strategy is $O(2|\theta| + |A'| + |X'|)$.

The overall complexity of CTRL can be represented as $O(N'^2d^2) + O(mLNd^2) + O(LN'd) + O(2|\theta| + |A'| + |X'|)$.
Note $N'<N$, we can drop the terms that only involve $N'$ and constants (e.g., the number of $L$ and $m$). 
The final complexity can be simplified as $O(mLNd^2)$,  thus it can be seen that
although the matching process is much finer, the complexity of CTRL still is linear to the number of nodes in the original graph. 

\textbf{Running time}.
We report the running time of the CTRL in two kinds of six groups of classification tasks. 
For node classification tasks, we vary the condensing ratio r in the range of $\left\{0.9\%, 1.8\%, 3.6\%\right\}$ for Citeseer, $\left\{1.3\%, 2.6\%, 5.2\%\right\}$ for Cora, $\left\{0.1\%, 0.5\%, 1.0\%\right\}$ for Ogbn-arxiv, all experiments are conducted five times on one single V100-SXM2 GPU. 
For graph classification tasks, We vary the condensing ratio \textit{r} in the range of $\left\{0.2\%, 1.7\%, 8.3\%\right\}$ for Ogbg-molbace, $\left\{0.1\%, 1.2\%, 6.1\%\right\}$ for Ogbg-molbbbp, $\left\{0.1\%, 0.5\%, 1.0\%\right\}$ for Ogbg-molhiv, all experiments are repeated 5 times on one single A100-SXM4 GPU. 

We also conduct experiments with GCond and Docscond using the same settings, respectively. As shown in Table \ref{tab:running time of NC} and \ref{tab:running time of GC}, our approach achieved a similar running time to GCond and Doscond, note that the comparison procedure here improves the cosine distance calculation of GCond, otherwise, our method would be faster on small datasets.

\input{tables/time}


\end{document}

%% file: 1.Introduction.tex
\section{Introduction} \label{introduction}
Graph neural networks (GNNs) have demonstrated superior performances in various graph analysis tasks, including node classification~\cite{li2019learning,xiao2022graph}, link prediction~\cite{chen2022gc,rossi2022explaining}, and graph generation~\cite{van2022declarative,vignac2022digress}. However, the training GNNs overheads on large-scale graphs comes at a high cost~\cite{wang2022adagcl,polisetty2023gsplit}. One of the most straightforward ideas is to reduce the redundancy sub-counterpart like graph sparsification~\cite{li2022graph,yu2022principle}, which
directly removes trivial edges or nodes, while graph coarsening~\cite{chen2022graph,kammer2022space,wang2022searching} shrink the graph size by merging similar nodes. 
Yet the rigid operation graph sparsification and coarsening conduct heavily rely on heuristic techniques~\cite{cai2021graph}, which leads to the disregard for local details and limitation in compress ratio~\cite{jin2022graph}. 

The recently ever-developed graph condensation effectively opens a potential path for the upcoming large-scale graph training and storage, by generating a synthetic sub-counterpart to mimic the original graph ~\cite{jin2022graph,jin2022condensing,zheng2023structure}. As one of the well-established graph condensation frameworks, GCond takes the \underline{first step} to adopt the condensation techniques in the computer vision realm~\cite{zhao2020dataset} and proposes a graph condensation concept based on gradient-matching. 
Concretely, GCond optimizes the synthetic graph by minimizing the gradient discrepancies of GNNs trained on the original graph and the synthetic graph.
As a result, GCond is capable of condensing the graph under an extremely low condensation ratio without introducing significant performance degradation.

\begin{figure*}[ht]
        \centering
        \subfigure[Cosine distance]
        {\includegraphics[width=0.3\linewidth, angle=0]{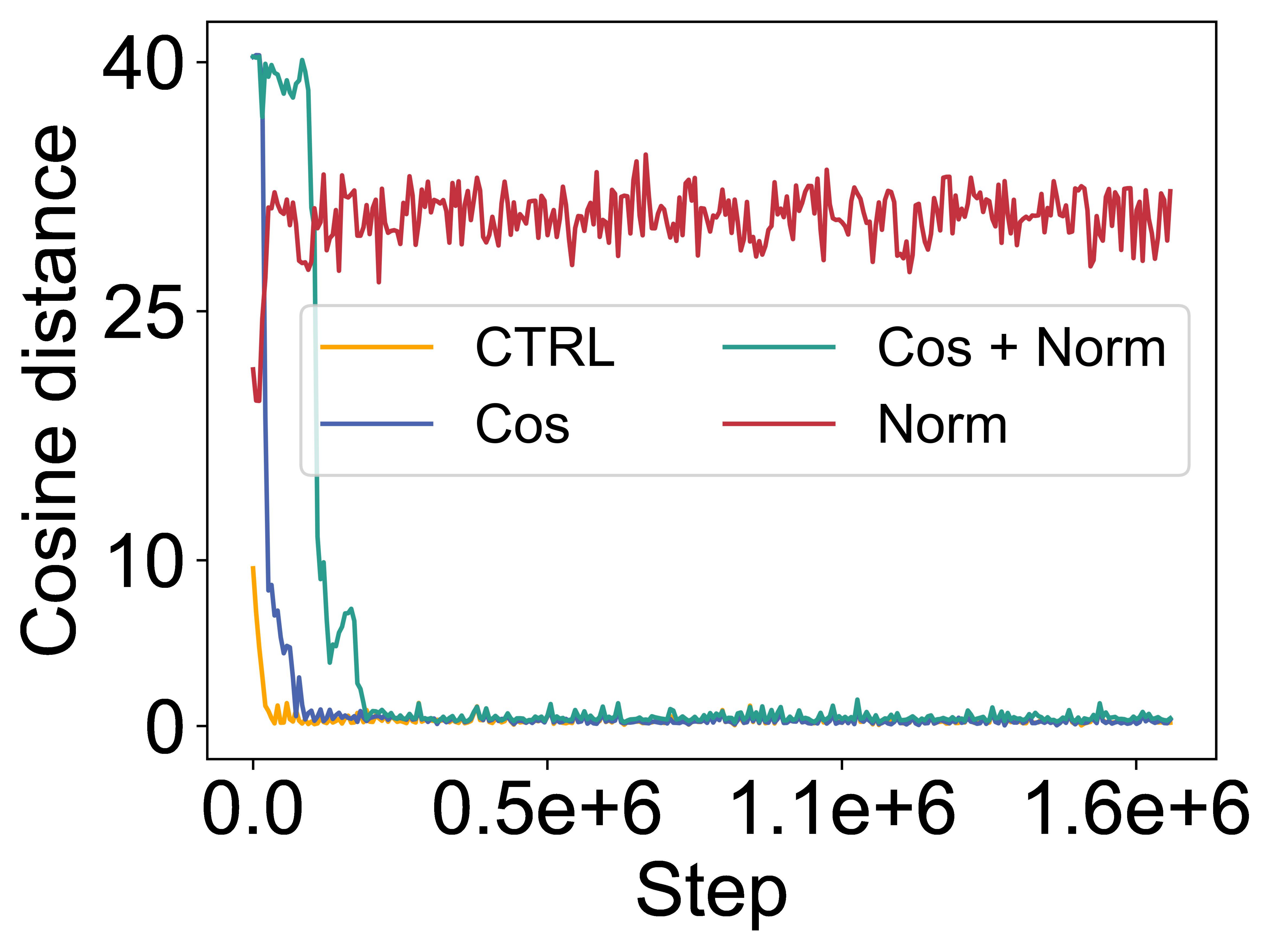}
        \label{fig:cos}}
         \hspace{3.5mm}
        \subfigure[Magnitude difference]
        {\includegraphics[width=0.3\linewidth, angle=0]{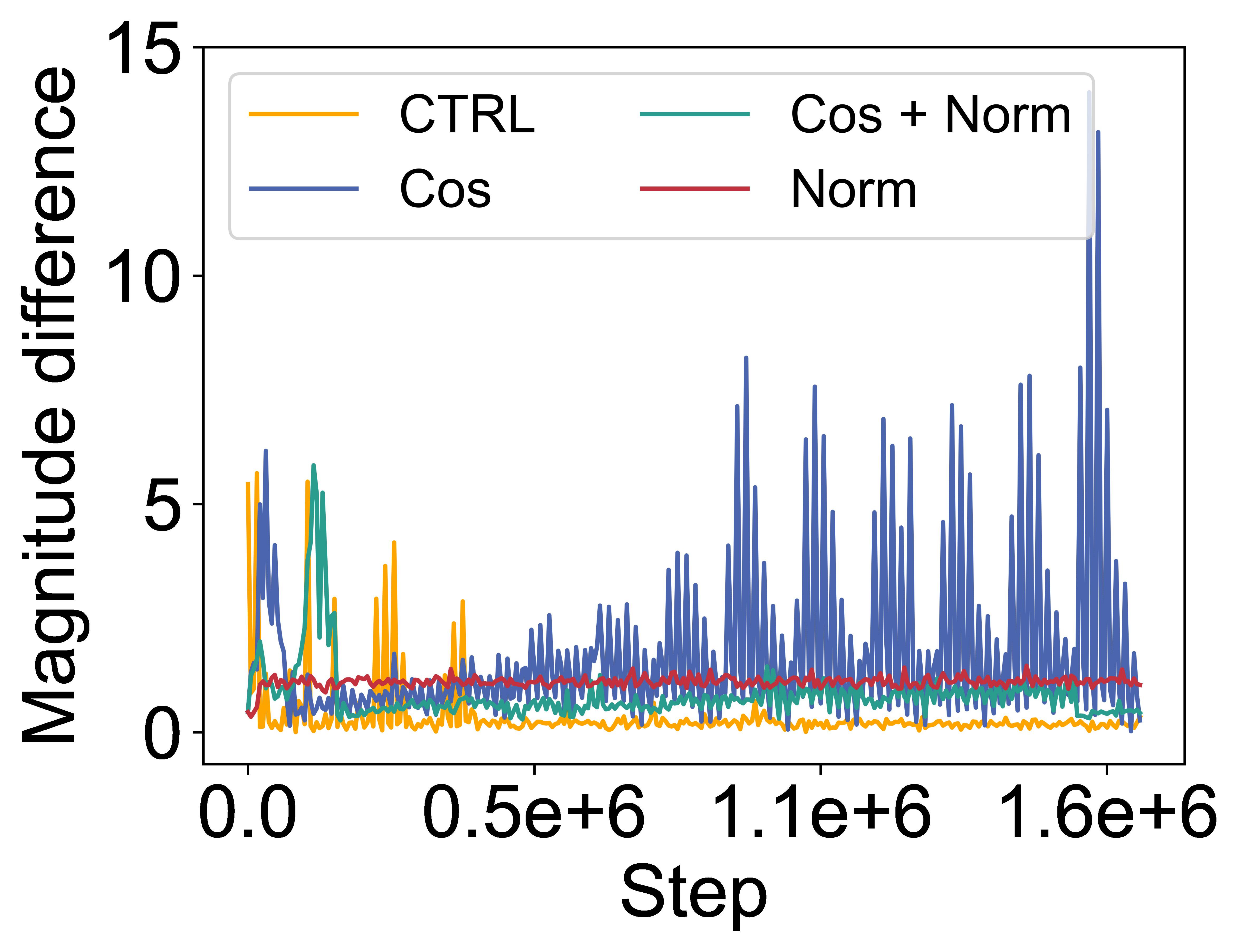}
        \label{fig:amp}}
        \hspace{3.5mm}
        \subfigure[Test accuracy]
        {\includegraphics[width=0.295\linewidth, angle=0]{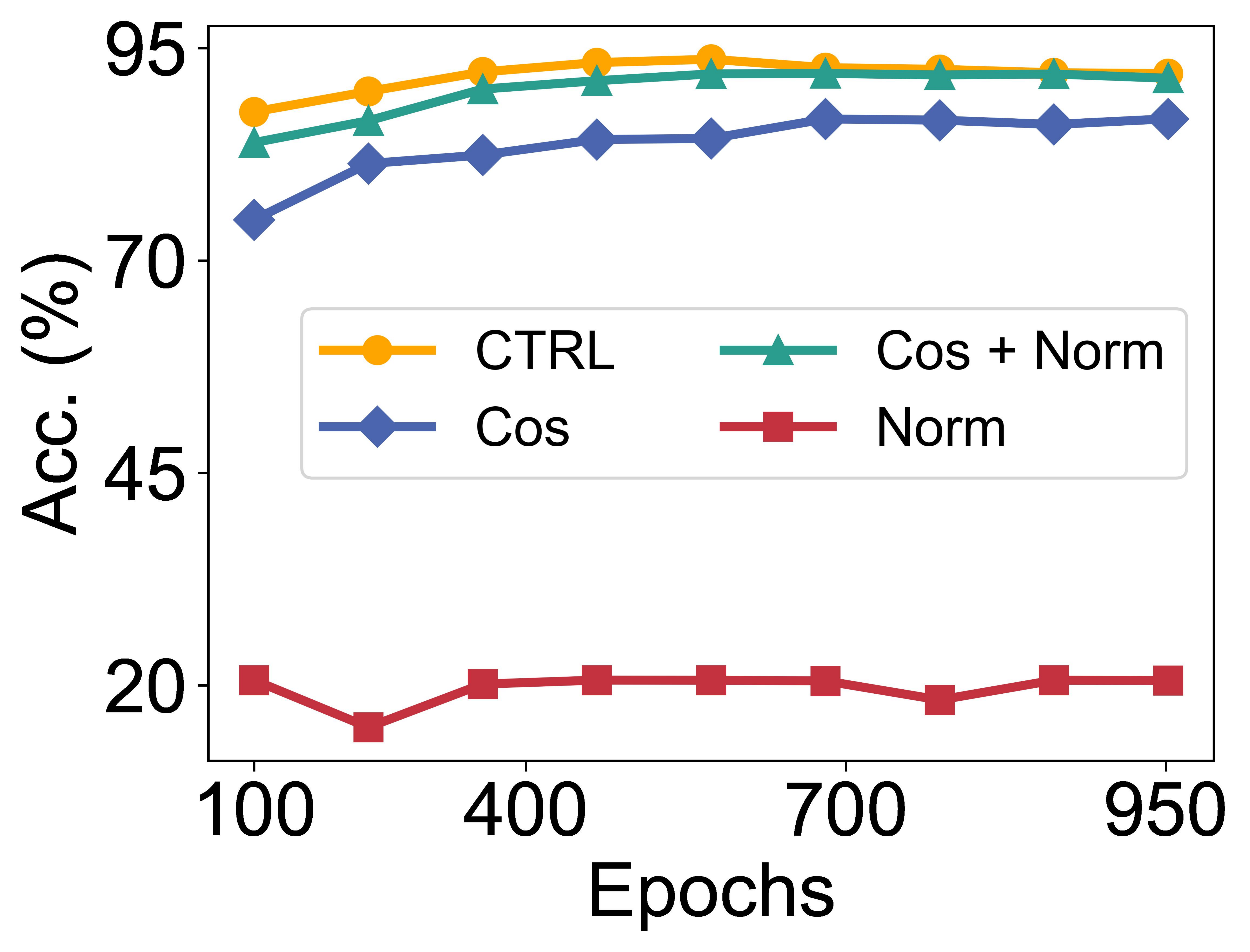}
        \label{fig:acc}}
        \vspace{-0.3cm}
    \caption{
    (a) and (b) illustrate the gradient difference during the optimization process of the condensed graphs using CTRL and the basic gradient matching methods with three gradient discrepancy metrics, step refers to the times of calculating gradient matching losses. 
    (c) shows the performance of the condensed graphs during the optimization process for various methods.
    }\label{fig_1}
    \vspace{-3mm} 
\end{figure*}
Building on this milestone, a deluge of endeavors have adopted a similar paradigm for graph condensation~\cite{jin2022condensing,app13169166,gao2023graph}. 
However, gradient-matching primarily focuses on minimizing the cosine distance between gradients (termed Cos), leading to a discrepancy in magnitude between the gradients produced by the synthetic graph and those by the original graph. 
As shown in Fig.\ref{fig_1}(a)$\sim$(c)\footnote{we provide detailed experimental settings in Appendix~\ref{sec:matching}}, 
using either cosine distance or Euclidean distance as the sole metric for gradient matching leads to optimization collapse due to the inability to address gradient magnitude differences and the neglect of gradient angle differences. Based on the aforementioned methods, matching errors in this aspect have inadvertently been left unresolved (\textbf{Drawback} \textcolor{black}{\ding{182}}). Worse still, this error can lead to a divergence in starting points between the condensation and evaluation phases, consequently further impacting the performance of the synthetic graph in the form of so-called accumulated errors~\cite{du2023minimizing}.



To address the above issue, a natural solution is to directly match the Euclidean distance between gradients (termed Norm) \cite{zheng2023structure,jin2022condensing}. However, as illustrated in Fig.~\ref{fig_1}(a)$\sim$(c), Norm performs poorly in graph data condensation tasks for node classification. One plausible reason is that, initially, the Euclidean distance between gradients is too small to guide the optimization of the synthetic graph effectively~\cite{jiang2022delving}. Additionally, the objective of minimizing the Euclidean distance overly emphasizes fitting the gradient magnitude while neglecting optimization of gradient orientations.
To further reduce matching errors, we enhance existing gradient matching methods by combining cosine and Euclidean distances (termed Cos + Norm). As depicted in Fig.~\ref{fig_1}(a)$\sim$(c), this approach effectively reduces the final matching errors. However, this way introduces new challenges,
tackle discrepancies in gradient orientations
which require additional steps to address the optimization
(\textbf{Drawback} \textcolor{black}{\ding{183}}). This may be attributed to the fact that simple random sampling initialization may result in synthetic data initially exhibiting similar features, as shown in Fig.~\ref{fig_1}(d), thereby introducing additional optimization challenges~\cite{liu2023dream,schuetzke2022universal}.

\begin{wrapfigure}{r}{0.3\textwidth}
    \centering
    \includegraphics[width=0.27\textwidth]{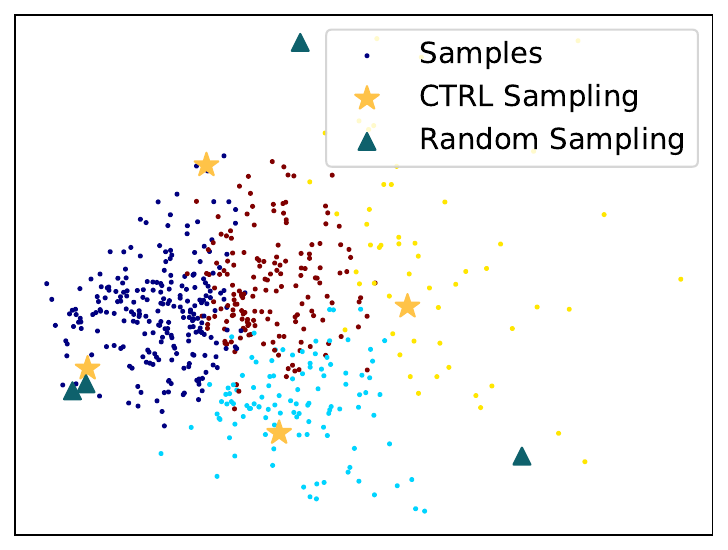}
    \caption{Sampling example}
    \label{fig:example}
\end{wrapfigure}


With this in mind, we introduce a novel graph condensation method called \textbf{C}raf\textbf{T}ing \textbf{R}ationa\textbf{L} gradient matching (CTRL). 
we consider both the direction and magnitude of gradients during gradient matching to minimize matching errors during the condensation phase, effectively reducing the impact of accumulated errors on the performance of synthetic graphs (\textbf{Benefit} \textcolor{black}{\ding{182}}).
Furthermore, we cluster each class of the original data into sub-clusters and sample from each sub-cluster as initialization values, as shown in Fig.~\ref{fig:example}, this approach can obtain initial synthetic data with even feature distribution, further assisting in reducing matching errors and accumulated errors (\textbf{Benefit} \textcolor{black}{\ding{183}}). 

Compared to previous methods, CTRL allows the systematic matching of gradients and obtains initial synthetic data with even feature distribution (Fig.~\ref{fig:example}), 
effectively reducing the accumulated errors caused by misalignment in gradient magnitude and gradient orientations. 
Moreover, by matching the gradient magnitude, CTRL can effectively capture the frequency distribution of signals, our studies reveal that the graph condensed by CTRL accurately mimics the frequency distribution of the original graph.

We further assess the performance of CTRL through comprehensive experiments, to evaluate its generalization capabilities, we extend its application to node classification and graph classification tasks. 
Additionally, our synthetic graphs exhibit outstanding performance in various downstream tasks, including \underline{cross-architecture} applications and \underline{neural architecture search.}

With greater specificity, CTRL effectively reduces the Euclidean distance between gradients to just 3.7\% of their original values while maintaining the cosine distance. Remarkably, we achieve lossless results on prominent datasets, including Citeseer, Cora, and Flickr for node classification, as well as the Ogbg-molbace and Ogbg-molbbbp datasets for graph classification. Notably, our approach yields new state-of-the-art results on the Ogbg-molbace and Ogbn-arxiv datasets, demonstrating improvements of 6.2\% and 6.3\%, respectively. Our main contributions are summarized as:

\ding{224} Based on the analysis of existing graph condensation area, which dominated by gradient matching, we introduce CTRL, a simple and highly generalizable method for graph dataset condensation through finer-grained gradient matching.

\ding{224} In CTRL, the optimization trajectory of our synthetic data closely approximates that of real data through a weighted combination of cosine distance and Euclidean distance and effectively captures the feature distribution of real data.

\ding{224} We conduct experimental evaluations across 18 node classification tasks and 18 graph classification tasks. The results highlight that our method achieves state-of-the-art performances in 34 cases of experiments on 12 datasets with lossless performances on 5 datasets.

%% file: 3.Method.tex
\section{Effective Gradient Matching via CTRL}
\subsection{Graph Condensation via Gradient Matching}
The objective of the graph condensation via gradient matching framework is to extract latent information from a large graph dataset $\mathcal{T}=\{\mathbf{A},\mathbf{X},\mathbf{Y}\}$ to synthesize a smaller dataset $\mathcal{S}=\{\mathbf{A'},\mathbf{X'},\mathbf{Y'}\}$, such that a model trained on $\mathcal{S}$ can achieve comparable results to one trained on $\mathcal{T}$, where 
$\mathbf{A}\in\mathbb{R}^{N\times N}$ denotes the adjacency  matrix, $\mathbf{X}\in\mathbb{R}^{N\times d}$ is the feature, and $\mathbf{Y}\in\mathbb{R}^{N\times 1}$ represents the labels, the first dimension of $\mathbf{A'},\mathbf{X'}$, and $\mathbf{Y'}$ are $\mathbf{N'}$. 
To summarize, the process essentially revolves around matching the gradients generated during the training of GNNs on both datasets. 
To achieve this alignment, the following steps are undertaken: Fitstly, both on the large original dataset $\mathcal{T}$ and the small synthetic dataset $\mathcal{S}$, we train a GNN model parameterized with 
$\mathrm{\boldsymbol{\theta}}$, denotes as $f_{\boldsymbol{\theta}}$ and computes the parameter gradients at each layer for this model. We then use the gradients from the synthetic dataset to update the GNN model.
The optimization goal is to minimize the distance $\mathcal{D}$ between the gradients at each layer, essentially, this aligns the training trajectories of the models over time, making them converge.

The above process can be written as follows:
\begin{gather}
\operatorname*{min}_{\mathcal{S}}\mathrm{E}_{\boldsymbol{\theta}_0\sim P_{\theta_0}}
\sum_{t=0}^{T-1}\Big[
\mathcal{D}\Big(
\nabla_{\boldsymbol{\theta},t}^\mathcal{S},\nabla_{\boldsymbol{\theta},t}^\mathcal{T}\Big)\Big],
\\
\mathrm{s.t.}\quad 
\nabla_{\boldsymbol{\theta},t}^\mathcal{S} = \frac{\partial \mathcal{L}(f_{\boldsymbol{\theta_t}}(\mathcal{S}), \mathbf{Y}^{\prime})}{\partial \theta}, 
\nabla_{\boldsymbol{\theta},t}^\mathcal{T} = \frac{\partial \mathcal{L}(f_{\boldsymbol{\theta_t}}(\mathcal{T}), \mathbf{Y})}{\partial \theta}.\nonumber 
\end{gather}
Where $T$ is the number of steps of the whole training trajectory, and 
$\boldsymbol{\theta}_t$ denote the model parameters at time step $t$. The distance $\mathcal{D}$ is further defined as the sum of the distance between two gradients at each layer.
Given two gradient $\mathbf{G}^{\mathcal{S}}\in\mathbb{R}^{d_1\times d_2}$ and $\mathbf{G}^{\mathcal{T}}\in\mathbb{R}^{d_1\times d_2}$ at a specific layer, we defined the distance $\mathrm{dis}(\cdot,\cdot)$ as follows:
\begin{gather}
\mathrm{dis}(\mathbf{G}^{\mathcal{S}},\mathbf{G}^{\mathcal{T}})
=\sum_{i=1}^{d_2}g(\mathbf{G}_i^{\mathcal{S}},\mathbf{G}_i^{\mathcal{T}}).
\end{gather}
Where $\mathbf{G}_i^{\mathcal{S}}$, $\mathbf{G}_i^{\mathcal{T}}$ are the $i$-th column vectors of the gradient matrices, and the metric $g$ serves as an indicator for quantifying the difference between two gradient vectors. 

\subsection{Graph Condensation via Crafting Rational Trajectory}\label{sec:CTRL}

\textbf{Refined matching criterion.}
Previous methods~\cite{jin2022condensing,zheng2023structure} often solely rely on either cosine or Euclidean distances between gradients to guide synthetic graph optimization. However, this approach typically neglects both gradient magnitudes and orientations.

To enhance gradient matching guidance, we combine cosine similarity and Euclidean distance into a linear combination. Initially, the optimization prioritizes aligning gradient directions due to the typically larger cosine distance between gradients in the initial phase. As the cosine distance diminishes, the focus of the Euclidean distance shifts primarily to measuring gradient magnitudes, allowing subsequent optimization processes to concentrate effectively on aligning the magnitudes of the gradients, ensuring they are well-aligned while avoiding interference with the alignment of gradient orientations. 
Additionally, We assign different weights to balance the significance of these two metrics through $\beta$. 
Formally, we determine the refined matching criterion $g_{\rm{ctrl}}$ as:
\begin{gather}\footnotesize
\label{eq_ctrl}
g_{\rm{ctrl}}(\mathbf{G}_i^{\mathcal{S}},\mathbf{G}_i^{\mathcal{T}}) = 
\beta g_1 + (1-\beta)g_2,
\\\nonumber
\mathrm{s.t.}\;\;\;g_1 = \|\mathbf{G_i}^{\mathcal{S}}-\mathbf{G_i}^{\mathcal{T}}\|, \;\;
g_2 = 1-\frac{\mathbf{G_i}^{\mathcal{S}}\cdot\mathbf{G_i}^{\mathcal{T}}}{\|\mathbf{G_i}^{\mathcal{S}}\|\|\mathbf{G_i}^{\mathcal{T}}\|}.
\end{gather}
\textbf{Initialization of the synthetic graph.}
Traditional graph condensation methods suffer from limitations in effectively capturing the feature distribution of real data. 
As shown in Fig.~\ref{fig_1}(d), the random sampling initialization they employed brings uneven feature distribution, which leads to suboptimal initial optimization points in the condensation process, subsequently impairing the condensation process~\cite{liu2023dream}.
To tackle this problem, CTRL initially employs a clustering algorithm (termed $Agg.$), dividing each class of the original graph into $M$ sub-clusters $S_{c}=\{S_{c,0},...,S_{c,m},...,S_{c,M}\}$, where $M$ is the number of nodes per class in the synthetic graph. Subsequently, a node feature is sampled from each cluster according to distribution $P_{\phi}$ to serve as the initial values for the corresponding node feature in the synthetic graph ${x'}_{c,m}$.
We formulate this initialization process as follows:
\begin{gather}
\label{eq_init}
{x'}_{c,m} \xleftarrow{\text{Init}} P_{\phi}(S_{c,m}),\;\;
\mathrm{s.t.}\;\;S_{c} = Agg.(\mathbf{X}_{c},M).
\end{gather}
After extensive experimental comparisons, we ultimately adopt the K-Means~\cite{hartigan1979algorithm,arthur2007k} and a uniform distribution as the methods for $Agg.$ and $P_{\phi}$ in our approach.
This method ensures the initial synthetic graph's feature distribution is closer to the original graph's, while it only requires clustering data from a specific class at a time, which is exceptionally cost-effective in terms of computation.

Next, We provide the theoretical analysis to demonstrate the rationality of the proposed CTRL.


\begin{wrapfigure}{l}{0.5\textwidth}
        \centering
        {\includegraphics[width=0.46\textwidth, angle=0]{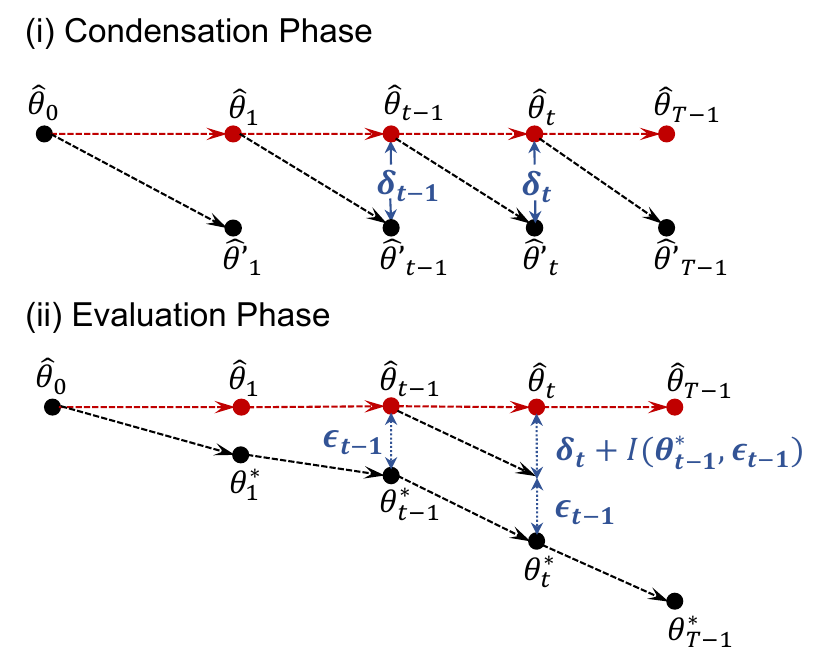}
        }
    \caption{Illustration of gradien matching.}\label{fig:method_ctrl}
    \vspace{-1mm} 
\end{wrapfigure}

\textbf{Theoretical understanding.}
According to \cite{du2023minimizing} and \cite{zhangnavigating}, the training trajectories of GNNs using original and synthetic graphs can be divided into \( T \) stages, donate as \( (\theta^*_{0}, \ldots, \theta^*_{t}, \ldots, \theta^*_{T-1}) \) and \( (\hat{\theta}_{0}, \ldots, \hat{\theta}_{t}, \ldots, \hat{\theta}_{T-1}) \), respectively, where \( \theta^*_{0}\) represents the initial parameters of all GNNs, i.e. \( \theta^*_{0} = \hat{\theta}_{0} \). Then for any stage \( t \), we define the following:

\textbf{Definition 2.1} 
\textit{Accumulated error}, which represents the parameter difference between models trained with synthetic and original data during the evaluation stage:
\begin{equation}
 \epsilon_t = \hat{\theta}_{t} - \theta^{*}_{t}. 
\end{equation}

This error originates from the challenge of completely eradicating gradient discrepancies 
during the condensation phase. 
As a result, the starting points of GNNs' training exhibit differences between the condensation and evaluation phases in the same stage, which progressively accumulate during the evaluation phase, leading to an even greater divergence between GNNs trained on $\mathcal{S}$ and $\mathcal{T}$, as illustrated in Fig.~\ref{fig:method_ctrl}. To further analyze the origins of accumulated errors, we introduce the following definitions:

\textbf{Definition 2.2} 
\textit{Matching Error} is the difference between the gradient trajectories generated by synthetic data and original data at this stage during condensation:
\begin{equation}
\delta_{t+1} = \nabla_{\theta}L_{S}(f_{\hat{\theta}_{t}}) - \nabla_{\theta}L_{T}(f_{\hat{\theta}_{t}}). 
\end{equation}

\textbf{Definition 2.3} 
\textit{Initialization Error} arises due to different initial parameters used during training. 
\begin{equation} I(\theta^*_{t}, \epsilon_{t}) = \nabla_{\theta}L_{T}(f_{\theta^{*}_{t} + \epsilon_{t}}) - \nabla_{\theta}L_{T}(f_{\theta^{*}_{t}}).
\end{equation}

It's important to acknowledge that because the initial training points in the condensation and evaluation phases differ, the training outcomes will inevitably vary, even if the synthetic data manages to generate gradients that are identical to those of the original data at certain stages. Specifically, if there's a stage $t$ at which $\delta_t = 0$, indicating no difference in gradients at that point, discrepancies in the final training results will still be present as long as there's an initial error in setting up the model. This underlines the significance of the initialization phase in the overall training process and its impact on the eventual performance of the model.


\noindent\fbox{%
    \parbox{\linewidth}{%
        \textbf{Theorem 2.4}
        During the evaluation process, the accumulated error at any stage is determined by the sum of matching and initialization errors.
    }%
}
\vspace{0.3em}

\begin{proof}[Proof of Theorem 2.4]
\begin{align}
\epsilon_{t+1} 
&= \hat{\theta}_{t+1} - \theta^{*}_{t+1}\nonumber\\
&= (\hat{\theta}_{t} + \nabla_{\theta}L_{S}(f_{\hat{\theta}_{t}})) - (\theta^{*}_{t}+\nabla_{\theta}L_{T}(f_{\theta^{*}_{t}})) \nonumber\\
&= (\hat{\theta}_{t} - \theta^{*}_{t}) + (\nabla_{\theta}L_{S}(f_{\hat{\theta}_{t}}) - \nabla_{\theta}L_{T}(f_{\hat{\theta}_{t}})) {\nabla_{\theta}}+ (\nabla_{\theta}L_{T}(f_{\theta^{*}_{t} + \epsilon_{t}}) - \nabla_{\theta}L_{T}(f_{\theta^{*}_{t}})) \nonumber\\
&= \epsilon_{t} + \delta_{t+1} + I(\theta^*_{t}, \epsilon_{t}) \nonumber\\
&= \sum_{i=0}^{t}\delta_{i+1} + \sum_{i=1}^{t}I(\theta^*_{i}, \epsilon_{i}) + \epsilon_0 \nonumber\\
&= \sum_{i=0}^{t}\delta_{i+1} + \sum_{i=1}^{t}I(\theta^*_{i}, \epsilon_{i}).
\end{align}
Where \( \epsilon_0 \) is the error introduced by network initialization, which can be eliminated over multiple iterations, thus the accumulated error can be represented as the sum of all matching errors in the condensation phase and all initialization errors in the evaluation phase.
\end{proof}

\textbf{Corollary 2.5}
Early matching errors continuously affect subsequent initialization errors.

\begin{proof}[Proof of corollary 2.5] 
\begin{align}
\label{eq_9}
I(\theta^*_{t}, \epsilon_{t}) 
&= \nabla_{\theta}L_{T}(f_{\theta^{*}_{t} + \epsilon_{t}}) - \nabla_{\theta}L_{T}(f_{\theta^{*}_{t}}) \nonumber
\\
&= \nabla_{\theta}L_{T}(f_{\theta^{*}_{t} + \sum_{i=0}^{t-1}\delta_{i+1} + \sum_{i=1}^{t-1}I(\theta^*_{i}, \epsilon_{i})}) {\nabla_{\theta}} -\nabla_{\theta}L_{T}(f_{\theta^{*}_{t}}).
\end{align}
As illustrated by Eq.~\ref{eq_9}, the initialization error at any given stage is derived from both the matching error and the initialization errors of the preceding stages. 
This implies that the source of the initialization errors is still the matching error. 
If the matching error is not eliminated during the condensation process, this will not only result in the emergence of initialization errors but also lead to their gradual expansion as the evaluation progresses, ultimately causing a rapid increase in cumulative errors, which in turn adversely affects the performance of the synthetic graph.
\end{proof}
\noindent\fbox{%
    \parbox{\linewidth}{%
    \textbf{Proposition 2.6}
Directly using \( \|\delta_i\|_2^2 \) as the optimization objective is limited.

    }%
}
\vspace{-0.3em}

\begin{wrapfigure}{r}{0.3\textwidth}
        \centering
        {\includegraphics[width=0.3\textwidth, angle=0]{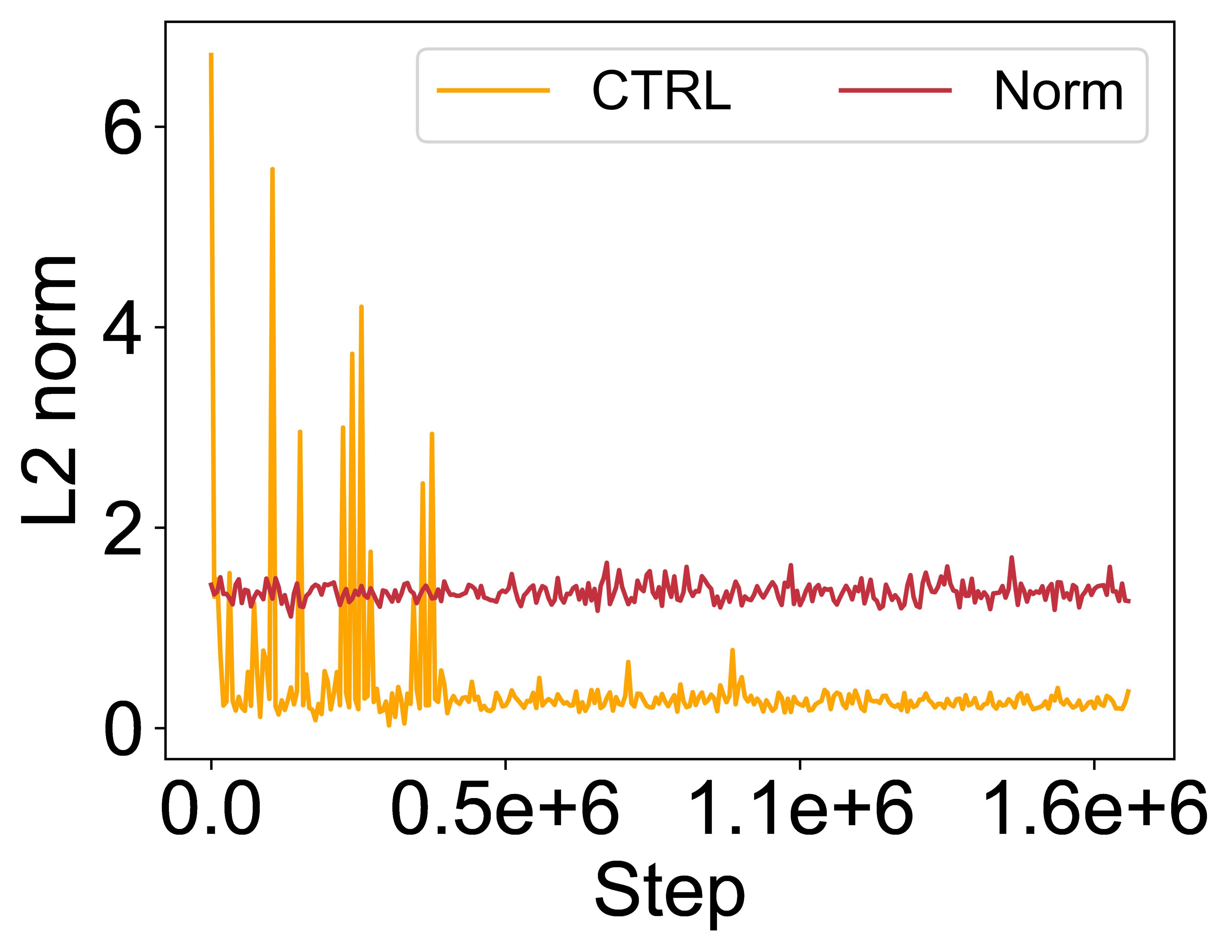}
        }
        \vspace{-4mm}
    \caption{L2 norm of the matching error.}\label{fig:norm}
    \vspace{-4mm} 
\end{wrapfigure}

To minimize accumulated error, an intuitive solution is to optimize synthetic data using the Euclidean distance between gradients as the loss function during the gradient matching stage. 
This method has achieved good results in image and graph classification tasks~\cite{mtt,jin2022condensing}. 
However, we found that in condensation tasks for node classification in graph datasets, using Euclidean distance as the loss function can lead to worse outcomes than at initialization (as shown in Fig.~\ref{fig_1}(c) and Fig.~\ref{fig:norm}).

This occurs because when the discrepancy in gradient magnitudes is minimal, employing Euclidean distance as the loss function can make it susceptible to early convergence into local minima~\cite{jiang2022delving}. Consequently, this hampers the optimization of the synthetic dataset and results in suboptimal outcomes.

\textbf{Corollary 2.7}
The proposed strategy can optimize the accumulated error by reducing the matching error during the condensation phase.

\begin{equation}\label{eqn-11}
\epsilon'_{t+1} = \sum_{i=0}^{t}\delta'_{i+1} + \sum_{i=1}^{t}I'(\theta^*_{i}, \epsilon_{i})<\epsilon_{t+1}.
\end{equation}
Where $\epsilon^{'}$, $\delta^{'}$, \(I'\) are the reduced $\epsilon$, \(I\), $\delta$, respectively. The above corollary implies that our approach can mitigate the initialization errors in the evaluation phase by significantly reducing the matching errors during the condensation phase, thereby effectively diminishing the impact of accumulated errors on the performance of the synthetic graph.

\noindent \textbf{Model summary.} In our approach, by introducing a synthetic graph initialization method that approximates the feature distribution of the original graph and a refined matching criterion, we generate a synthetic graph that more accurately matches the magnitude and direction of gradients produced by the original graph, thereby effectively reducing matching errors, as depicted in Fig.~\ref{fig:norm}.
This, consequently, leads to a significant enhancement in the performance of the synthetic graph by substantially mitigating the impact of accumulated errors.
It is noteworthy that although the final L2 norm of the matching errors between the two methods does not significantly differ in absolute terms, the quality of the resulting synthetic graph is vastly different, as shown in Fig.~\ref{fig_1}(c). \textbf{This outcome is consistent with our theoretical analysis.}

%% file: 4.Experiments.tex
\input{tables/nc}
\begin{figure*}[ht]
        \centering
        \subfigure[Accuracies on Reddit]
        {\includegraphics[width=0.3\linewidth, angle=0]{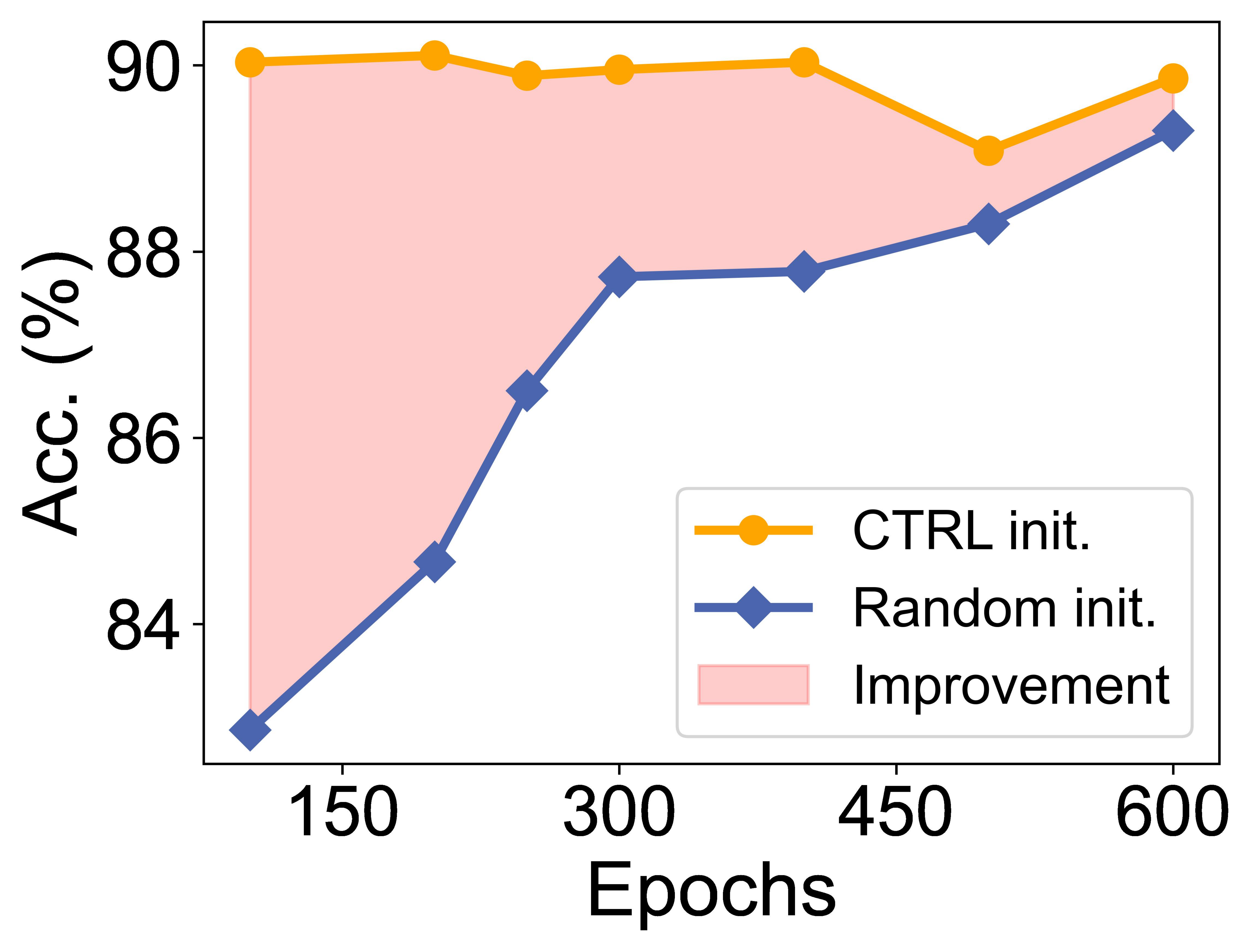}
        \hspace{3.5mm}
        \label{fig:reddit}}
        \subfigure[Accuracies on Ogbn-arxiv]
        {\includegraphics[width=0.3\linewidth, angle=0]{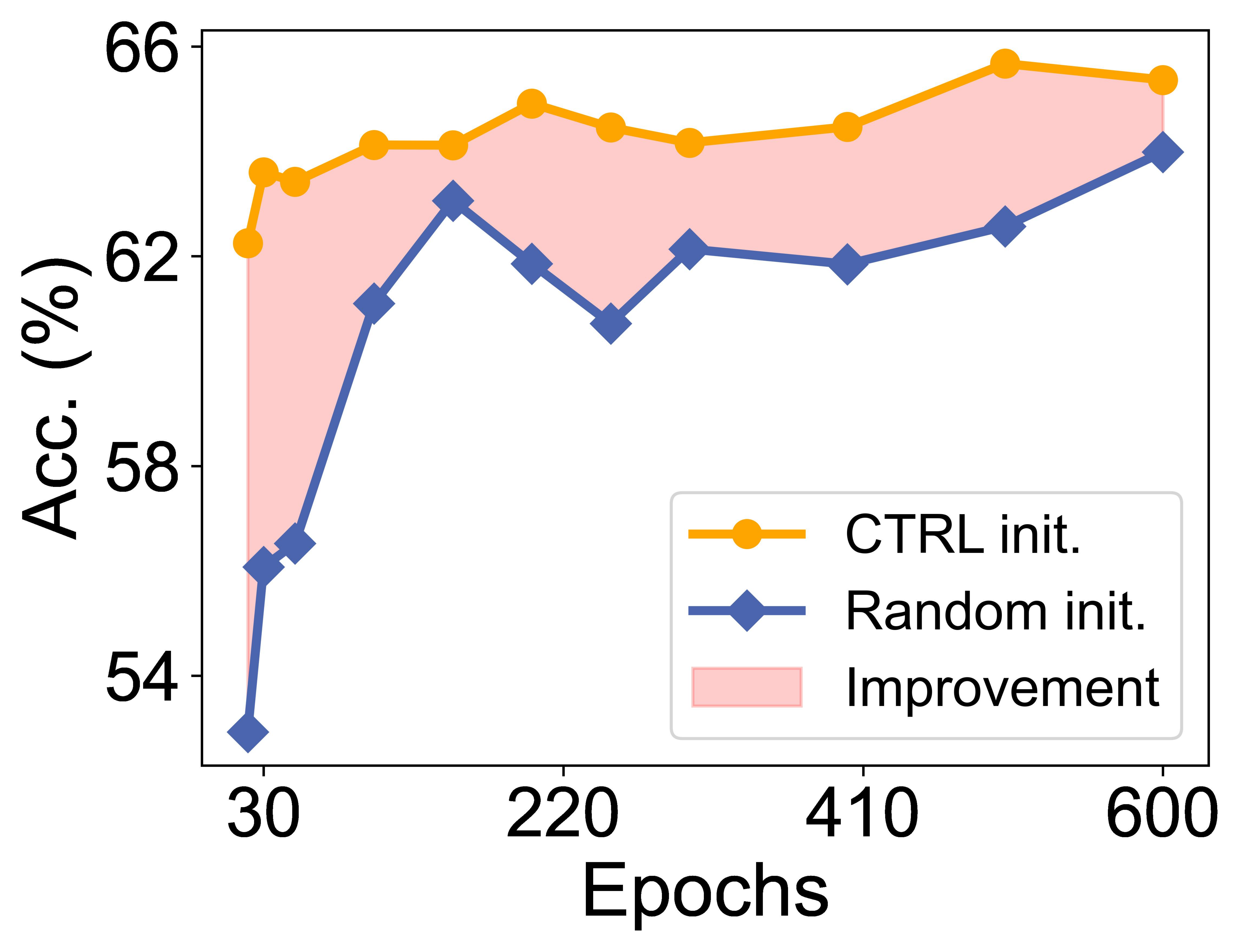}
        \hspace{3.5mm}
        \label{fig:ogb}}
        \subfigure[Sensitivity Analysis of $\beta$]
        {\includegraphics[width=0.297\linewidth, angle=0]{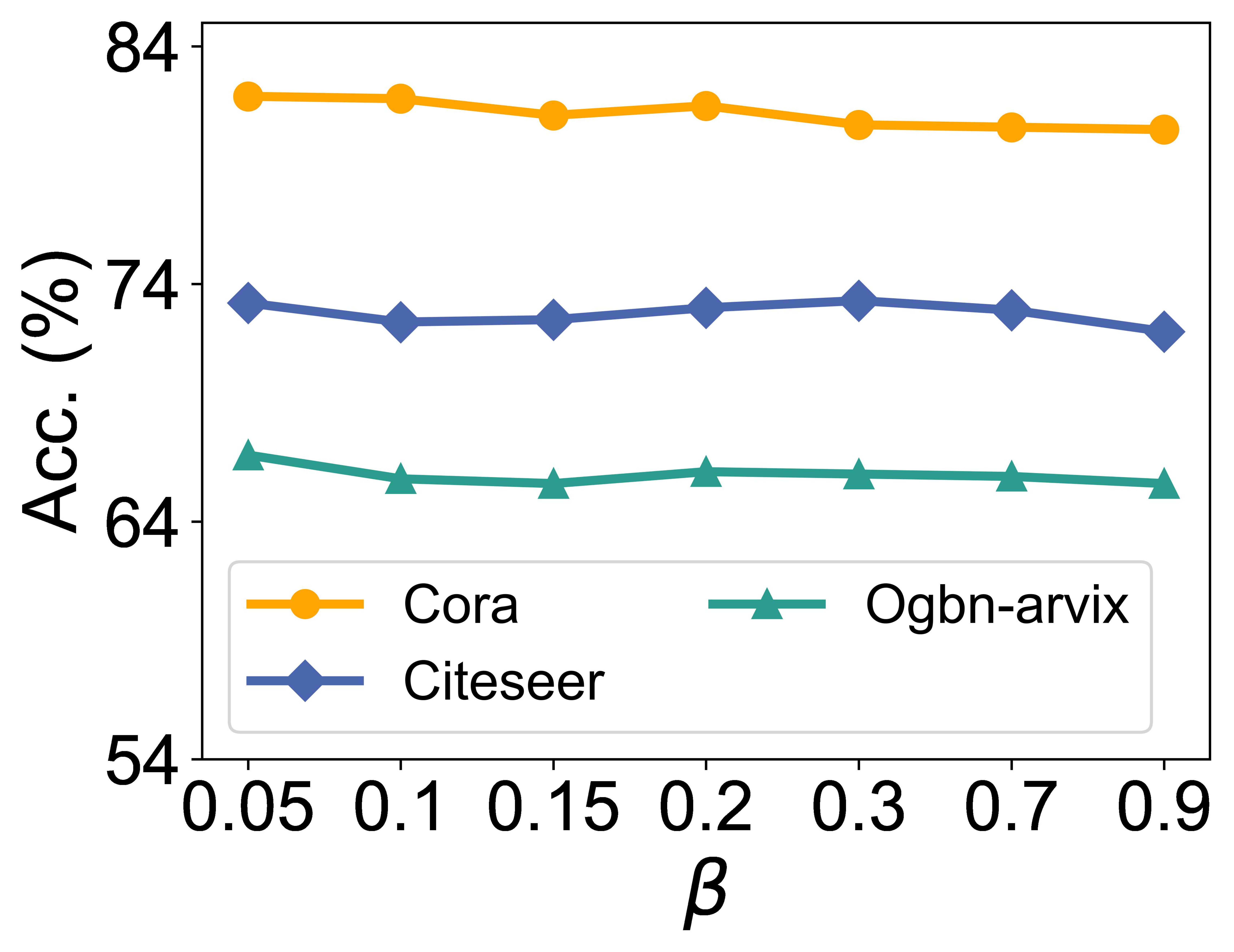}
        \label{fig:beta}}
    \vspace{-1.3mm}
    \caption{(a) and (b) show the improvement by employing the initialization method of CTRL. (c) indicates the sensitivity analysis experiments conducted on Cora, Citeseer, and Ogbn-arxiv, with condensation ratios of 1.3\%, 0.9\%, and 0.25\%, respectively.}\label{fig:A_Better}
    \vspace{-3mm} 
\end{figure*}
\vspace{-1.0em}
\section{Experiments}
\vspace{-0.5em}
\subsection{Datasets \& Implementation Details}\label{sec:exp-setting}
\vspace{-0.5em}

\textbf{Datasets.} To better evaluate the performance of CTRL, we conducted experiments on six node classification datasets: Cora, Citeseer~\cite{kipf2016semi}, Ogbn-arxiv~\cite{hu2020open}, Ogbn-xrt~\cite{chien2021node}, Flickr~\cite{zeng2020graphsaint}, and Reddit~\cite{hamilton2017inductive}, as well as six graph classification datasets, multiple molecular datasets from Open Graph Benchmark (OGB)~\cite{hu2020open}, TU Datasets (MUTAG and NCI1)~\cite{morris2020tudataset}, and one superpixel dataset CIFAR10~\cite{dwivedi2020benchmarking}.
For the dataset partitioning settings, we follow the settings of  \cite{jin2022graph} and  \cite{jin2022condensing}. 

\textbf{Implementation details.} For node classification, without specific designation, in the condense stage, we adopt the 2-layer GCN with 128 hidden units as the backbone, and we adopt the settings on \cite{jin2022graph} and \cite{jin2022condensing}. For graph classification, we adopt a 3-layer GCN with 128 hidden units as the model for one-step gradient matching. More details can be found in Appendix \ref{section: details}.

\textbf{Evaluation.} We first train a GCN classifier on synthetic graphs generated by different baseline algorithms. Then, we evaluate the classifier's performance on the test sets of real graphs.
In the context of node classification tasks, we condense the entire graph and the training set graphs for transductive datasets and inductive datasets, respectively.
For graph classification tasks, we repeat the generation process of condensed graphs 5 times with different random seeds and train GCN on these graphs with 10 different random seeds. 
Finally, we report the average performance and variance across all kinds of experiments.

\vspace{-0.5em}
\subsection{Comparison with Baselines}
\vspace{-0.5em}

For node classification tasks, we compare our proposed CTRL with 7 existing popular compression baselines: graph coarsening methods~\cite{huang2021scaling}, Random, which randomly selected nodes to form the original graph, core set methods (Herding~\cite{welling2009herding} and K-Center~\cite{sener2017active}), DC-Graph~\cite{jin2022graph}, and the state-of-the-art graph condensation methods GCond~\cite{jin2022graph} and SFGC~\cite{zheng2023structure}. For graph classification tasks, we compare the proposed method with the above three coreset methods: Random, Herding, and K-Center, the above DC-Graph, and Doscond~\cite{jin2022condensing}. 
We showcase the performances in Table \ref{tab:nc} and Table \ref{tab:gc}, from which we can make the following observations:

\textbf{Obs 1.}
In most cases of the node classification tasks, CTRL consistently surpasses the baseline methods, establishing a new benchmark.
Compared with GCond, a conventional gradient matching approach, CTRL demonstrates superior performance across three traditional transductive datasets: Cora, Citeseer, and Ogbn-arxiv. Specifically, it delivers up to a 6.4\%  enhancement in performance, averaging an increase of 3.2\%, and achieves \textbf{lossless} condensation on the first two datasets.
Even compared to the state-of-the-art unstructured dataset condensation methods, our approach still achieves better results in the majority of tasks with less time and space consumption. Additionally, \textbf{our method is currently the only one capable of achieving lossless condensation on the Flickr.}

\textbf{Obs 2.}
Across all datasets for graph classification, CTRL outperformed the current state-of-the-art condensation method, DosCond, particularly on the ogbg-molbace dataset, where it achieved an average improvement of 5.9\% and a peak of 6.2\%. Furthermore, \textbf{CTRL is the first method accomplishing lossless condensation on both the ogbg-molbace and ogbg-molbbbp datasets.} 
This finding indicates that the issue arising from measuring gradient differences using a single metric is not confined to graph condensation tasks based on node classification alone. It further demonstrates the generalizability and necessity of our approach, suggesting that our method has the potential to be directly applied to other condensation methods.

\textbf{Remark.}
To further validate the generality of our method, we also applied a more refined gradient matching strategy to image dataset condensation, resulting in an average improvement of about 0.2\% and a maximum improvement of about 1.1\%. Although there is a certain degree of improvement, it is not as remarkable as the improvements observed in graph dataset condensation. A possible reason for this could be that matching the magnitude of gradients helps the synthetic graph to fit the frequency domain distribution of the original graph. This point is elaborated in detail in Appendix~\ref{sec_cv} and subsequent experiments.

\input{tables/gc}
\input{tables/Signal}
\vspace{-0.5em}
\subsection{Analysis of Components and Performance}
\vspace{-0.5em}
\textbf{Ablation study \& Sensitivity analysis.}
We further study the impact of our initialization method and the sensitivity of $\beta$, with specific experimental details provided in Appendix~\ref{sec_abl}. Here we can make the following observations:

\textbf{Obs 3.} As shown in Fig.~\ref{fig:reddit} and~\ref{fig:ogb}, the initialization strategy in CTRL not only improves the performances by approximately$\sim10\%$ at the beginning stages but also leads to significant improvements throughout the condensation process. 
This suggests that CTRL's initialization method offers a superior starting point for synthetic graphs.

\textbf{Obs 4.}
As indicated in Fig.~\ref{fig:beta}, when the value of $\beta$ varies, the change in test accuracy does not exceed 1.5\%. 
These results indicate that our proposed matching criterion does not overly rely on the choice of the $\beta$.

\textbf{Quantitative analysis of frequency domain similarity.}\label{sec:Quantitative}

To assess the fidelity with which CTRL-generated synthetic graphs replicate the frequency domain distribution of the original graph, we first calculate various metrics separately for both the synthetic and original graphs. The chosen metrics for evaluating the frequency domain distribution are detailed in Appendix~\ref{sec-metric}. We then visually represent the correlation by computing the Pearson coefficient, which illustrates the general trend. Tab.~\ref{tab:signals} exhibits that:

\textbf{Obs 5.}
The comparison a greater similarity in the frequency distribution between the CTRL-generated synthetic graph and the original graph, which provides support for the previously mentioned capability of matching gradient magnitudes to approximate the frequency domain distribution.

\textbf{Neural architecture search evaluation.}
Following the setting in \cite{jin2022graph}, we search 480 neural architectures on condensed graphs of the Cora, Citeseer, and Pubmed datasets. 
We documented the Pearson correlation between the validation accuracies achieved by architectures trained on condensed graphs versus those trained on original graphs and the average test accuracy of the searched architectures.

\textbf{Obs 6.}
The results reported in Table \ref{tab:NAS} demonstrate the consistent superiority of our method, evidenced by higher Pearson correlation coefficients~\cite{9737295} and improved test performance, with enhancements of up to 0.14 and 0.4\%, respectively. 
Notably, \textbf{our method is capable of discovering architectures that outperform those searched through the entire Cora and Citeseer.}

\input{tables/cross}
\input{tables/crossarch_cora}

\textbf{Cross-architecture generalization analysis.}
To further demonstrate the transferability of our method, we first utilize SGC to generate synthetic graphs, then evaluate them on a variety of architectures, including APPNP, GCN, SGC, GraphSAGE~\cite{hamilton2017inductive}, Cheby and GAT~\cite{velivckovic2017graph}, as well as a standard MLP. 
Furthermore, We conduct condensation with different structures on the Cora and Citeseer datasets and evaluate the cross-architecture generalization performance of the condensed graphs produced by different structures.
Tab.~\ref{tab:cross} and \ref{tab:crossarch_cora} document the results of those experiments, from which we can draw the following conclusions:

\textbf{Obs 7.}
As shown in Tab.~\ref{tab:cross}, compared to GCond, our synthetic graphs generally exhibit better cross-architecture performance(up to 2\%), demonstrating that the finer-grained gradient matching does not lead the condensed graph to overfit to a single neural network architecture.

\textbf{Remark.}
To make a fair comparison,
we incorporated our refined matching criterion into SFGC and conducted a parallel comparison. The empirical results confirm that this improvement can further enhance the cross-architecture performance. Details are included in the Appendix~\ref{tab:cross_sfgc}.

\textbf{Obs 8.}
The results in Tab.~\ref{tab:crossarch_cora} show that our approach achieves improved performance over GCond on the majority of model-dataset combinations, demonstrating that CTRL is versatile and achieves consistent gains across diverse architectures and datasets.

%% file: tables/nc.tex
\definecolor{lightblue}{RGB}{153,255,255}
\sethlcolor{lightblue}
\begin{table*}[ht]
  \centering
  \caption{Performance comparison to baselines in the node classification tasks. 
  We report test accuracy (\%) on Citeseer, Cora, Ogbn-arxiv, Ogbn-xrt, Flickr and Reddit. \textbf{Bold entries} are best results, \hl{highlight} mark the lossless results. The accuracy with the error bar reported in the main table is the maximum of top-1 test results 5 times repeated.}
  \label{tab:nc}
  \fontsize{6.5}{6}\selectfont
   \begin{tabular}{c@{\hskip 0.2cm}c|c@{\hskip 0.2cm}c@{\hskip 0.2cm}c@{\hskip 0.2cm}c@{\hskip 0.2cm}c@{\hskip 0.2cm}c@{\hskip 0.2cm}c|c|c}
    \toprule
    Dataset & Ratio & Random & Herding & K-Center & Coarsening & DC-Graph & GCond & SFGC & CTRL(ours) & Whole Dataset\\
    \midrule
    \multirow{3}{*}{Citeseer} & 0.90\% & 54.4$_{\pm 4.4}$ & 57.1$_{\pm 1.5}$ & 52.4$_{\pm 2.8}$ & 52.2$_{\pm 0.4}$ & 66.8$_{\pm 1.5}$ & 70.5$_{\pm 1.2}$ & 71.4$_{\pm 0.5}$ & \textbf{\hl{73.3$_{\pm 0.4}$}} & \multirow{3}{*}{71.7$_{\pm 0.1}$} \\
    & 1.80\% & 64.2$_{\pm 1.7}$ & 66.7$_{\pm 1.0}$ & 64.3$_{\pm 1.0}$ &  66.9$_{\pm 0.9}$ & 59.0$_{\pm 0.5}$ & 70.6$_{\pm 0.9}$ & \hl{72.4$_{\pm 0.4}$} & \textbf{\hl{73.5$_{\pm 0.1}$}} \\
     & 3.60\% & 69.1$_{\pm 0.1}$ & 69.0$_{\pm 0.1}$ & 69.1$_{\pm 0.1}$ & 65.3$_{\pm 0.5}$  & 66.3$_{\pm 1.5}$ & 69.8$_{\pm 1.4}$ & 70.6$_{\pm 0.7}$ & \textbf{\hl{73.4$_{\pm 0.2}$}} \\
    \midrule
    \multirow{3}{*}{Cora} & 1.30\%  & 63.6$_{\pm 3.7}$ & 67.0$_{\pm 1.3}$ & 64.0$_{\pm 2.3}$ & 31.2$_{\pm 0.2}$ &  67.3$_{\pm 1.9}$ & 79.8$_{\pm 1.3}$ & 80.1$_{\pm 0.4}$ & \textbf{\hl{81.9$_{\pm 0.1}$}} & \multirow{3}{*}{81.2$_{\pm 0.2}$} \\
    & 2.60\% &72.8$_{\pm 1.1}$ & 73.4$_{\pm 1.0}$ & 73.2$_{\pm 1.2}$  & 65.2$_{\pm 0.6}$ &  67.6$_{\pm 3.5}$ & 80.1$_{\pm 0.6}$ & \hl{81.7$_{\pm 0.5}$} & \textbf{\hl{81.8$_{\pm 0.1}$}} \\
     & 5.20\%  & 76.8$_{\pm 0.1}$ & 76.8$_{\pm 0.1}$ & 76.7$_{\pm 0.1}$ & 70.6$_{\pm 0.1}$ & 67.7$_{\pm 2.2}$ & 79.3$_{\pm 0.3}$ & \hl{81.6$_{\pm 0.8}$} & \textbf{\hl{81.8$_{\pm 0.1}$}} \\
    \midrule
    \multirow{3}{*}{Ogbn-arxiv} & 0.05\% & 47.1$_{\pm 3.9}$ & 52.4$_{\pm 1.8}$ & 47.2$_{\pm 3.0}$ & 35.4$_{\pm 0.3}$ & 58.6$_{\pm 0.4}$ & 59.2$_{\pm 1.1}$ & 65.5$_{\pm 0.7}$ & \textbf{65.6$_{\pm 0.3}$} & \multirow{3}{*}{71.4$_{\pm 0.1}$} \\
    & 0.25\%  &57.3$_{\pm 1.1}$ & 58.6$_{\pm 1.2}$ & 56.8$_{\pm 0.8}$ & 43.5$_{\pm 0.2}$ &59.9$_{\pm 0.3}$ & 63.2$_{\pm 0.3}$ & 66.1$_{\pm 0.4}$ & \textbf{66.5$_{\pm 0.3}$} \\
     & 0.50\%  &60.0$_{\pm 0.9}$ & 60.4$_{\pm 0.8}$ & 60.3$_{\pm 0.4}$ & 50.4$_{\pm 0.1}$ &  59.5$_{\pm 0.3}$ & 64.0$_{\pm 1.4}$ & 66.8$_{\pm 0.4}$ & \textbf{67.6$_{\pm 0.2}$} \\
      \midrule
    \multirow{3}{*}{Ogbn-xrt} & 0.05\% &55.1$_{\pm 2.1}$ & 60.6$_{\pm 2.3}$ & 52.7$_{\pm 3.5}$& 48.4$_{\pm 0.4}$ &  60.4$_{\pm 0.5}$ & 55.1$_{\pm 0.3}$ & 55.4$_{\pm 0.1}$ & \textbf{61.1$_{\pm 0.5}$} & \multirow{3}{*}{73.4$_{\pm 0.1}$} \\
    & 0.25\% & 64.2$_{\pm 1.4}$ & 64.4$_{\pm 1.7}$ & 60.8$_{\pm 2.2}$ & 51.3$_{\pm 0.3}$ & 61.5$_{\pm 0.2}$ & 68.1$_{\pm 0.4}$ & 67.7$_{\pm 0.1}$ & \textbf{69.4$_{\pm 0.4}$} \\
     & 0.50\% &65.7$_{\pm 2.1}$ & 68.3$_{\pm 2.2}$ & 62.3$_{\pm 3.4}$ & 56.9$_{\pm 0.1}$ & 68.2$_{\pm 0.3}$ & 69.3$_{\pm 0.4}$ & 66.5$_{\pm 0.1}$ & \textbf{70.4$_{\pm 0.1}$} \\
      \midrule
    \multirow{3}{*}{Flickr} & 0.10\% &41.8$_{\pm 2.0}$ & 42.5$_{\pm 1.8}$ & 42.0$_{\pm 0.7}$  & 41.9$_{\pm 0.2}$  &46.3$_{\pm 0.2}$ & 46.5$_{\pm 0.4}$ & 46.6$_{\pm 0.2}$ & \textbf{47.1$_{\pm 0.1}$} & \multirow{3}{*}{47.2$_{\pm 0.1}$} \\
    & 0.50\%  & 44.0$_{\pm 0.4}$ & 43.9$_{\pm 0.9}$ & 43.2$_{\pm 0.1}$ & 44.5$_{\pm 0.1}$ &  45.9$_{\pm 0.1}$ & 47.1$_{\pm 0.1}$ & 47.0$_{\pm 0.1}$ & \textbf{\hl{47.4$_{\pm 0.1}$}} \\
     & 1.0\% & 44.6$_{\pm 0.2}$ & 44.4$_{\pm 0.6}$ & 44.1$_{\pm 0.4}$ & 44.6$_{\pm 0.1}$ &  45.8$_{\pm 0.1}$ & 47.1$_{\pm 0.1}$ & 47.1$_{\pm 0.1}$ & \textbf{\hl{47.5$_{\pm 0.1}$}} \\
      \midrule
    \multirow{3}{*}{Reddit} & 0.01\%  &46.1$_{\pm 4.4}$ & 53.1$_{\pm 2.5}$ & 46.6$_{\pm 2.3}$ & 40.9$_{\pm 0.5}$ & 88.2$_{\pm 0.2}$ & 88.0$_{\pm 1.8}$ & \textbf{89.7$_{\pm 0.2}$} & 89.2$_{\pm 0.2}$ & \multirow{3}{*}{93.9$_{\pm 0.0}$} \\
    & 0.05\% &58.0$_{\pm 2.2}$ & 62.7$_{\pm 1.0}$ & 53.0$_{\pm 3.3}$  & 42.8$_{\pm 0.8}$ &  89.5$_{\pm 0.1}$ & 89.6$_{\pm 0.7}$ & 90.0$_{\pm 0.3}$ & \textbf{90.6$_{\pm 0.2}$} \\
     & 0.50\%  &66.3$_{\pm 1.9}$ & 71.0$_{\pm 1.6}$ & 58.5$_{\pm 2.1}$ & 47.4$_{\pm 0.9}$ &  90.5$_{\pm 1.2}$ & 90.1$_{\pm 0.5}$  & 90.3$_{\pm 0.3}$ & \textbf{91.9$_{\pm 0.4}$} \\
    \bottomrule
    \end{tabular}%
\end{table*}

%% file: tables/gc.tex
\definecolor{lightblue}{RGB}{153,255,255}
\sethlcolor{lightblue}
\begin{table*}[ht]
  \centering
  \caption{The graph classification performance comparison to baselines. \textit{CTRL achieves the highest results in all cases
  on graph classification and lossless results in 2 of 6 datasets.} We report the ROC-AUC for the first three datasets and accuracies (\%) for others. Whole Dataset indicates the performance of the original dataset. \textbf{Bold entries} are best results, 
  \hl{highlight} mark the lossless results.}
  \label{tab:gc}
   \fontsize{6.5}{5}\selectfont
   \begin{tabular}{c@{\hskip 0.2cm}c|c@{\hskip 0.2cm}c@{\hskip 0.2cm}c@{\hskip 0.2cm}c@{\hskip 0.2cm}c|c@{\hskip 0.2cm}c}
    \toprule
    Dataset & Ratio & Random & Herding & K-Center & DCG & DosCond & CTRL(ours) & Whole Dataset\\
    \midrule
    \multirow{3}{*}{ogbg-molbace} & 0.20\% & 0.580$_{\pm 0.067}$ & 0.548$_{\pm 0.034}$ & 0.548$_{\pm 0.034}$ &  0.623$_{\pm 0.046}$ & 0.657$_{\pm0.034}$  & \textbf{\hl{0.716$_{\pm 0.025}$}} & \multirow{3}{*}{0.724$_{\pm 0.005}$} \\
    & 1.70\% & 0.598$_{\pm 0.073}$ & 0.639$_{\pm 0.039}$ & 0.591$_{\pm 0.056}$ & 0.655$_{\pm 0.033}$ & 0.674$_{\pm 0.035}$ & \textbf{\hl{0.736$_{\pm 0.014}$}} \\ 
    & 8.30\% & 0.632$_{\pm 0.047}$ & 0.683$_{\pm 0.022}$ & 0.589$_{\pm 0.025}$ & 0.652$_{\pm 0.013}$ & 0.688$_{\pm 0.012}$& \textbf{\hl{0.745$_{\pm 0.009}$}} \\
    \midrule
    \multirow{3}{*}{ogbg-molbbbp} & 0.10\% & 0.519$_{\pm 0.016}$ & 0.546$_{\pm 0.019}$ & 0.546$_{\pm 0.019}$ &  0.559$_{\pm 0.044}$ & 0.581$_{\pm0.005}$  & \textbf{0.592$_{\pm 0.011}$} & \multirow{3}{*}{0.646$_{\pm 0.004}$} \\
    & 1.20\% & 0.586$_{\pm 0.040}$ & 0.605$_{\pm 0.019}$ & 0.530$_{\pm 0.039}$ & 0.568$_{\pm 0.032}$ & 0.605$_{\pm 0.008}$ & \textbf{0.629$_{\pm 0.006}$} \\ 
    & 6.10\% & 0.606$_{\pm 0.020}$ & 0.617$_{\pm 0.003}$ & 0.576$_{\pm 0.019}$ & 0.579$_{\pm 0.032}$ & 0.620$_{\pm 0.007}$& \textbf{\hl{0.650$_{\pm 0.005}$}} \\
    \midrule
    \multirow{3}{*}{ogbg-molhiv} & 0.01\% & 0.719$_{\pm 0.009}$ & 0.721$_{\pm 0.002}$ & 0.721$_{\pm 0.002}$ &  0.718$_{\pm 0.013}$ & 0.726$_{\pm0.003}$  & \textbf{0.732$_{\pm 0.006}$} & \multirow{3}{*}{0.757$_{\pm 0.008 }$} \\
    & 0.06\% & 0.720$_{\pm 0.011}$ & 0.725$_{\pm 0.006}$ & 0.713$_{\pm 0.009}$ & 0.728$_{\pm 0.002}$ & 0.728$_{\pm 0.005}$ & \textbf{0.734$_{\pm 0.008}$} \\ 
    & 0.30\% & 0.721$_{\pm 0.014}$ & 0.725$_{\pm 0.003}$ & 0.725$_{\pm 0.006}$ & 0.726$_{\pm 0.010}$ & 0.731$_{\pm 0.004}$& \textbf{0.737$_{\pm 0.006}$}\\
    \midrule
    \multirow{2}{*}{MUTAG} & 1.30\% & 67.47$_{\pm 9.74}$ & 70.84$_{\pm 7.71}$ & 70.84$_{\pm 7.71}$ &  75.00$_{\pm 8.16}$ & 82.21$_{\pm1.61}$  & \textbf{83.06$_{\pm3.15}$} & 
    \multirow{2}{*}{88.63$_{\pm 1.44}$} \\
    & 13.30\% & 77.89$_{\pm 7.55}$ & 80.42$_{\pm 1.89}$ & 81.00$_{\pm 2.51}$ & 82.66$_{\pm 0.68}$ & 82.76$_{\pm 2.31}$ & 
    \textbf{83.16$_{\pm 3.62}$} \\ 
    \midrule
    \multirow{3}{*}{NCI1} & 0.10\% & 51.27$_{\pm 1.22}$ & 53.98$_{\pm 0.67}$ & 53.98$_{\pm 0.67}$ &  51.14$_{\pm 1.08}$ & 56.58$_{\pm0.48}$  & \textbf{56.69$_{\pm 0.69}$} & \multirow{3}{*}{71.70$_{\pm 0.20}$} \\
    & 0.60\% & 54.33$_{\pm 3.14}$ & 57.11$_{\pm 0.56}$ & 53.21$_{\pm 1.44}$ & 51.86$_{\pm 0.81}$ & 58.02$_{\pm 1.05}$ & \textbf{58.04$_{\pm 1.28}$} \\ 
    & 3.10\% & 58.51$_{\pm 1.73}$ & 58.94$_{\pm 0.83}$ & 56.58$_{\pm 3.08}$ & 52.17$_{\pm 1.90}$ & 60.07$_{\pm 1.58}$& \textbf{60.14$_{\pm 1.73}$} \\
    \midrule
    \multirow{3}{*}{CIFAR10} & 0.06\% & 15.61$_{\pm 0.52}$ & 22.38$_{\pm 0.49}$ & 22.37$_{\pm 0.50}$ & 21.60$_{\pm 0.42}$ & 24.70$_{\pm0.70}$  & \textbf{29.30$_{\pm 0.27}$} & \multirow{3}{*}{50.75$_{\pm 0.14}$} \\
    & 0.20\% & 23.07$_{\pm 0.76}$ & 28.81$_{\pm 0.35}$ & 20.93$_{\pm 0.62}$ & 29.27$_{\pm 0.77}$ & 30.70$_{\pm 0.23}$ & \textbf{31.21$_{\pm 0.20}$} \\ 
    & 1.10\% & 30.56$_{\pm 0.81}$ & 33.94$_{\pm 0.37}$ & 24.17$_{\pm 0.51}$ & 34.47$_{\pm 0.52}$ & 35.34$_{\pm 0.14}$& \textbf{35.53$_{\pm 0.38}$} \\

    \bottomrule
    \end{tabular}%
\end{table*}

%% file: tables/Signal.tex
\begin{table*}[ht]
    \begin{minipage}[t]{0.36\textwidth} 
        \centering
        \fontsize{6.5}{8}\selectfont
        \caption{
         Comparison of the frequency domain similarity between synthetic graphs from CTRL and GCond with original graphs.}
        \begin{tabularx}{\linewidth}{*{2}{>{\centering\arraybackslash}X}@{\hspace{0.55cm}}*{1}{>{\centering\arraybackslash}X}}
         \toprule
         \multirow{2}{*}{Dataset} & \multicolumn{2}{c}{Pearson Correlation / P-value}\\
          \cmidrule(lr){2-3} 
          & GCond & CTRL\\
          \midrule
         Cora & 0.90/0.016 & \textbf{0.96/0.003} \\
         Citeseer & 0.92/0.009 & \textbf{0.95/0.003}\\
         Pumbed & 0.89/0.016 & \textbf{0.93/0.014} \\
          \bottomrule
        \end{tabularx}
        \label{tab:signals}
      \end{minipage}
  \hfill
  \begin{minipage}[t]{0.60\textwidth}
    \centering
     \fontsize{6.5}{8}\selectfont
    \caption{Neural Architecture Search. Methods are compared in validation accuracy correlation and test accuracy obtained by searched architecture. Whole refers to the test accuracy obtained by the architecture searched using the whole dataset.}
    \begin{tabularx}{\linewidth}{*{6}{>{\centering\arraybackslash}X}}
    \toprule
     \multirow{2}{*}{Dataset} & \multicolumn{4}{c}{Pearson Correlation / Performance(\%)} & \multirow{2}{*}{Whole} \\
     \cmidrule(lr){2-5} 
     &Random&Herding&GCond&CTRL&Acc.(\%)\\
     \midrule
    Cora & 0.40/82.9 & 0.21/82.9  & 0.76/83.1 & \textbf{0.86}/\textbf{83.2} & 82.6 \\
    Citeseer & 0.56/71.4 & 0.29/71.3 & 0.79/71.3 & \textbf{0.93}/\textbf{72.0} &71.6 \\
    Pubmed & 0.55/80.0 & 0.21/79.9 & 0.81/79.7 &\textbf{0.83}/\textbf{80.1} & 80.5 \\
    \bottomrule
  \label{tab:NAS}%
   \end{tabularx}
    \label{tab:table1}
  \end{minipage}%
  \vfill
  \vspace{-2.5em}
\end{table*}

%% file: tables/cross.tex
\begin{table*}[ht]
\centering
\fontsize{6.5}{5}\selectfont
\setlength{\extrarowheight}{0.8pt} 
\caption{Performance across different GNN architectures. \colorbox{orange!30}{\makebox(14,6){Avg.}} and \colorbox{red!30}{\makebox(14,6){Std.}}: the average performance and the standard deviation of the results of APPNP, Cheby, GCN, SAGE and SGC, \colorbox{cyan!20}{\makebox(20,6){$\Delta$(\%)}} denotes the improvements upon the DG-Graph. \textbf{Bold entries} are the best results.}
\label{tab:cross}
\resizebox{0.9\textwidth}{!}{%
\begin{tabular}{cccccccccccccc}
\cline{1-12}
                          &         & \multicolumn{7}{c}{Architectures}                & \multicolumn{3}{c}{Statistics}                                                                                         \\ \cline{3-12}
\multirow{-2}{*}{Datasets} & \multirow{-2}{*}{Methods} & MLP  & GAT  & APPNP & Cheby & GCN & SAGE & SGC & \cellcolor[HTML]{FFE6CC}Avg.          & \cellcolor[HTML]{FFB2B2}Std.          & \cellcolor[HTML]{C6EFFC}$\Delta$(\%)    \\ \cline{1-9}

                          & DC-Graph & 67.2 & - &67.1 &67.7 &67.9 &66.2 &72.8 &  \cellcolor[HTML]{FFE6CC}68.3          & \cellcolor[HTML]{FFB2B2}2.6          & \cellcolor[HTML]{C6EFFC}-               \\
                           & GCond & 73.1 &66.2 &78.5 &76.0 &80.1 &78.2 &79.3 & \cellcolor[HTML]{FFE6CC}78.4          & \cellcolor[HTML]{FFB2B2}4.9         & $\uparrow$ \cellcolor[HTML]{C6EFFC}10.1            \\
\multirow{-3.3}{*}{Cora}  & CTRL & 77.4 & 66.7 & 80.8 & 77.0  &81.6 & 79.0 & 81.5 & \cellcolor[HTML]{FFE6CC}\textbf{80.0} & \cellcolor[HTML]{FFB2B2}\textbf{2.0} & $\uparrow$ \cellcolor[HTML]{C6EFFC}\textbf{11.7}   \\ \cline{1-9}
\multirow{-3.7}{*}{\centering (Ratio = 1.80\%)}

                          & DC-Graph & 59.9& -& 60.0& 55.7 &59.8 &60.0 & 60.4 & \cellcolor[HTML]{FFE6CC}59.2          & \cellcolor[HTML]{FFB2B2}\textbf{2.0}           & \cellcolor[HTML]{C6EFFC}-               \\
                           & GCond & 62.2& 60.0 &63.4 &54.9& 63.2 &62.6& 63.7&  \cellcolor[HTML]{FFE6CC}61.6          & \cellcolor[HTML]{FFB2B2}3.7          & $\uparrow$ \cellcolor[HTML]{C6EFFC}2.4             \\
\multirow{-3.3}{*}{Ogbn-arxiv}   & CTRL & 64.1 & 60.3 &65.3 & 57.7 &65.8 & 62.9 & 66.1  & \cellcolor[HTML]{FFE6CC}\textbf{63.6} & \cellcolor[HTML]{FFB2B2}3.5  & $\uparrow$ \cellcolor[HTML]{C6EFFC}\textbf{4.4}   \\ \cline{1-9}
\multirow{-3.7}{*}{\centering (Ratio = 0.25\%)}

                          & DC-Graph &43.1 & -& 45.7 &43.8 &45.9 &45.8 &45.6  & \cellcolor[HTML]{FFE6CC}45.4        & \cellcolor[HTML]{FFB2B2}\textbf{0.9}          & \cellcolor[HTML]{C6EFFC}-              \\
                           & GCond & 44.8 & 40.1 & 45.9 & 42.8 & 47.1& 46.2& 46.1 & \cellcolor[HTML]{FFE6CC}45.6         & \cellcolor[HTML]{FFB2B2}1.6           & $\uparrow$ \cellcolor[HTML]{C6EFFC}0.2             \\
\multirow{-3.3}{*}{Flickr}     & CTRL & 42.6 & 41.0 & 46.2 & 43.1 & 47.2 &46.5 &46.7 & \cellcolor[HTML]{FFE6CC}\textbf{45.9} & \cellcolor[HTML]{FFB2B2}1.6  & $\uparrow$ \cellcolor[HTML]{C6EFFC}\textbf{0.5}  \\ \cline{1-9}
\multirow{-3.7}{*}{\centering (Ratio = 0.50\%)}
                           & DC-Graph & 50.3 & - & 81.2 & 77.5 & 89.5 & 89.7 & 90.5 & \cellcolor[HTML]{FFE6CC}85.7         & \cellcolor[HTML]{FFB2B2}\textbf{5.9}           & \cellcolor[HTML]{C6EFFC}-              \\
                          & GCond & 42.5 & 60.2 & 87.8 & 75.5 & 89.4 & 89.1 & 89.6 & \cellcolor[HTML]{FFE6CC}86.3          & \cellcolor[HTML]{FFB2B2}6.1           & $\uparrow$ \cellcolor[HTML]{C6EFFC}0.6           \\
\multirow{-3.3}{*}{Reddit} & CTRL & 43.2 & 60.0 & 87.9 & 75.3 & 90.3 & 89.1 & 89.7 & \cellcolor[HTML]{FFE6CC}\textbf{86.5}  & \cellcolor[HTML]{FFB2B2}6.3  & \cellcolor[HTML]{C6EFFC}$\uparrow$ \textbf{0.8}   \\ \cline{1-12}
\multirow{-3.7}{*}{\centering (Ratio = 0.10\%)}
\end{tabular}%
}

\end{table*}

%% file: tables/crossarch_cora.tex
\begin{table*}
  \centering
  \caption{Comparison of the cross-architecture generalization performance between GCond (on the left of /) and CTRL (on the right of /) on Cora(a) and Ogbn-arxiv(b). \textbf{Bold} entries are the best results. $\textcolor{red}{\uparrow}$/$\textcolor{blue}{\downarrow}$ : our
method show increase or decrease performance. }
\vspace{-0.5em}
  \begin{minipage}{0.45\textwidth}
      \centering
      \fontsize{6.5}{8}\selectfont
        \caption*{(a)  Cora, Ratio = 2.60\%}
        \label{subtab:cora}
         \begin{tabular}{c|c@{\hskip 0.1cm}c@{\hskip 0.1cm}c@{\hskip 0.1cm}c@{\hskip 0.1cm}c}
            \toprule
            C/T & APPNP & Cheby & GCN & SAGE & SGC \\
            \midrule
            APPNP & 72.1/\textbf{76.1}$\textcolor{red}{\uparrow}$ & 60.8/\textbf{75.1}$\textcolor{red}{\uparrow}$ & 73.5/\textbf{76.7}$\textcolor{red}{\uparrow}$ & 72.3/\textbf{72.7}$\textcolor{red}{\uparrow}$& 73.1/\textbf{78.0}$\textcolor{red}{\uparrow}$\\
            Cheby & 75.3/\textbf{78.5}$\textcolor{red}{\uparrow}$ & 71.8/\textbf{75.8}$\textcolor{red}{\uparrow}$& \textbf{76.8}/74.1$\textcolor{blue}{\downarrow}$ & \textbf{76.4}/75.2$\textcolor{blue}{\downarrow}$ & 75.5/\textbf{75.6}$\textcolor{red}{\uparrow}$\\
            GCN & 69.8/\textbf{72.5}$\textcolor{red}{\uparrow}$ & 53.2/\textbf{62.4}$\textcolor{red}{\uparrow}$& 70.6/\textbf{72.3}$\textcolor{red}{\uparrow}$& 60.2/\textbf{60.6}$\textcolor{red}{\uparrow}$& 68.7/\textbf{73.1}$\textcolor{red}{\uparrow}$\\
            SAGE & 77.1/\textbf{77.2}$\textcolor{red}{\uparrow}$ & 69.3/\textbf{75.9}$\textcolor{red}{\uparrow}$& 77.0/\textbf{79.3}$\textcolor{red}{\uparrow}$& \textbf{76.1}/75.7$\textcolor{blue}{\downarrow}$& 77.7/\textbf{79.1}$\textcolor{red}{\uparrow}$\\
            SGC & 78.5/\textbf{80.9}$\textcolor{red}{\uparrow}$ & 76.0/\textbf{77.5}$\textcolor{red}{\uparrow}$ & 80.1/\textbf{80.7}$\textcolor{red}{\uparrow}$ & \textbf{78.2}/74.8$\textcolor{blue}{\downarrow}$ & \textbf{79.3}/76.5$\textcolor{blue}{\downarrow}$\\
            \bottomrule
        \end{tabular}%
  \end{minipage}
  \hspace{10mm}
  \begin{minipage}{0.45\textwidth}
      \centering
      \fontsize{6.5}{8}\selectfont
        \caption*{(b)  Ogbn-arxiv, Ratio = 0.05\%}
        \label{subtab:arxiv}
         \begin{tabular}{c|c@{\hskip 0.1cm}c@{\hskip 0.1cm}c@{\hskip 0.1cm}c@{\hskip 0.1cm}c}
            \toprule
            C/T & APPNP & Cheby & GCN & SAGE & SGC \\
            \midrule
            APPNP & 60.3/\textbf{62.5}$\textcolor{red}{\uparrow}$ & 51.8/\textbf{57.3}$\textcolor{red}{\uparrow}$ & 59.9/\textbf{63.6}$\textcolor{red}{\uparrow}$ & 59.0/\textbf{61.3}$\textcolor{red}{\uparrow}$ & 61.2/\textbf{62.5}$\textcolor{red}{\uparrow}$\\
            Cheby & 57.4/\textbf{59.2} $\textcolor{red}{\uparrow}$& 53.5/\textbf{55.2}$\textcolor{red}{\uparrow}$& 57.4/\textbf{59.0}$\textcolor{red}{\uparrow}$& \textbf{57.1}/55.2$\textcolor{blue}{\downarrow}$ & \textbf{58.2}/57.7$\textcolor{blue}{\downarrow}$\\
            GCN & 59.3/\textbf{61.4}$\textcolor{red}{\uparrow}$ & 51.8/\textbf{55.8}$\textcolor{red}{\uparrow}$ & 60.3/\textbf{61.1}$\textcolor{red}{\uparrow}$ & 60.2/\textbf{60.3}$\textcolor{red}{\uparrow}$ & 59.2/\textbf{61.9}$\textcolor{red}{\uparrow}$\\
            SAGE & 57.6/\textbf{60.7}$\textcolor{red}{\uparrow}$ & \textbf{53.9}/53.3$\textcolor{blue}{\downarrow}$ & 58.1/\textbf{62.9}$\textcolor{red}{\uparrow}$& \textbf{57.8}/55.5$\textcolor{blue}{\downarrow}$& 59.0/\textbf{61.7}$\textcolor{red}{\uparrow}$\\
            SGC & 57.6/\textbf{60.6}$\textcolor{red}{\uparrow}$ & 53.9/\textbf{55.9}$\textcolor{red}{\uparrow}$& 58.1/\textbf{61.8}$\textcolor{red}{\uparrow}$& 57.8/\textbf{58.9}$\textcolor{red}{\uparrow}$& 59.0/\textbf{61.6}$\textcolor{red}{\uparrow}$\\
            \bottomrule
        \end{tabular}%
  \end{minipage}
  \label{tab:crossarch_cora}
\end{table*}

%% file: 5.Conclusion.tex
\vspace{-2mm}
\section{Conclusion}
\vspace{-2mm}

In this paper, we introduce a novel approach for graph condensation, which effectively reduces the impact of accumulated errors arising from the disparities between gradients produced by synthetic and original data through finer-grained gradient matching and a more rational initialization process.
Extensive experimental results and analysis confirm the superiority of the proposed method in synthesizing high-quality small-scale graphs, including node classification, graph classification, cross-architecture tasks, and neural architecture search tasks. Furthermore, considering the universality of the CTRL concept, it can be effectively integrated into various data condensation methods.

\textbf{Limitations and future work.} CTRL is limited to gradient matching framework, we will explore its applications in more matching approaches (metrics) in the future.

%% file: 2.Related_work.tex
\section{Related work}
\vspace{-0.5em}

\textbf{Dataset distillation or condensation.}
Dataset Distillation (DD)~\cite{wang2018dataset,bohdal2020flexible,nguyen2020dataset,liu2022dataset,zhou2023dataset,wang2023dim,lu2023can} aims to condense large datasets into smaller versions for comparable model training, whereas Dataset Condensation (DC)~\cite{zhao2020dataset,zhao2021dataset} enhances this process, focusing on efficiency improvements through techniques like gradient matching. 
However, the previous approaches still have some open questions, such as how to design an effective matching strategy~\cite{sucholutsky2020softlabel, lu2023can}. 
In this work, we focus on designing a novel matching strategy for efficient graph condensation.


\textbf{Graph signal processing.}
Our work is also related to graph signal processing, which studies the analysis and processing of signals defined on graphs~\cite{ortega2018graph,Nica_2018}. Graph signal processing extends classical signal processing concepts like frequency, filtering, and sampling to graph signals~\cite{Shuman_2013,sandryhaila2013discrete}, which are functions that assign values to the nodes or edges of a graph, providing tools for feature extraction, denoising, compression, and learning on graph-structured data~\cite{hammond2009wavelets,Loukas_2015, S3GCL_ICML24}.

\textbf{Graph neural networks (GNNs).}
As the generalization of deep neural networks to graph data, Graph Neural Networks (GNNs) enhance the representation of individual nodes by utilizing information from their neighboring nodes\cite{kipf2016semi,you2019position,wang2024heterophilic, FGSSL_IJCAI23, Epi_Survey_KDD24}. Due to their powerful capability in handling graph-structured data, GNNs have achieved remarkable performance on various real-world tasks, such as social networks\cite{hoff2002latent,perozzi2014deepwalk, EARTH_Arxiv},physical\cite{bear2020learning,raffo2021shrec,mckay2022surge}, and chemical interactions\cite{he2016enumerating,battaglia2018relational,wu2020comprehensive,zhou2020graph}, and knowledge graphs\cite{noy2019industry,hogan2021knowledge,ji2021survey}.

\textbf{Graph coarsening \& Graph sparsification.}
Graph coarsening minimizes the size of graph data while preserving its basic properties~\cite{chen2022graph,kammer2022space,kelley2022parallel} by grouping similar nodes into supernodes. Graph Sparsification reduces the number of edges to make the graph sparser~\cite{li2022graph,yu2022graphfm,chen2022gc,wang2023brave, zhang2024two, zhang2024graph1}, as there are many
redundant relations in the graphs. Both two methods are based on the idea of coreset selection and data pruning~\cite{chen2012super,campbell2019automated,qin2023infobatch,zhang2024graph2}, aiming to remove less important information from the original graph.
However, these methods rely heavily on heuristic unsupervised techniques, resulting in poor generalization performance in downstream tasks. Moreover, they are unable to condense graphs with extremely low condensation ratios~\cite{zhangnavigating,hashemi2024comprehensive}. 

%% file: tables/NC_dataset.tex
\begin{table*}[htbp]
  \centering
  \caption{Dataset statistics(node classification). The first four are transductive datasets and the last two are inductive
datasets.}
  \label{tab:datasets}
  \fontsize{10}{11}\selectfont
   \begin{tabular}{c@{\hspace{2em}}c@{\hspace{2em}}c@{\hspace{2em}}c@{\hspace{2em}}c@{\hspace{2em}}c}
    \toprule
    Dataset & \#Nodes & \#Edges & \#Classes & \#Features & Training/Validation/Test\\
    \midrule
    Cora & 2,708 & 5,429 & 7 & 1,433 & 140/500/1000 \\
    Citeseer & 3,327 & 4,732 & 6 & 3,703 & 120/500/1000\\
Ogbn-arxiv & 169,343 & 1,166,243 & 40 & 128 & 90,941/29,799/48,603\\
Ogbn-xrt & 169,343 & 1,166,243 & 40 & 768 & 90,941/29,799/48,603\\
\midrule
Flickr & 89,250 & 899,756 & 7& 500 & 44,625/22312/22313\\
Reddit & 232,965 & 57,307,946 & 210 & 602 & 15,3932/23,699/55,334\\
    \bottomrule
    \end{tabular}%
  \label{tab:ncdatasets}%
\end{table*}

%% file: tables/GC_dataset.tex
\begin{table*}[htbp]
  \centering
  \caption{Dataset statistics(graph classification).}
  \fontsize{10}{13}\selectfont
   \begin{tabular}{c@{\hspace{2em}}c@{\hspace{2em}}c@{\hspace{2em}}c@{\hspace{2em}}c@{\hspace{2em}}c}
    \toprule
Dataset & Type & \#Clases & \#Graphs & \#Avg.Nodes & \#Avg. Edges\\
    \midrule
    CIFAR10 & Superpixel & 10 & 60,000 & 117.6 & 941.07 \\
    ogbg-molhiv & Molecule & 2 & 41,127 & 25.5 & 54.9\\
    ogbg-molbace & Molecule & 2 & 1,513 & 34.1 & 36.9\\
    ogbg-molbbbp & Molecule & 2 & 2,039 & 24.1 & 26.0\\
    MUTAG & Molecule & 2 & 188 & 17.93 & 19.79\\
    NCI1 & Molecule & 2 & 4,110 & 29.87 & 32.30\\
    \bottomrule
    \end{tabular}%
  \label{tab:gcdatasets}%
\end{table*}

%% file: tables/Corre.tex
  \begin{table*}[ht]
      \centering
        \caption{Spearman correlation coefficients of high-frequency area and gradient amplitude under different graph models and GNN architectures. Strong correlations were shown under almost all combinations}
      \begin{tabular}{c|cccc}
       \toprule
           & SGC & Cheby & APPNP & GCN \\
          \midrule
    Erdos-Renyi & 0.953 & 0.965 & 0.955 & 0.859\\
    \midrule
    Barabasi-Albert & 0.955 & 0.961 & 0.875 & 0.818 \\
    \midrule
    Watts-Strogatz  & 0.946 & 0.955 & 0.939 & 0.679\\
            \bottomrule
      \end{tabular}
    
      \label{tab:corre}
  \end{table*}

%% file: tables/cv.tex
\begin{table}[h!]
\centering
\caption{ MD denotes the method proposed in~\cite{jiang2022delving} using both cosine distance and Euclidean distance(equal weight), while MDA is the improved version of MD, all results in the table are averaged over three repetitions for accuracy and consistency.}
\begin{tabular}{cccccc}
\hline
\textbf{Dataset} & \textbf{IPC} & \textbf{MD} & \textbf{MD+CTRL} & \textbf{MDA} & \textbf{MDA+CTRL} \\
\hline
FashionMNIST & 10 & 85.4$\pm$0.3 & 85.4$\pm$0.3 & 84.2$\pm$0.3 & 84.7$\pm$0.1 \\
FashionMNIST & 50 & 87.4$\pm$0.2 & 87.6$\pm$0.2 & 87.9$\pm$0.2 & 87.9$\pm$0.2 \\
SVHN & 10 & 75.9$\pm$0.7 & 75.9$\pm$0.4 & 75.9$\pm$0.7 & 76.3$\pm$0.4 \\
SVHN & 50 & 82.9$\pm$0.2 & 83.0$\pm$0.1 & 83.2$\pm$0.3 & 83.4$\pm$0.2 \\
CIFAR-10 & 1 & 30.0$\pm$0.6 & 31.0$\pm$0.4 & - & - \\
CIFAR-10 & 10 & 49.5$\pm$0.5 & 49.5$\pm$0.3 & 50.2$\pm$0.6 & 51.3$\pm$0.1 \\
\hline
\end{tabular}
\label{table:results}
\end{table}

%% file: tables/cross_sfgc.tex
\begin{table*}[ht]
\centering
\tiny
\setlength{\extrarowheight}{0.8pt} 
\caption{Performance across different GNN architectures. \colorbox{orange!30}{\makebox(14,6){Avg.}} and \colorbox{red!30}{\makebox(14,6){Std.}}: the average performance and the standard deviation of the results of all architectures, \colorbox{cyan!20}{\makebox(20,6){$\Delta$(\%)}} denotes the improvements upon the DC-Graph. We mark the best performance by \textbf{bold}.}
\label{tab:cross_sfgc}
\resizebox{0.9\textwidth}{!}{%
\begin{tabular}{cccccccccccccc}
\cline{1-12}
                          &         & \multicolumn{7}{c}{Architectures}                & \multicolumn{3}{c}{Statistics}                                                                                        &  &  \\ \cline{3-12}
\multirow{-2}{*}{Datasets} & \multirow{-2}{*}{Methods} & MLP  & GAT  & APPNP & Cheby & GCN & SAGE & SGC & \cellcolor[HTML]{FFE6CC}Avg.          & \cellcolor[HTML]{FFB2B2}Std.          & \cellcolor[HTML]{C6EFFC}$\Delta$(\%)  &  &  \\ \cline{1-9}

                                                      & DC-Graph & 66.2 & - &66.4 &64.9 &66.2 &65.9 &69.6 &  \cellcolor[HTML]{FFE6CC}66.5          & \cellcolor[HTML]{FFB2B2}1.5         & \cellcolor[HTML]{C6EFFC}-             &  &  \\

                          & GCond & 63.9 &55.4 &69.6 &68.3 &70.5 &66.2 &70.3 & \cellcolor[HTML]{FFE6CC}66.3         & \cellcolor[HTML]{FFB2B2}5.0         & $\downarrow$ \cellcolor[HTML]{C6EFFC}0.2          &  &  \\
                           & SFGC & 71.3& 72.1 &70.5 &71.8& 71.6 &71.7& 71.8&  \cellcolor[HTML]{FFE6CC}71.5          & \cellcolor[HTML]{FFB2B2}0.5          & $\uparrow$ \cellcolor[HTML]{C6EFFC}5.0         &  &  \\

\multirow{-5}{*}{\centering Citeseer}
& CTRL & 71.6 & 72.4 & 71.2 & 71.4  &71.8 & 72.1 & 72.1 & \cellcolor[HTML]{FFE6CC}\textbf{71.8} & \cellcolor[HTML]{FFB2B2}0.4 & $\uparrow$ \cellcolor[HTML]{C6EFFC}\textbf{5.3}  &  &  \\ \cline{1-9}
\multirow{-5}{*}{\centering (Ratio = 1.80\%)}

                                                          & DC-Graph & 67.2 & - &67.1 &67.7 &67.9 &66.2 &72.8 &  \cellcolor[HTML]{FFE6CC}68.1          & \cellcolor[HTML]{FFB2B2}2.1          & \cellcolor[HTML]{C6EFFC}-             &  &  \\

                          & GCond & 73.1 &66.2 &78.5 &76.0 &80.1 &78.2 &79.3 & \cellcolor[HTML]{FFE6CC}75.9          & \cellcolor[HTML]{FFB2B2}4.5         &  $\uparrow$ \cellcolor[HTML]{C6EFFC}7.8         &  &  \\
                           & SFGC & 81.1& 80.8 &78.8 &79.0& 81.1 &81.9& 79.1&  \cellcolor[HTML]{FFE6CC}80.3         & \cellcolor[HTML]{FFB2B2}1.2          & $\uparrow$ \cellcolor[HTML]{C6EFFC}12.2           &  &  \\

\multirow{-5}{*}{Cora}  & CTRL & 81.1 & 81.0 & 78.7 & 79.2  &81.2 & 82.0 & 80.1 & \cellcolor[HTML]{FFE6CC}\textbf{80.5} & \cellcolor[HTML]{FFB2B2}1.1 & $\uparrow$ \cellcolor[HTML]{C6EFFC}\textbf{12.4}  &  &  \\ \cline{1-9}
\multirow{-5}{*}{\centering (Ratio = 2.60\%)}

                          & DC-Graph & 59.9& -& 60.0& 55.7 &59.8 &60.0 & 60.4 & \cellcolor[HTML]{FFE6CC}59.3          & \cellcolor[HTML]{FFB2B2}1.6           & \cellcolor[HTML]{C6EFFC}-             &  &  \\

                          & GCond & 62.2& 60.0 &63.4 &54.9& 63.2 &62.6& 63.7&  \cellcolor[HTML]{FFE6CC}61.4          & \cellcolor[HTML]{FFB2B2}2.9          &  $\uparrow$ \cellcolor[HTML]{C6EFFC}2.1            &  &  \\
                           & SFGC & 65.1& 65.7 &63.9 &60.7& 65.1 &64.8& 64.8&  \cellcolor[HTML]{FFE6CC}64.3          & \cellcolor[HTML]{FFB2B2}1.6         & $\uparrow$ \cellcolor[HTML]{C6EFFC}5.0           &  &  \\
           
\multirow{-5}{*}{Ogbn-arxiv}   & CTRL & 65.2 & 66.0 &63.6 & 61.0 &65.3 & 64.9 & 64.8  & \cellcolor[HTML]{FFE6CC}\textbf{64.4} & \cellcolor[HTML]{FFB2B2}1.5  & $\uparrow$ \cellcolor[HTML]{C6EFFC}\textbf{5.1} &  &  \\ \cline{1-9}
\multirow{-5}{*}{\centering (Ratio = 0.25\%)}

                          & DC-Graph &43.1 & -& 45.7 &43.8 &45.9 &45.8 &45.6  & \cellcolor[HTML]{FFE6CC}45.0        & \cellcolor[HTML]{FFB2B2}1.1         & \cellcolor[HTML]{C6EFFC}-             &  &  \\
                           & GCond & 44.8 & 40.1 & 45.9 & 42.8 & 47.1& 46.2& 46.1 & \cellcolor[HTML]{FFE6CC}44.7         & \cellcolor[HTML]{FFB2B2}2.3           & $\downarrow$ \cellcolor[HTML]{C6EFFC}0.3            &  &  \\
                           & SFGC & 47.1 & 45.3 & 40.7 & 45.4 & 47.1& 47.0& 42.5 & \cellcolor[HTML]{FFE6CC}45.0        & \cellcolor[HTML]{FFB2B2}2.3           & \cellcolor[HTML]{C6EFFC}-           &  &  \\
\multirow{-5}{*}{Flickr}     & CTRL & 47.1 & 45.6 & 40.9 & 45.3 & 47.2 &47.1 &43.2 & \cellcolor[HTML]{FFE6CC}\textbf{45.2} & \cellcolor[HTML]{FFB2B2}2.2  & $\uparrow$ \cellcolor[HTML]{C6EFFC}\textbf{0.2}  &  &  \\ \cline{1-9}
\multirow{-5}{*}{\centering (Ratio = 0.50\%)}

                           & DC-Graph & 50.3 & - & 81.2 & 77.5 & 89.5 & 89.7 & 90.5 & \cellcolor[HTML]{FFE6CC}79.8         & \cellcolor[HTML]{FFB2B2}14.0          & \cellcolor[HTML]{C6EFFC}-             &  &  \\
                           & GCond & 42.5 & 60.2 & 87.8 & 75.5 & 89.4 & 89.1 & 89.6 & \cellcolor[HTML]{FFE6CC}76.3          & \cellcolor[HTML]{FFB2B2}17.1          &  $\downarrow$ \cellcolor[HTML]{C6EFFC}3.5           &  &  \\
                          & SFGC & 89.5 & 87.1 & 88.3 & 82.8 & 89.7 & 90.3 & 89.5 & \cellcolor[HTML]{FFE6CC}88.2          & \cellcolor[HTML]{FFB2B2}2.4           & $\uparrow$ \cellcolor[HTML]{C6EFFC}8.4           &  &  \\
\multirow{-5}{*}{Reddit} & CTRL & 89.7 & 87.6 & 88.5 & 83.0 & 90.0 & 90.5 & 89.8 & \cellcolor[HTML]{FFE6CC}\textbf{88.4}  & \cellcolor[HTML]{FFB2B2}2.4 & \cellcolor[HTML]{C6EFFC}$\uparrow$ \textbf{8.6} &  &  \\ \cline{1-12}
\multirow{-5}{*}{\centering (Ratio = 0.10\%)}
                          &         &      &      &       &       &      &      &      &                                       &                                       &                                       &  & 
\end{tabular}%
}
\vspace{-2.5em}
\end{table*}

%% file: tables/time.tex
\begin{table*}[ht]
  \begin{minipage}[t]{0.45\textwidth}
  \centering
  \caption{Runing time on Citeseer, Cora and Ogbn-arxiv for 50 epochs.}
  \small
   \begin{tabular}{cccc}
    \toprule
    Dataset & \textit{r} &GCond & CTRL\\
    \midrule
    \multirow{3}{*}{Citeseer} & 0.9\% & 68.7$\pm2.4$s & 71.3$\pm 2.5$s \\
    & 1.8\% & 70.2$\pm 2.5$s & 73.7$\pm 2.6$s \\
     & 3.6\% & 78.1$\pm 2.3$s & 88.6$\pm 2.1$s \\
    \midrule
    \multirow{3}{*}{Cora} & 1.3\%  & 76.4$\pm 3.3$s & 76.8$\pm 2.8$s \\
    & 2.6\% & 77.7$\pm 2.2$s & 78.7$\pm 3.4$s \\
     & 5.2\%  & 85.3$\pm 3.4$s & 89.2$\pm 2.5$s \\
    \midrule
    \multirow{3}{*}{Ogbn-arxiv} & 0.1\% & 939.3$\pm 4.9$s & 967.6$\pm 5.8$s\\
    & 0.5\%  &1008.4$\pm 3.1$s & 1033.4$\pm 4.2$s \\
     & 1.0\%  &1061.3$\pm 2.9$s & 1087.6$\pm 1.8$s\\
    \bottomrule
    \end{tabular}%
  \label{tab:running time of NC}%
  \end{minipage}%
  \hfill
  \begin{minipage}[t]{0.45\textwidth} 
  \centering
  \caption{Runing time on Ogbg-molbace, molbbbp and molhiv for 100 epochs.}
  \small
   \begin{tabular}{cccc}
    \toprule
    Dataset & \textit{r} & Doscond & CTRL\\
    \midrule
    \multirow{3}{*}{Ogbg-molbace} & 0.2\% & 18.1$\pm1.4$s & 21.3$\pm 1.5$s \\
    & 1.7\% & 23.1$\pm 1.5$s & 27.4$\pm 1.6$s \\
     & 8.3\% & 26.6$\pm 1.3$s & 29.5$\pm 1.1$s \\
    \midrule
    \multirow{3}{*}{Ogbg-molbbbp} & 0.1\%  & 17.2$\pm 1.3$s & 18.9$\pm 1.8$s \\
    & 1.2\% & 20.2$\pm 1.2$s & 23.4$\pm 1.4$s \\
     & 6.1\%  & 22.1$\pm 1.4$s & 24.9$\pm 1.5$s \\
    \midrule
    \multirow{3}{*}{Ogbg-molhiv} & 0.1\% & 39.6$\pm 1.9$s & 42.3$\pm 1.8$s\\
    & 0.5\%  & 40.2$\pm 1.1$s & 43.1$\pm 2.2$s \\
     & 1.0\% & 40.8$\pm 1.9$s & 43.9$\pm 1.8$s\\
    \bottomrule
    \end{tabular}%
  \label{tab:running time of GC}%
  \end{minipage}
  \vfill
\end{table*}

%% file: neurips_2024.bbl
\begin{thebibliography}{100}
\providecommand{\url}[1]{#1}
\csname url@samestyle\endcsname
\providecommand{\newblock}{\relax}
\providecommand{\bibinfo}[2]{#2}
\providecommand{\BIBentrySTDinterwordspacing}{\spaceskip=0pt\relax}
\providecommand{\BIBentryALTinterwordstretchfactor}{4}
\providecommand{\BIBentryALTinterwordspacing}{\spaceskip=\fontdimen2\font plus
\BIBentryALTinterwordstretchfactor\fontdimen3\font minus \fontdimen4\font\relax}
\providecommand{\BIBforeignlanguage}[2]{{%
\expandafter\ifx\csname l@#1\endcsname\relax
\typeout{** WARNING: IEEEtran.bst: No hyphenation pattern has been}%
\typeout{** loaded for the language `#1'. Using the pattern for}%
\typeout{** the default language instead.}%
\else
\language=\csname l@#1\endcsname
\fi
#2}}
\providecommand{\BIBdecl}{\relax}
\BIBdecl

\bibitem{hoff2002latent}
P.~D. Hoff, A.~E. Raftery, and M.~S. Handcock, ``Latent space approaches to social network analysis,'' \emph{Journal of the american Statistical association}, vol.~97, no. 460, pp. 1090--1098, 2002.

\bibitem{sahili2023spatiotemporal}
Z.~A. Sahili and M.~Awad, ``Spatio-temporal graph neural networks: A survey,'' 2023.

\bibitem{hashemi2024comprehensive}
M.~Hashemi, S.~Gong, J.~Ni, W.~Fan, B.~A. Prakash, and W.~Jin, ``A comprehensive survey on graph reduction: Sparsification, coarsening, and condensation,'' \emph{arXiv preprint}, 2024.

\bibitem{liu2022dataset}
S.~Liu, K.~Wang, X.~Yang, J.~Ye, and X.~Wang, ``Dataset distillation via factorization,'' 2022.

\bibitem{zhangnavigating}
Y.~Zhang, T.~Zhang, K.~Wang, Z.~Guo, Y.~Liang, X.~Bresson, W.~Jin, and Y.~You, ``Navigating complexity: Toward lossless graph condensation via expanding window matching,'' in \emph{Forty-first International Conference on Machine Learning}.

\bibitem{lu2023can}
Y.~Lu, X.~Chen, Y.~Zhang, J.~Gu, T.~Zhang, Y.~Zhang, X.~Yang, Q.~Xuan, K.~Wang, and Y.~You, ``Can pre-trained models assist in dataset distillation?'' \emph{arXiv preprint arXiv:2310.03295}, 2023.

\bibitem{wang2023dim}
K.~Wang, J.~Gu, D.~Zhou, Z.~Zhu, W.~Jiang, and Y.~You, ``Dim: Distilling dataset into generative model,'' 2023.

\bibitem{zhou2023dataset}
D.~Zhou, K.~Wang, J.~Gu, X.~Peng, D.~Lian, Y.~Zhang, Y.~You, and J.~Feng, ``Dataset quantization,'' 2023.

\bibitem{qin2023infobatch}
Z.~Qin, K.~Wang, Z.~Zheng, J.~Gu, X.~Peng, Z.~Xu, D.~Zhou, L.~Shang, B.~Sun, X.~Xie, and Y.~You, ``Infobatch: Lossless training speed up by unbiased dynamic data pruning,'' 2023.

\bibitem{li2021data}
Y.~Li and W.~Li, ``Data distillation for text classification,'' 2021.

\bibitem{wang2023brave}
K.~Wang, Y.~Liang, X.~Li, G.~Li, B.~Ghanem, R.~Zimmermann, H.~Yi, Y.~Zhang, Y.~Wang \emph{et~al.}, ``Brave the wind and the waves: Discovering robust and generalizable graph lottery tickets,'' \emph{IEEE Transactions on Pattern Analysis and Machine Intelligence}, 2023.

\bibitem{wang2022searching}
K.~Wang, Y.~Liang, P.~Wang, X.~Wang, P.~Gu, J.~Fang, and Y.~Wang, ``Searching lottery tickets in graph neural networks: A dual perspective,'' in \emph{The Eleventh International Conference on Learning Representations}, 2022.

\bibitem{Loukas_2015}
\BIBentryALTinterwordspacing
A.~Loukas, A.~Simonetto, and G.~Leus, ``Distributed autoregressive moving average graph filters,'' \emph{{IEEE} Signal Processing Letters}, vol.~22, no.~11, pp. 1931--1935, nov 2015. [Online]. Available: \url{https://doi.org/10.1109%2Flsp.2015.2448655}
\BIBentrySTDinterwordspacing

\bibitem{Shuman_2013}
\BIBentryALTinterwordspacing
D.~I. Shuman, S.~K. Narang, P.~Frossard, A.~Ortega, and P.~Vandergheynst, ``The emerging field of signal processing on graphs: Extending high-dimensional data analysis to networks and other irregular domains,'' \emph{{IEEE} Signal Processing Magazine}, vol.~30, no.~3, pp. 83--98, may 2013. [Online]. Available: \url{https://doi.org/10.1109%2Fmsp.2012.2235192}
\BIBentrySTDinterwordspacing

\bibitem{Leus_2023}
\BIBentryALTinterwordspacing
G.~Leus, A.~G. Marques, J.~M. Moura, A.~Ortega, and D.~I. Shuman, ``Graph signal processing: History, development, impact, and outlook,'' \emph{{IEEE} Signal Processing Magazine}, vol.~40, no.~4, pp. 49--60, jun 2023. [Online]. Available: \url{https://doi.org/10.1109%2Fmsp.2023.3262906}
\BIBentrySTDinterwordspacing

\bibitem{ortega2018graph}
A.~Ortega, P.~Frossard, J.~Kovačević, J.~M.~F. Moura, and P.~Vandergheynst, ``Graph signal processing: Overview, challenges and applications,'' 2018.

\bibitem{sandryhaila2013discrete}
A.~Sandryhaila and J.~M.~F. Moura, ``Discrete signal processing on graphs: Frequency analysis,'' 2013.

\bibitem{hammond2009wavelets}
D.~K. Hammond, P.~Vandergheynst, and R.~Gribonval, ``Wavelets on graphs via spectral graph theory,'' 2009.

\bibitem{Nica_2018}
\BIBentryALTinterwordspacing
B.~Nica, \emph{A Brief Introduction to Spectral Graph Theory}.\hskip 1em plus 0.5em minus 0.4em\relax {EMS} Press, may 2018. [Online]. Available: \url{https://doi.org/10.4171%2F188}
\BIBentrySTDinterwordspacing

\bibitem{sucholutsky2020softlabel}
I.~Sucholutsky and M.~Schonlau, ``Soft-label dataset distillation and text dataset distillation,'' 2020.

\bibitem{zhang2023iterative}
H.~Zhang, S.~Lin, W.~Liu, P.~Zhou, J.~Tang, X.~Liang, and E.~P. Xing, ``Iterative graph self-distillation,'' 2023.

\bibitem{du2023minimizing}
J.~Du, Y.~Jiang, V.~Y. Tan, J.~T. Zhou, and H.~Li, ``Minimizing the accumulated trajectory error to improve dataset distillation,'' in \emph{Proceedings of the IEEE/CVF Conference on Computer Vision and Pattern Recognition}, 2023, pp. 3749--3758.

\bibitem{jiang2022delving}
Z.~Jiang, J.~Gu, M.~Liu, and D.~Z. Pan, ``Delving into effective gradient matching for dataset condensation,'' 2022.

\bibitem{wu2023multimodal}
X.~Wu, Z.~Deng, and O.~Russakovsky, ``Multimodal dataset distillation for image-text retrieval,'' 2023.

\bibitem{Ram_rez_2021}
\BIBentryALTinterwordspacing
D.~Ram{\'{\i}}rez, A.~G. Marques, and S.~Segarra, ``Graph-signal reconstruction and blind deconvolution for structured inputs,'' \emph{Signal Processing}, vol. 188, p. 108180, nov 2021. [Online]. Available: \url{https://doi.org/10.1016%2Fj.sigpro.2021.108180}
\BIBentrySTDinterwordspacing

\bibitem{dong2022privacy}
T.~Dong, B.~Zhao, and L.~Lyu, ``Privacy for free: How does dataset condensation help privacy?'' in \emph{International Conference on Machine Learning}.\hskip 1em plus 0.5em minus 0.4em\relax PMLR, 2022, pp. 5378--5396.

\bibitem{xiong2023feddm}
Y.~Xiong, R.~Wang, M.~Cheng, F.~Yu, and C.-J. Hsieh, ``Feddm: Iterative distribution matching for communication-efficient federated learning,'' in \emph{Proceedings of the IEEE/CVF Conference on Computer Vision and Pattern Recognition}, 2023, pp. 16\,323--16\,332.

\bibitem{xu2023splitgnn}
X.~Xu, L.~Lyu, Y.~Dong, Y.~Lu, W.~Wang, and H.~Jin, ``Splitgnn: Splitting gnn for node classification with heterogeneous attention,'' 2023.

\bibitem{Veli_kovi__2023}
\BIBentryALTinterwordspacing
P.~Veli{\v{c}}kovi{\'{c}}, ``Everything is connected: Graph neural networks,'' \emph{Current Opinion in Structural Biology}, vol.~79, p. 102538, apr 2023. [Online]. Available: \url{https://doi.org/10.1016%2Fj.sbi.2023.102538}
\BIBentrySTDinterwordspacing

\bibitem{borgatti1992notions}
S.~P. Borgatti and M.~G. Everett, ``Notions of position in social network analysis,'' \emph{Sociological methodology}, pp. 1--35, 1992.

\bibitem{perozzi2014deepwalk}
B.~Perozzi, R.~Al-Rfou, and S.~Skiena, ``Deepwalk: Online learning of social representations,'' in \emph{Proceedings of the 20th ACM SIGKDD international conference on Knowledge discovery and data mining}, 2014, pp. 701--710.

\bibitem{ebel2002coevolutionary}
H.~Ebel and S.~Bornholdt, ``Coevolutionary games on networks,'' \emph{Physical Review E}, vol.~66, no.~5, p. 056118, 2002.

\bibitem{bear2020learning}
D.~Bear, C.~Fan, D.~Mrowca, Y.~Li, S.~Alter, A.~Nayebi, J.~Schwartz, L.~F. Fei-Fei, J.~Wu, J.~Tenenbaum \emph{et~al.}, ``Learning physical graph representations from visual scenes,'' \emph{Advances in Neural Information Processing Systems}, vol.~33, pp. 6027--6039, 2020.

\bibitem{raffo2021shrec}
A.~Raffo, U.~Fugacci, S.~Biasotti, W.~Rocchia, Y.~Liu, E.~Otu, R.~Zwiggelaar, D.~Hunter, E.~I. Zacharaki, E.~Psatha \emph{et~al.}, ``Shrec 2021: Retrieval and classification of protein surfaces equipped with physical and chemical properties,'' \emph{Computers \& Graphics}, vol.~99, pp. 1--21, 2021.

\bibitem{mckay2022surge}
B.~D. McKay, M.~A. Yirik, and C.~Steinbeck, ``Surge: a fast open-source chemical graph generator,'' \emph{Journal of Cheminformatics}, vol.~14, no.~1, p.~24, 2022.

\bibitem{daller2018local}
{\'E}.~Daller, S.~Bougleux, L.~Brun, and O.~L{\'e}zoray, ``Local patterns and supergraph for chemical graph classification with convolutional networks,'' in \emph{Joint IAPR International Workshops on Statistical Techniques in Pattern Recognition (SPR) and Structural and Syntactic Pattern Recognition (SSPR)}.\hskip 1em plus 0.5em minus 0.4em\relax Springer, 2018, pp. 97--106.

\bibitem{battaglia2018relational}
P.~W. Battaglia, J.~B. Hamrick, V.~Bapst, A.~Sanchez-Gonzalez, V.~Zambaldi, M.~Malinowski, A.~Tacchetti, D.~Raposo, A.~Santoro, R.~Faulkner \emph{et~al.}, ``Relational inductive biases, deep learning, and graph networks,'' \emph{arXiv preprint arXiv:1806.01261}, 2018.

\bibitem{wu2020comprehensive}
Z.~Wu, S.~Pan, F.~Chen, G.~Long, C.~Zhang, and S.~Y. Philip, ``A comprehensive survey on graph neural networks,'' \emph{IEEE transactions on neural networks and learning systems}, vol.~32, no.~1, pp. 4--24, 2020.

\bibitem{zhou2020graph}
J.~Zhou, G.~Cui, S.~Hu, Z.~Zhang, C.~Yang, Z.~Liu, L.~Wang, C.~Li, and M.~Sun, ``Graph neural networks: A review of methods and applications,'' \emph{AI open}, vol.~1, pp. 57--81, 2020.

\bibitem{wale2010trends}
N.~Wale, X.~Ning, and G.~Karypis, ``Trends in chemical graph data mining,'' \emph{Managing and mining graph data}, pp. 581--606, 2010.

\bibitem{he2016enumerating}
F.~He, A.~Hanai, H.~Nagamochi, and T.~Akutsu, ``Enumerating naphthalene isomers of tree-like chemical graphs,'' in \emph{International Conference on Bioinformatics Models, Methods and Algorithms}, vol.~4.\hskip 1em plus 0.5em minus 0.4em\relax SCITEPRESS, 2016, pp. 258--265.

\bibitem{hogan2021knowledge}
A.~Hogan, E.~Blomqvist, M.~Cochez, C.~d’Amato, G.~D. Melo, C.~Gutierrez, S.~Kirrane, J.~E.~L. Gayo, R.~Navigli, S.~Neumaier \emph{et~al.}, ``Knowledge graphs,'' \emph{ACM Computing Surveys (Csur)}, vol.~54, no.~4, pp. 1--37, 2021.

\bibitem{ji2021survey}
S.~Ji, S.~Pan, E.~Cambria, P.~Marttinen, and S.~Y. Philip, ``A survey on knowledge graphs: Representation, acquisition, and applications,'' \emph{IEEE transactions on neural networks and learning systems}, vol.~33, no.~2, pp. 494--514, 2021.

\bibitem{nickel2015review}
M.~Nickel, K.~Murphy, V.~Tresp, and E.~Gabrilovich, ``A review of relational machine learning for knowledge graphs,'' \emph{Proceedings of the IEEE}, vol. 104, no.~1, pp. 11--33, 2015.

\bibitem{noy2019industry}
N.~Noy, Y.~Gao, A.~Jain, A.~Narayanan, A.~Patterson, and J.~Taylor, ``Industry-scale knowledge graphs: Lessons and challenges: Five diverse technology companies show how it’s done,'' \emph{Queue}, vol.~17, no.~2, pp. 48--75, 2019.

\bibitem{smith2002constructing}
L.~D. Smith, L.~A. Best, D.~A. Stubbs, A.~B. Archibald, and R.~Roberson-Nay, ``Constructing knowledge: The role of graphs and tables in hard and soft psychology.'' \emph{American Psychologist}, vol.~57, no.~10, p. 749, 2002.

\bibitem{xu2018powerful}
K.~Xu, W.~Hu, J.~Leskovec, and S.~Jegelka, ``How powerful are graph neural networks?'' \emph{arXiv preprint arXiv:1810.00826}, 2018.

\bibitem{scarselli2008graph}
F.~Scarselli, M.~Gori, A.~C. Tsoi, M.~Hagenbuchner, and G.~Monfardini, ``The graph neural network model,'' \emph{IEEE transactions on neural networks}, vol.~20, no.~1, pp. 61--80, 2008.

\bibitem{you2019position}
J.~You, R.~Ying, and J.~Leskovec, ``Position-aware graph neural networks,'' in \emph{International conference on machine learning}.\hskip 1em plus 0.5em minus 0.4em\relax PMLR, 2019, pp. 7134--7143.

\bibitem{shlomi2020graph}
J.~Shlomi, P.~Battaglia, and J.-R. Vlimant, ``Graph neural networks in particle physics,'' \emph{Machine Learning: Science and Technology}, vol.~2, no.~2, p. 021001, 2020.

\bibitem{wu2022graph}
L.~Wu, P.~Cui, J.~Pei, L.~Zhao, and X.~Guo, ``Graph neural networks: foundation, frontiers and applications,'' in \emph{Proceedings of the 28th ACM SIGKDD Conference on Knowledge Discovery and Data Mining}, 2022, pp. 4840--4841.

\bibitem{li2019learning}
B.~Li and D.~Pi, ``Learning deep neural networks for node classification,'' \emph{Expert Systems with Applications}, vol. 137, pp. 324--334, 2019.

\bibitem{xiao2022graph}
S.~Xiao, S.~Wang, Y.~Dai, and W.~Guo, ``Graph neural networks in node classification: survey and evaluation,'' \emph{Machine Vision and Applications}, vol.~33, pp. 1--19, 2022.

\bibitem{chen2022gc}
J.~Chen, X.~Wang, and X.~Xu, ``Gc-lstm: Graph convolution embedded lstm for dynamic network link prediction,'' \emph{Applied Intelligence}, pp. 1--16, 2022.

\bibitem{rossi2022explaining}
A.~Rossi, D.~Firmani, P.~Merialdo, and T.~Teofili, ``Explaining link prediction systems based on knowledge graph embeddings,'' in \emph{Proceedings of the 2022 international conference on management of data}, 2022, pp. 2062--2075.

\bibitem{van2022declarative}
D.~Van~Assche, T.~Delva, G.~Haesendonck, P.~Heyvaert, B.~De~Meester, and A.~Dimou, ``Declarative rdf graph generation from heterogeneous (semi-) structured data: A systematic literature review,'' \emph{Journal of Web Semantics}, p. 100753, 2022.

\bibitem{vignac2022digress}
C.~Vignac, I.~Krawczuk, A.~Siraudin, B.~Wang, V.~Cevher, and P.~Frossard, ``Digress: Discrete denoising diffusion for graph generation,'' \emph{arXiv preprint arXiv:2209.14734}, 2022.

\bibitem{polisetty2023gsplit}
S.~Polisetty, J.~Liu, K.~Falus, Y.~R. Fung, S.-H. Lim, H.~Guan, and M.~Serafini, ``Gsplit: Scaling graph neural network training on large graphs via split-parallelism,'' \emph{arXiv preprint arXiv:2303.13775}, 2023.

\bibitem{wang2022adagcl}
Y.~Wang, K.~Zhou, R.~Miao, N.~Liu, and X.~Wang, ``Adagcl: Adaptive subgraph contrastive learning to generalize large-scale graph training,'' in \emph{Proceedings of the 31st ACM International Conference on Information \& Knowledge Management}, 2022, pp. 2046--2055.

\bibitem{yu2022graphfm}
H.~Yu, L.~Wang, B.~Wang, M.~Liu, T.~Yang, and S.~Ji, ``Graphfm: Improving large-scale gnn training via feature momentum,'' in \emph{International Conference on Machine Learning}.\hskip 1em plus 0.5em minus 0.4em\relax PMLR, 2022, pp. 25\,684--25\,701.

\bibitem{duan2022comprehensive}
K.~Duan, Z.~Liu, P.~Wang, W.~Zheng, K.~Zhou, T.~Chen, X.~Hu, and Z.~Wang, ``A comprehensive study on large-scale graph training: Benchmarking and rethinking,'' \emph{Advances in Neural Information Processing Systems}, vol.~35, pp. 5376--5389, 2022.

\bibitem{li2022graph}
J.~Li, T.~Zhang, H.~Tian, S.~Jin, M.~Fardad, and R.~Zafarani, ``Graph sparsification with graph convolutional networks,'' \emph{International Journal of Data Science and Analytics}, pp. 1--14, 2022.

\bibitem{yu2022principle}
S.~Yu, F.~Alesiani, W.~Yin, R.~Jenssen, and J.~C. Principe, ``Principle of relevant information for graph sparsification,'' in \emph{Uncertainty in Artificial Intelligence}.\hskip 1em plus 0.5em minus 0.4em\relax PMLR, 2022, pp. 2331--2341.

\bibitem{chen2022weighted}
Y.~Chen, S.~Khanna, and H.~Li, ``On weighted graph sparsification by linear sketching,'' in \emph{2022 IEEE 63rd Annual Symposium on Foundations of Computer Science (FOCS)}.\hskip 1em plus 0.5em minus 0.4em\relax IEEE, 2022, pp. 474--485.

\bibitem{chen2022graph}
J.~Chen, Y.~Saad, and Z.~Zhang, ``Graph coarsening: from scientific computing to machine learning,'' \emph{SeMA Journal}, pp. 1--37, 2022.

\bibitem{kelley2022parallel}
B.~Kelley and S.~Rajamanickam, ``Parallel, portable algorithms for distance-2 maximal independent set and graph coarsening,'' in \emph{2022 IEEE International Parallel and Distributed Processing Symposium (IPDPS)}.\hskip 1em plus 0.5em minus 0.4em\relax IEEE, 2022, pp. 280--290.

\bibitem{kammer2022space}
F.~Kammer and J.~Meintrup, ``Space-efficient graph coarsening with applications to succinct planar encodings,'' \emph{arXiv preprint arXiv:2205.06128}, 2022.

\bibitem{jin2022graph}
W.~Jin, L.~Zhao, S.~Zhang, Y.~Liu, J.~Tang, and N.~Shah, ``Graph condensation for graph neural networks,'' 2022.

\bibitem{cai2021graph}
C.~Cai, D.~Wang, and Y.~Wang, ``Graph coarsening with neural networks,'' \emph{arXiv preprint arXiv:2102.01350}, 2021.

\bibitem{9737295}
G.~Li, A.~Zhang, Q.~Zhang, D.~Wu, and C.~Zhan, ``Pearson correlation coefficient-based performance enhancement of broad learning system for stock price prediction,'' \emph{IEEE Transactions on Circuits and Systems II: Express Briefs}, vol.~69, no.~5, pp. 2413--2417, 2022.

\bibitem{jin2022condensing}
W.~Jin, X.~Tang, H.~Jiang, Z.~Li, D.~Zhang, J.~Tang, and B.~Yin, ``Condensing graphs via one-step gradient matching,'' in \emph{Proceedings of the 28th ACM SIGKDD Conference on Knowledge Discovery and Data Mining}, 2022, pp. 720--730.

\bibitem{zheng2023structure}
X.~Zheng, M.~Zhang, C.~Chen, Q.~V.~H. Nguyen, X.~Zhu, and S.~Pan, ``Structure-free graph condensation: From large-scale graphs to condensed graph-free data,'' \emph{arXiv preprint arXiv:2306.02664}, 2023.

\bibitem{jiang2023delving}
Z.~Jiang, J.~Gu, M.~Liu, and D.~Z. Pan, ``Delving into effective gradient matching for dataset condensation,'' in \emph{2023 IEEE International Conference on Omni-layer Intelligent Systems (COINS)}.\hskip 1em plus 0.5em minus 0.4em\relax IEEE, 2023, pp. 1--6.

\bibitem{amari1998natural}
S.-I. Amari, ``Natural gradient works efficiently in learning,'' \emph{Neural computation}, vol.~10, no.~2, pp. 251--276, 1998.

\bibitem{pascanu2013revisiting}
R.~Pascanu and Y.~Bengio, ``Revisiting natural gradient for deep networks,'' \emph{arXiv preprint arXiv:1301.3584}, 2013.

\bibitem{martens2020new}
J.~Martens, ``New insights and perspectives on the natural gradient method,'' \emph{The Journal of Machine Learning Research}, vol.~21, no.~1, pp. 5776--5851, 2020.

\bibitem{masset2022natural}
P.~Masset, J.~Zavatone-Veth, J.~P. Connor, V.~Murthy, and C.~Pehlevan, ``Natural gradient enables fast sampling in spiking neural networks,'' \emph{Advances in Neural Information Processing Systems}, vol.~35, pp. 22\,018--22\,034, 2022.

\bibitem{rumelhart1986learning}
D.~E. Rumelhart, G.~E. Hinton, and R.~J. Williams, ``Learning representations by back-propagating errors,'' \emph{nature}, vol. 323, no. 6088, pp. 533--536, 1986.

\bibitem{ruder2016overview}
S.~Ruder, ``An overview of gradient descent optimization algorithms,'' \emph{arXiv preprint arXiv:1609.04747}, 2016.

\bibitem{fukuda2011convergence}
E.~H. Fukuda and L.~G. Drummond, ``On the convergence of the projected gradient method for vector optimization,'' \emph{Optimization}, vol.~60, no. 8-9, pp. 1009--1021, 2011.

\bibitem{xu1998snakes}
C.~Xu and J.~L. Prince, ``Snakes, shapes, and gradient vector flow,'' \emph{IEEE Transactions on image processing}, vol.~7, no.~3, pp. 359--369, 1998.

\bibitem{cantrijn1992gradient}
F.~Cantrijn, M.~de~Le{\'o}n, and E.~Lacomba, ``Gradient vector fields on cosymplectic manifolds,'' \emph{Journal of Physics A: Mathematical and General}, vol.~25, no.~1, p. 175, 1992.

\bibitem{rumelhart1985learning}
D.~E. Rumelhart, G.~E. Hinton, R.~J. Williams \emph{et~al.}, ``Learning internal representations by error propagation,'' 1985.

\bibitem{sutskever2013importance}
I.~Sutskever, J.~Martens, G.~Dahl, and G.~Hinton, ``On the importance of initialization and momentum in deep learning,'' in \emph{International conference on machine learning}.\hskip 1em plus 0.5em minus 0.4em\relax PMLR, 2013, pp. 1139--1147.

\bibitem{huang2020improving}
X.~S. Huang, F.~Perez, J.~Ba, and M.~Volkovs, ``Improving transformer optimization through better initialization,'' in \emph{International Conference on Machine Learning}.\hskip 1em plus 0.5em minus 0.4em\relax PMLR, 2020, pp. 4475--4483.

\bibitem{he2004initialization}
J.~He, M.~Lan, C.-L. Tan, S.-Y. Sung, and H.-B. Low, ``Initialization of cluster refinement algorithms: A review and comparative study,'' in \emph{2004 IEEE international joint conference on neural networks (IEEE Cat. No. 04CH37541)}, vol.~1.\hskip 1em plus 0.5em minus 0.4em\relax IEEE, 2004, pp. 297--302.

\bibitem{forgy1965cluster}
E.~W. Forgy, ``Cluster analysis of multivariate data: efficiency versus interpretability of classifications,'' \emph{biometrics}, vol.~21, pp. 768--769, 1965.

\bibitem{hartigan1979algorithm}
J.~A. Hartigan and M.~A. Wong, ``Algorithm as 136: A k-means clustering algorithm,'' \emph{Journal of the royal statistical society. series c (applied statistics)}, vol.~28, no.~1, pp. 100--108, 1979.

\bibitem{arthur2007k}
D.~Arthur and S.~Vassilvitskii, ``K-means++ the advantages of careful seeding,'' in \emph{Proceedings of the eighteenth annual ACM-SIAM symposium on Discrete algorithms}, 2007, pp. 1027--1035.

\bibitem{liu2023dream}
Y.~Liu, J.~Gu, K.~Wang, Z.~Zhu, W.~Jiang, and Y.~You, ``Dream: Efficient dataset distillation by representative matching,'' \emph{arXiv preprint arXiv:2302.14416}, 2023.

\bibitem{zheng2023auto}
X.~Zheng, M.~Zhang, C.~Chen, Q.~Zhang, C.~Zhou, and S.~Pan, ``Auto-heg: Automated graph neural network on heterophilic graphs,'' \emph{arXiv preprint arXiv:2302.12357}, 2023.

\bibitem{rosasco2021distilled}
A.~Rosasco, A.~Carta, A.~Cossu, V.~Lomonaco, and D.~Bacciu, ``Distilled replay: Overcoming forgetting through synthetic samples,'' in \emph{International Workshop on Continual Semi-Supervised Learning}.\hskip 1em plus 0.5em minus 0.4em\relax Springer, 2021, pp. 104--117.

\bibitem{gao2021graph}
Y.~Gao, H.~Yang, P.~Zhang, C.~Zhou, and Y.~Hu, ``Graph neural architecture search,'' in \emph{International joint conference on artificial intelligence}.\hskip 1em plus 0.5em minus 0.4em\relax International Joint Conference on Artificial Intelligence, 2021.

\bibitem{CaT}
Y.~Liu, R.~Qiu, and Z.~Huang, ``Cat: Balanced continual graph learning with graph condensation,'' vol. abs/2309.09455, 2023.

\bibitem{kipf2016semi}
T.~N. Kipf and M.~Welling, ``Semi-supervised classification with graph convolutional networks,'' \emph{arXiv preprint arXiv:1609.02907}, 2016.

\bibitem{wang2018dataset}
T.~Wang, J.-Y. Zhu, A.~Torralba, and A.~A. Efros, ``Dataset distillation,'' \emph{arXiv preprint arXiv:1811.10959}, 2018.

\bibitem{bohdal2020flexible}
O.~Bohdal, Y.~Yang, and T.~Hospedales, ``Flexible dataset distillation: Learn labels instead of images,'' \emph{arXiv preprint arXiv:2006.08572}, 2020.

\bibitem{nguyen2020dataset}
T.~Nguyen, Z.~Chen, and J.~Lee, ``Dataset meta-learning from kernel ridge-regression,'' \emph{arXiv preprint arXiv:2011.00050}, 2020.

\bibitem{zhao2020dataset}
B.~Zhao, K.~R. Mopuri, and H.~Bilen, ``Dataset condensation with gradient matching,'' \emph{arXiv preprint arXiv:2006.05929}, 2020.

\bibitem{zhao2021dataset}
B.~Zhao and H.~Bilen, ``Dataset condensation with differentiable siamese augmentation,'' in \emph{International Conference on Machine Learning}.\hskip 1em plus 0.5em minus 0.4em\relax PMLR, 2021, pp. 12\,674--12\,685.

\bibitem{chen2012super}
Y.~Chen, M.~Welling, and A.~Smola, ``Super-samples from kernel herding,'' \emph{arXiv preprint arXiv:1203.3472}, 2012.

\bibitem{campbell2019automated}
T.~Campbell and T.~Broderick, ``Automated scalable bayesian inference via hilbert coresets,'' \emph{The Journal of Machine Learning Research}, vol.~20, no.~1, pp. 551--588, 2019.

\bibitem{cheng2022graph}
C.~Cheng, Y.~Chen, J.~Y. Lee, and Q.~Sun, ``Graph fourier transforms on directed product graphs,'' \emph{arXiv preprint arXiv:2209.01336}, 2022.

\bibitem{nt2019revisiting}
H.~Nt and T.~Maehara, ``Revisiting graph neural networks: All we have is low-pass filters,'' \emph{arXiv preprint arXiv:1905.09550}, 2019.

\bibitem{chen2020bridging}
Z.~Chen, F.~Chen, L.~Zhang, T.~Ji, K.~Fu, L.~Zhao, F.~Chen, L.~Wu, C.~Aggarwal, and C.-T. Lu, ``Bridging the gap between spatial and spectral domains: A survey on graph neural networks,'' \emph{arXiv preprint arXiv:2002.11867}, 2020.

\bibitem{wu2023extracting}
L.~Wu, H.~Lin, Y.~Huang, T.~Fan, and S.~Z. Li, ``Extracting low-/high-frequency knowledge from graph neural networks and injecting it into mlps: An effective gnn-to-mlp distillation framework,'' \emph{arXiv preprint arXiv:2305.10758}, 2023.

\bibitem{bo2021beyond}
D.~Bo, X.~Wang, C.~Shi, and H.~Shen, ``Beyond low-frequency information in graph convolutional networks,'' in \emph{Proceedings of the AAAI Conference on Artificial Intelligence}, vol.~35, 2021, pp. 3950--3957.

\bibitem{wang2019demystifying}
\BIBentryALTinterwordspacing
Y.~Wang, Z.~Hu, Y.~Ye, and Y.~Sun, ``Demystifying graph neural network via graph filter assessment,'' 2019. [Online]. Available: \url{https://api.semanticscholar.org/CorpusID:213408668}
\BIBentrySTDinterwordspacing

\bibitem{liu2022graph}
M.~Liu, S.~Li, X.~Chen, and L.~Song, ``Graph condensation via receptive field distribution matching,'' \emph{arXiv preprint arXiv:2206.13697}, 2022.

\bibitem{welling2009herding}
M.~Welling, ``Herding dynamical weights to learn,'' in \emph{Proceedings of the 26th Annual International Conference on Machine Learning}, 2009, pp. 1121--1128.

\bibitem{sener2017active}
O.~Sener and S.~Savarese, ``Active learning for convolutional neural networks: A core-set approach,'' \emph{arXiv preprint arXiv:1708.00489}, 2017.

\bibitem{huang2021scaling}
Z.~Huang, S.~Zhang, C.~Xi, T.~Liu, and M.~Zhou, ``Scaling up graph neural networks via graph coarsening,'' in \emph{Proceedings of the 27th ACM SIGKDD conference on knowledge discovery \& data mining}, 2021, pp. 675--684.

\bibitem{hu2020open}
W.~Hu, M.~Fey, M.~Zitnik, Y.~Dong, H.~Ren, B.~Liu, M.~Catasta, and J.~Leskovec, ``Open graph benchmark: Datasets for machine learning on graphs,'' \emph{Advances in neural information processing systems}, vol.~33, pp. 22\,118--22\,133, 2020.

\bibitem{pmlr-v162-tang22b}
\BIBentryALTinterwordspacing
J.~Tang, J.~Li, Z.~Gao, and J.~Li, ``Rethinking graph neural networks for anomaly detection,'' in \emph{Proceedings of the 39th International Conference on Machine Learning}, ser. Proceedings of Machine Learning Research, K.~Chaudhuri, S.~Jegelka, L.~Song, C.~Szepesvari, G.~Niu, and S.~Sabato, Eds., vol. 162.\hskip 1em plus 0.5em minus 0.4em\relax PMLR, 17--23 Jul 2022, pp. 21\,076--21\,089. [Online]. Available: \url{https://proceedings.mlr.press/v162/tang22b.html}
\BIBentrySTDinterwordspacing

\bibitem{luo2022fast}
T.~Luo, Z.~Mo, and S.~J. Pan, ``Fast graph generation via spectral diffusion,'' 2022.

\bibitem{martinkus2022spectre}
K.~Martinkus, A.~Loukas, N.~Perraudin, and R.~Wattenhofer, ``Spectre: Spectral conditioning helps to overcome the expressivity limits of one-shot graph generators,'' 2022.

\bibitem{chien2021node}
E.~Chien, W.-C. Chang, C.-J. Hsieh, H.-F. Yu, J.~Zhang, O.~Milenkovic, and I.~S. Dhillon, ``Node feature extraction by self-supervised multi-scale neighborhood prediction,'' \emph{arXiv preprint arXiv:2111.00064}, 2021.

\bibitem{hamilton2017inductive}
W.~Hamilton, Z.~Ying, and J.~Leskovec, ``Inductive representation learning on large graphs,'' \emph{Advances in neural information processing systems}, vol.~30, 2017.

\bibitem{morris2020tudataset}
C.~Morris, N.~M. Kriege, F.~Bause, K.~Kersting, P.~Mutzel, and M.~Neumann, ``Tudataset: A collection of benchmark datasets for learning with graphs,'' \emph{arXiv preprint arXiv:2007.08663}, 2020.

\bibitem{dwivedi2020benchmarking}
V.~P. Dwivedi, C.~K. Joshi, A.~T. Luu, T.~Laurent, Y.~Bengio, and X.~Bresson, ``Benchmarking graph neural networks,'' \emph{arXiv preprint arXiv:2003.00982}, 2020.

\bibitem{stankovic2019graph}
L.~Stankovic, D.~Mandic, M.~Dakovic, M.~Brajovic, B.~Scalzo, and A.~G. Constantinides, ``Graph signal processing--part ii: Processing and analyzing signals on graphs,'' \emph{arXiv preprint arXiv:1909.10325}, 2019.

\bibitem{Bengio+chapter2007}
Y.~Bengio and Y.~LeCun, ``Scaling learning algorithms towards {AI},'' in \emph{Large Scale Kernel Machines}.\hskip 1em plus 0.5em minus 0.4em\relax MIT Press, 2007.

\bibitem{Hinton06}
G.~E. Hinton, S.~Osindero, and Y.~W. Teh, ``A fast learning algorithm for deep belief nets,'' \emph{Neural Computation}, vol.~18, pp. 1527--1554, 2006.

\bibitem{goodfellow2016deep}
I.~Goodfellow, Y.~Bengio, A.~Courville, and Y.~Bengio, \emph{Deep learning}.\hskip 1em plus 0.5em minus 0.4em\relax MIT Press, 2016, vol.~1.

\bibitem{gasteiger2018predict}
J.~Gasteiger, A.~Bojchevski, and S.~G{\"u}nnemann, ``Predict then propagate: Graph neural networks meet personalized pagerank,'' \emph{arXiv preprint arXiv:1810.05997}, 2018.

\bibitem{gao2023graph}
X.~Gao, T.~Chen, Y.~Zang, W.~Zhang, Q.~V.~H. Nguyen, K.~Zheng, and H.~Yin, ``Graph condensation for inductive node representation learning,'' 2023.

\bibitem{mtt}
G.~Cazenavette, T.~Wang, A.~Torralba, A.~A. Efros, and J.-Y. Zhu, ``Dataset distillation by matching training trajectories,'' in \emph{Proceedings of the IEEE/CVF Conference on Computer Vision and Pattern Recognition}, 2022, pp. 4750--4759.

\bibitem{app13169166}
\BIBentryALTinterwordspacing
R.~Mao, W.~Fan, and Q.~Li, ``Gcare: Mitigating subgroup unfairness in graph condensation through adversarial regularization,'' \emph{Applied Sciences}, vol.~13, no.~16, 2023. [Online]. Available: \url{https://www.mdpi.com/2076-3417/13/16/9166}
\BIBentrySTDinterwordspacing

\bibitem{schuetzke2022universal}
J.~Schuetzke, N.~J. Szymanski, and M.~Reischl, ``A universal synthetic dataset for machine learning on spectroscopic data,'' 2022.

\bibitem{zhang2024graph1}
G.~Zhang, K.~Wang, W.~Huang, Y.~Yue, Y.~Wang, R.~Zimmermann, A.~Zhou, D.~Cheng, J.~Zeng, and Y.~Liang, ``Graph lottery ticket automated,'' in \emph{The Twelfth International Conference on Learning Representations}, 2024.

\bibitem{zhang2024two}
G.~Zhang, Y.~Yue, K.~Wang, J.~Fang, Y.~Sui, K.~Wang, Y.~Liang, D.~Cheng, S.~Pan, and T.~Chen, ``Two heads are better than one: Boosting graph sparse training via semantic and topological awareness,'' \emph{arXiv preprint arXiv:2402.01242}, 2024.

\bibitem{zhang2024graph2}
G.~Zhang, X.~Sun, Y.~Yue, K.~Wang, T.~Chen, and S.~Pan, ``Graph sparsification via mixture of graphs,'' \emph{arXiv preprint arXiv:2405.14260}, 2024.

\bibitem{wu2019simplifying}
F.~Wu, A.~Souza, T.~Zhang, C.~Fifty, T.~Yu, and K.~Weinberger, ``Simplifying graph convolutional networks,'' in \emph{International conference on machine learning}.\hskip 1em plus 0.5em minus 0.4em\relax PMLR, 2019, pp. 6861--6871.

\bibitem{defferrard2016convolutional}
M.~Defferrard, X.~Bresson, and P.~Vandergheynst, ``Convolutional neural networks on graphs with fast localized spectral filtering,'' \emph{Advances in neural information processing systems}, vol.~29, 2016.

\bibitem{Sandryhaila_2013}
\BIBentryALTinterwordspacing
A.~Sandryhaila and J.~M.~F. Moura, ``Discrete signal processing on graphs,'' \emph{{IEEE} Transactions on Signal Processing}, vol.~61, no.~7, pp. 1644--1656, apr 2013. [Online]. Available: \url{https://doi.org/10.1109%2Ftsp.2013.2238935}
\BIBentrySTDinterwordspacing

\bibitem{stamatelatos2022exact}
G.~Stamatelatos and P.~S. Efraimidis, ``An exact, linear time barab\'asi-albert algorithm,'' 2022.

\bibitem{Erd_s_2013}
\BIBentryALTinterwordspacing
L.~Erd{\H{o}}s, A.~Knowles, H.-T. Yau, and J.~Yin, ``Spectral statistics of erd{\H{o}}s{\textendash}r{\'{e}}nyi graphs i: Local semicircle law,'' \emph{The Annals of Probability}, vol.~41, no.~3B, may 2013. [Online]. Available: \url{https://doi.org/10.1214%2F11-aop734}
\BIBentrySTDinterwordspacing

\bibitem{velivckovic2017graph}
P.~Veli{\v{c}}kovi{\'c}, G.~Cucurull, A.~Casanova, A.~Romero, P.~Lio, and Y.~Bengio, ``Graph attention networks,'' \emph{arXiv preprint arXiv:1710.10903}, 2017.

\bibitem{Song_2014}
\BIBentryALTinterwordspacing
H.~F. Song and X.-J. Wang, ``Simple, distance-dependent formulation of the watts-strogatz model for directed and undirected small-world networks,'' \emph{Physical Review E}, vol.~90, no.~6, dec 2014. [Online]. Available: \url{https://doi.org/10.1103%2Fphysreve.90.062801}
\BIBentrySTDinterwordspacing

\bibitem{ma2018streaming}
Y.~Ma, Z.~Guo, Z.~Ren, E.~Zhao, J.~Tang, and D.~Yin, ``Streaming graph neural networks,'' 2018.

\bibitem{Jin_2022}
\BIBentryALTinterwordspacing
W.~Jin, X.~Tang, H.~Jiang, Z.~Li, D.~Zhang, J.~Tang, and B.~Yin, ``Condensing graphs via one-step gradient matching,'' in \emph{Proceedings of the 28th {ACM} {SIGKDD} Conference on Knowledge Discovery and Data Mining}.\hskip 1em plus 0.5em minus 0.4em\relax {ACM}, aug 2022. [Online]. Available: \url{https://doi.org/10.1145%2F3534678.3539429}
\BIBentrySTDinterwordspacing

\bibitem{hamilton2018inductive}
W.~L. Hamilton, R.~Ying, and J.~Leskovec, ``Inductive representation learning on large graphs,'' 2018.

\bibitem{S3GCL_ICML24}
G.~Wan, Y.~Tian, W.~Huang, N.~V. Chawla, and M.~Ye, ``S3gcl: Spectral, swift, spatial graph contrastive learning,'' in \emph{Forty-first International Conference on Machine Learning}, 2024.

\bibitem{FGSSL_IJCAI23}
W.~Huang, G.~Wan, M.~Ye, and B.~Du, ``Federated graph semantic and structural learning,'' in \emph{Proceedings of the Thirty-Second International Joint Conference on Artificial Intelligence}, 2023, pp. 3830--3838.

\bibitem{Epi_Survey_KDD24}
Z.~Liu, G.~Wan, B.~A. Prakash, M.~S. Lau, and W.~Jin, ``A review of graph neural networks in epidemic modeling,'' in \emph{Proceedings of the 30th ACM SIGKDD Conference on Knowledge Discovery and Data Mining}, 2024, pp. 6577--6587.

\bibitem{balın2023layerneighbor}
M.~F. Balın and Ümit V.~Çatalyürek, ``Layer-neighbor sampling -- defusing neighborhood explosion in gnns,'' 2023.

\bibitem{EARTH_Arxiv}
G.~Wan, Z.~Liu, M.~S. Lau, B.~A. Prakash, and W.~Jin, ``Epidemiology-aware neural ode with continuous disease transmission graph,'' \emph{arXiv preprint arXiv}, 2024.

\bibitem{wang2024heterophilic}
K.~Wang, G.~Zhang, X.~Zhang, J.~Fang, X.~Wu, G.~Li, S.~Pan, W.~Huang, and Y.~Liang, ``The heterophilic snowflake hypothesis: Training and empowering gnns for heterophilic graphs,'' in \emph{Proceedings of the 30th ACM SIGKDD Conference on Knowledge Discovery and Data Mining}, 2024, pp. 3164--3175.

\bibitem{NEURIPS2020_a7789ef8}
\BIBentryALTinterwordspacing
M.~Chen, Z.~Wei, B.~Ding, Y.~Li, Y.~Yuan, X.~Du, and J.-R. Wen, ``Scalable graph neural networks via bidirectional propagation,'' in \emph{Advances in Neural Information Processing Systems}, H.~Larochelle, M.~Ranzato, R.~Hadsell, M.~Balcan, and H.~Lin, Eds., vol.~33.\hskip 1em plus 0.5em minus 0.4em\relax Curran Associates, Inc., 2020, pp. 14\,556--14\,566. [Online]. Available: \url{https://proceedings.neurips.cc/paper_files/paper/2020/file/a7789ef88d599b8df86bbee632b2994d-Paper.pdf}
\BIBentrySTDinterwordspacing

\bibitem{zeng2020graphsaint}
H.~Zeng, H.~Zhou, A.~Srivastava, R.~Kannan, and V.~Prasanna, ``Graphsaint: Graph sampling based inductive learning method,'' 2020.

\bibitem{li2022training}
G.~Li, M.~Müller, B.~Ghanem, and V.~Koltun, ``Training graph neural networks with 1000 layers,'' 2022.

\bibitem{fey2019fast}
M.~Fey and J.~E. Lenssen, ``Fast graph representation learning with pytorch geometric,'' \emph{arXiv preprint arXiv:1903.02428}, 2019.

\bibitem{csikvári2022note}
P.~Csikvári, ``Note on the sum of the smallest and largest eigenvalues of a triangle-free graph,'' 2022.

\bibitem{yu2011limit}
H.~Yu, ``On the limit points of the smallest eigenvalues of regular graphs,'' 2011.

\bibitem{lin2023laplacian}
Z.~Lin, J.~Wang, and M.~Cai, ``The laplacian spectral ratio of connected graphs,'' 2023.

\bibitem{ao2022spectral}
G.~Ao, R.~Liu, and J.~Yuan, ``Spectral radius and spanning trees of graphs,'' 2022.

\bibitem{gallier2016spectral}
J.~Gallier, ``Spectral theory of unsigned and signed graphs. applications to graph clustering: a survey,'' 2016.

\bibitem{gao2017spectral}
J.~Gao and X.~Hou, ``The spectral radius of graphs without long cycles,'' 2017.

\bibitem{fan2022toughness}
D.~Fan, H.~Lin, and H.~Lu, ``Toughness, hamiltonicity and spectral radius in graphs,'' 2022.

\bibitem{sharma2019note}
R.~Sharma, A.~Sharma, and R.~Saini, ``A note on variance bounds and location of eigenvalues,'' 2019.

\bibitem{Min_2016}
\BIBentryALTinterwordspacing
C.~Min and Y.~Chen, ``On the variance of linear statistics of hermitian random matrices,'' \emph{Acta Physica Polonica B}, vol.~47, no.~4, p. 1127, 2016. [Online]. Available: \url{https://doi.org/10.5506%2Faphyspolb.47.1127}
\BIBentrySTDinterwordspacing

\end{thebibliography}
